\documentclass[12pt]{article}

\usepackage{bbding}

\usepackage{amsmath,amsfonts}
\usepackage{graphicx,psfrag,epsf}
\usepackage{enumerate}

\usepackage{natbib}

\usepackage{url} 
\usepackage[nodisplayskipstretch]{setspace}
\usepackage{mathtools}
\usepackage{hyperref}
\usepackage{epsfig}
\usepackage{epstopdf}
\usepackage{wrapfig}
\usepackage{color}

\usepackage{booktabs, multirow}
\usepackage{nicefrac}
\usepackage{microtype}
\usepackage{times}
\usepackage{adjustbox}
\usepackage{wasysym}
\usepackage{footnote}
\usepackage{enumitem}

\usepackage{comment}
\usepackage{cleveref}

\usepackage{caption}
\usepackage[flushleft]{threeparttable}

\newtheorem{lemma}{Lemma}
\newtheorem{definition}{Definition}
\newtheorem{theorem}{Theorem}
\newtheorem{corollary}{Corollary}
\newtheorem{assumption}{Assumption}

\usepackage{algorithm}
\usepackage{algpseudocode}

\DeclareMathOperator*{\argmax}{argmax}

\usepackage{array}
\newcolumntype{L}[1]{>{\raggedright\let\newline\\\arraybackslash\hspace{0pt}}m{#1}}
\newcolumntype{C}[1]{>{\centering\let\newline\\\arraybackslash\hspace{0pt}}m{#1}}
\newcolumntype{R}[1]{>{\raggedleft\let\newline\\\arraybackslash\hspace{0pt}}m{#1}}

\newcommand{\blind}{0}

\newtheorem{innercustomthm}{Lemma}
\newenvironment{customthm}[1]
{\renewcommand\theinnercustomthm{#1}\innercustomthm}
{\endinnercustomthm}

\newtheorem{innercustomdef}{Definition}
\newenvironment{customtdef}[1]
{\renewcommand\theinnercustomdef{#1}\innercustomdef}
{\endinnercustomdef}

\newtheorem{innercustomassum}{Assumption}
\newenvironment{customassum}[1]
{\renewcommand\theinnercustomassum{#1}\innercustomassum}
{\endinnercustomthm}

\begin{document}

\def\spacingset#1{\renewcommand{\baselinestretch}%
{#1}\small\normalsize} \spacingset{1}


\if0\blind
{
  \title{\bf Deep Distributional Learning with Non-crossing Quantile Network}
  \author{Guohao Shen\thanks{Dr. Shen is partially supported by the Hong Kong Research Grants Council (Grant No. 15305523) and research grants from The Hong Kong Polytechnic University.}\\
    Department of Applied Mathematics, The Hong Kong Polytechnic University\\
    Runpeng Dai \\
    Department of Biostatistics, University of North Carolina at Chapel Hill\\
    Guojun Wu \\
    ByteDance, Beijing, China\\
    Shikai Luo \\
    ByteDance, Beijing, China\\
    Chengchun Shi\thanks{Dr. Shi is partially supported by an EPSRC grant EP/W014971/1.} \\
    Department of Statistics, London School of Economics and Political Science \\
    Hongtu Zhu\thanks{Dr. Zhu is partially supported by the Gillings Innovation Laboratory on generative AI. Drs. Luo, Shi, and Zhu are the senior authors of this paper. } \\
    Department of Biostatistics, University of North Carolina at Chapel Hill}
    \date{}
  \maketitle
} \fi


\bigskip
\begin{abstract}
In this paper, we introduce a non-crossing quantile (NQ) network for conditional distribution learning. By leveraging non-negative activation functions, the NQ network ensures that the learned distributions remain monotonic, effectively addressing the issue of quantile crossing. Furthermore, the NQ network-based deep distributional learning framework is highly adaptable, applicable to a wide range of applications, from classical non-parametric quantile regression to more advanced tasks such as causal effect estimation and distributional reinforcement learning (RL). We also develop a comprehensive theoretical foundation for the deep NQ estimator and its application to distributional RL, providing an in-depth analysis that demonstrates its effectiveness across these domains. Our experimental results further highlight the robustness and versatility of the NQ network.
\end{abstract}

\noindent%
\vfill

\newpage
\spacingset{1.75} 
\addtolength{\textheight}{.5in}%

\section{Introduction}
\label{introduction}



Deep neural networks are extensively utilized across various domains due to their capability to capture complex patterns in the dataset. Most deep learning algorithms focus on learning the conditional expectation of an outcome. However, in many applications, it is more beneficial to estimate the entire distribution of an outcome to understand its variability and uncertainty. For instance, in applications like causal effect estimation or policy evaluation, where the reward distribution might be skewed or heavy-tailed, quantile-based metrics are preferred due to their robustness \citep{xu2022quantile, guojun2023dnet, zhang2023estimation, li2024evaluating, kallus2019localized, cheng2022doubly}. In reinforcement learning (RL), estimating the full distribution of returns rather than just their average can greatly improve policy optimization \citep{bellemare2017distributional, dabney2018distributional, sun2022does} and safeguard against undesirable outcomes \citep{wang2018quantile, fang2023fairness, qi2023robustness}.

In quantile regression, there is a longstanding problem of lack of monotonicity in the estimated conditional and structural quantile functions, also known as the crossing problem \citep{bassett1982empirical}.
To address this crossing problem, a variety of methods have been proposed. These methods can be roughly divided into two categories. The first category studies parametric quantile regression, including the location-scale regression estimator \citep{he1997quantile}, the constrained regression estimator \citep{koenker2005inequality,bondell2010noncrossing}, and the isotonic estimator \citep{mammen1991nonparametric,chernozhukov2010quantile}. The second category focuses on non-parametric quantile regression, especially for deep quantile regression \citep{zhou2020non,brando2022deep,padilla2022quantile,guojun2023dnet,yan2023ensemble,shen2024nonparametric}. We discuss these methods in Section \ref{related_work} in detail.


While the aforementioned methods have been applied to a variety of applications, much less is known regarding the theoretical properties of these estimations. This paper aims to introduce and study a general framework for deep distributional learning using a non-crossing quantile (NQ) network -- a neural network architecture designed to address the lack of monotonicity in distributional learning. We delve into a detailed theoretical analysis of deep NQ learning, demonstrating that it produces non-crossing quantile estimators that enjoy a minimax optimal rate of convergence. Our empirical results further illustrate the efficacy of deep NQ learning across diverse applications. 
To summarize, our major contributions are as follows:

\begin{enumerate}
    \item[(a)] We propose a deep distributional learning framework using a non-crossing quantile (NQ) network to handle quantile crossing problems. The NQ network ensures the quantile estimators are non-crossing and are flexible to be applied to a variety of distributional learning tasks, including deep reinforcement learning and causal inference.

    \item[(b)]  We lay the theoretical foundations of distributional learning using NQ networks. We establish the upper bounds of the prediction error of the quantile estimators, which achieve the minimax optimal rate of convergence. We also show that the NQ network can adapt to the low-dimensional structure of data to mitigate the curse of dimensionality. 
    \item[(c)] We provide novel theoretical guarantees for the application of the proposed NQ network to distributional reinforcement learning (DRL). Specifically, our analysis relaxes two commonly imposed restrictive conditions in the RL literature: 
    (i) We require only that the reward function has a bounded $p$th order absolute moment for any $p>1$, which is more flexible than the typical bounded or sub-Gaussian reward assumptions found in existing studies. This formally demonstrates the advantages of DRL over classical RL algorithms in handling heavy-tailed rewards. While these benefits have been observed empirically and discussed in the context of policy evaluation, our work rigorously prove these insights in the policy learning setting using deep learning.
    (ii) Our analysis does not assume that the data are independent and identically distributed \citep[i.i.d.,][]{chen2019information}, nor does it rely on mixing or ergodicity conditions \citep[see e.g.,][]{ramprasad2023online}. These assumptions, though common in the literature, often do not hold in practical RL applications, making our approach more applicable to real-world scenarios.

\end{enumerate}

\begin{table*}[t]
	\caption{A comparison of recent results on quantile neural networks.}
	\vskip 0.1in
	\label{tab:compare}
	\begin{tabular}{cc c c c c}
			\toprule
			Paper &  One-step & Non-cross & Theory & Low-dim Result & Optimal Rate \\ \midrule
                \cite{zhou2020non} &  \Checkmark & \Checkmark & \XSolidBrush & \XSolidBrush & \textbf{\large ?} \\
			 \cite{padilla2022quantile} & \Checkmark & \Checkmark & \XSolidBrush & \XSolidBrush & \textbf{\large ?} \\
	
			 \cite{yan2023ensemble} & \XSolidBrush & \Checkmark & \XSolidBrush & \XSolidBrush & \textbf{\large ?}  \\
            \cite{shen2024nonparametric} & \Checkmark &  \textbf{\large ?} & \Checkmark & \XSolidBrush & \XSolidBrush  \\
    
			This paper  & \Checkmark & \Checkmark & \Checkmark & \Checkmark & \Checkmark\\ 
			\bottomrule
		\end{tabular}%
	
    {\raggedleft \footnotesize \singlespacing Notes on the question marks:  The non-crossing constraint is not 
    strictly imposed in \cite{shen2024nonparametric}. 
    Learning guarantees for non-crossing quantile estimators are not derived in \cite{zhou2020non,padilla2022quantile,yan2023ensemble} and the optimality is unclear.}

\end{table*}

The rest of this paper is structured as follows:
Section \ref{related_work} reviews related works on distributional learning and quantile regression, with a particular focus on nonparametric methods, since they are closely related to our proposal. 
Section \ref{methodology} introduces the proposed framework for distributional learning using the NQ Network. Section \ref{sec_theory} establishes the learning guarantees including the minimax optimality and its capability to mitigate the curse of dimensionality. 
Section \ref{example} investigates the adaptation of the NQ network to DRL and derives the corresponding learning theories. 
Section \ref{sec_numerical} empirically 
compares the NQ network-based quantile regression against existing deep non-crossing quantile learning methods. We conclude our paper in Section \ref{conclusion}.  Additional simulation results, details of our simulation implementation, details of all the proofs, and supporting lemmas can be found in the appendix.

{\singlespacing\section{A Review of Non-parametric Quantile and Distributional Learning Methods}\label{related_work}}
\noindent Over the past twenty years, non-parametric methods for estimating conditional quantile functions have been proposed in the literature. 
The classical methods do not use deep learning, relying on traditional nonparametric statistical and machine learning techniques, such as random forests \citep{meinshausen2006quantile}, smoothing splines and kernels \citep{bondell2010noncrossing, sangnier2016joint,he2022scalable,HE2023367}, and vector-valued reproducing kernel Hilbert spaces \citep[RKHS,][]{takeuchi2006nonparametric}. For spline- and RKHS-based methods, hard constraints were imposed in the aforementioned papers to ensure the monotonicity of the estimated conditional quantiles. There are also generic methods such as inversion, monotonization and rearrangement \citep[see e.g.,][]{dette2008non, chernozhukov2010quantile} that are not reliant on particular machine learning algorithms designed to address the crossing issues. Furthermore, non-deep learning-based methods have also been explored in DRL \citep[see e.g.,][]{zhang2023estimation}.

Recently, 
deep learning-based methods 
have been developed to non-parametrically estimate the conditional distribution or quantile functions. 
In particular, \citet{shen2021deep} and \citet{padilla2022quantile} studied the deep quantile regressions for a single quantile and showed that ReLU-activated neural networks achieve the minimax optimal rate for non-parametric quantile regressions. \cite{padilla2022quantile} 
discussed the extension 
to simultaneously estimate multiple quantile functions across different levels while ensuring the monotonicity of the resulting estimator. The idea is to jointly estimate a lower-level quantile function and the non-negative gaps between this quantile and higher quantiles using a ReLU neural network with a positive activation function in the output layer. Non-crossing quantile neural networks based on similar ideas have been subsequently developed and applied to distributional causal learning and distributional reinforcement learning \citep{zhou2020non,guojun2023dnet,yan2023ensemble}. 
However, as 
both the neural network architecture and the algorithm are quite complicated, 
there is a lack of learning theories for the aforementioned estimators. 
Instead of estimating the conditional quantiles at multiple levels, \citet{brando2022deep} and \citet{shen2024nonparametric} proposed to model the entire conditional quantile process and address the quantile crossing problem by restricting or encouraging the positivity of the partial derivative of the quantile process with respect to the quantile level. Specifically, \cite{shen2024nonparametric} employed the rectified quadratic units (ReQU) activated neural networks to 
estimate the quantile process, and proposed to penalize the estimator's partial derivative to encourage the monotonicity or the non-crossing of the quantile process. The resulting estimator is shown to attain a nonparametric rate of convergence and the crossing issues can be controlled to some degree.

 In this work, we propose a non-crossing quantile network using a non-negative activation function for the nonparametric distributional learning which ensures the estimated quantiles are non-crossing. We show that the learned quantiles enjoy a minimax optimal rate of convergence toward the target quantiles and can adapt to the low-dimensional structure of the high-dimensional data to mitigate the curse of dimensionality. Our proposal is closely related to, yet different from, the aforementioned deep learning-based works, both theoretically and methodologically. These differences are summarized in Table \ref{tab:compare}.

\section{Methodology}\label{methodology}

In this section, we introduce the framework of deep distributional learning, present the proposed NQ network structure, and detail our 
estimating procedure.

\subsection{Background and problem formulation} 
 
Let $(X, Y)$ be a pair of 
$d_0$-dimensional predictor and univariate response following the joint distribution $P_{X, Y}$. Unlike traditional regression methods that estimate the conditional mean $\mathbb{E}(Y\mid X)$, quantile regression concerns the $\tau$th conditional quantile of the response given the predictors $X=x$, i.e., $Q^\tau_Y(x) = F^{-1}_{Y\mid X=x}(\tau),$
for $\tau\in(0,1)$,  where $F^{-1}_{Y\mid X=x}$ denotes the inverse of the cumulative distribution function of $Y$ given $X=x$. Given $\tau\in(0,1)$, the $\tau$th conditional quantile can be consistently estimated by minimizing the expected check loss function $\mathbb{E}_{X, Y}[\rho_\tau(Y-f(X))]$ over some proper class of functions $\mathcal{F}$, 
where $\rho_{\tau}(a)=a[\tau-1(a<0)]$ denotes the check loss function.

\begin{figure}[H]\centering\includegraphics[width=0.75\textwidth]{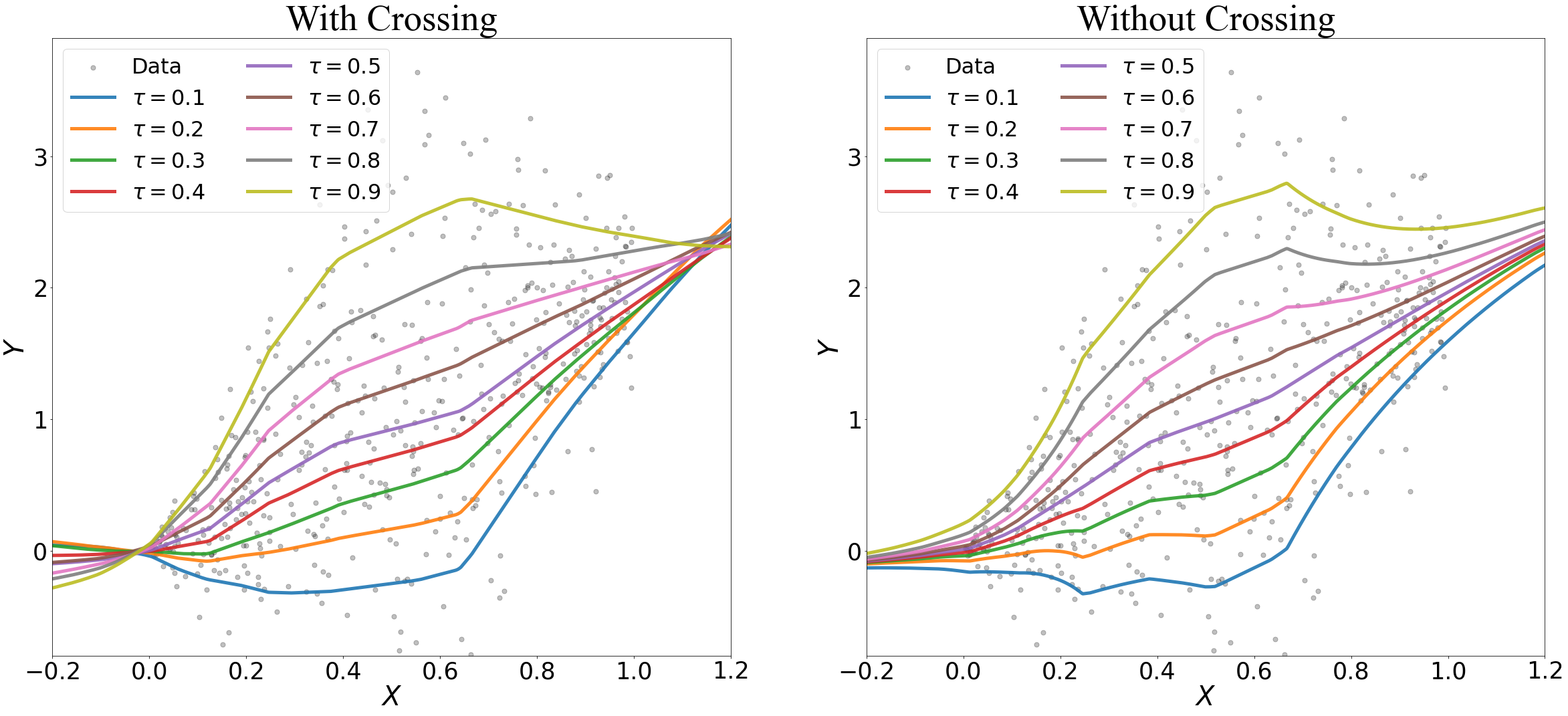}
    \caption{A demonstration of quantile crossing on a simulated dataset. The estimated quantile curves at $\tau=0.1, 0.2,\ldots, 0.9$ and the observations are depicted. The left panel presents the estimates from the deep quantile regression without any constraint and there appear crossings. The right figure presents our proposed quantile estimations with non-crossing constraints and there is no crossing.
    }
    \label{fig:demo}
\end{figure}

Distributional learning provides a more comprehensive understanding of the relationship between the predictors and the response by estimating different quantiles of the conditional distribution. Let $\tau = (\tau_1, \ldots, \tau_K)$ denote a set of $K$ pre-specified non-decreasing quantile levels, the objective of distributional learning is to model and estimate the conditional quantiles $Q^{\tau_1}_Y(x),\ldots, Q^{\tau_K}_Y(x)$ by minimizing the aggregated objective functions
\begin{equation}\label{obj}
    \sum_{k=1}^K\mathbb{E}_{X, Y}[\rho_{\tau_k}(Y-f_k(X))]
\end{equation}
over some hypothesis space of $f_1,\ldots,f_K$. 
A common issue in quantile regression is that the learned quantile curves, denoted as $\hat{f}_1, \ldots, \hat{f}_K$, may exhibit crossing-quantile problems, where the intended constraint $\hat{f}_1(x) \leq \hat{f}_2(x) \leq \ldots \leq \hat{f}_K(x)$ is not maintained for some $x$; see Figure~\ref{fig:demo} for a demonstration of the crossing issues. This violation not only reduces the accuracy of the estimations but also hurts the model's interpretability \citep{zhou2020non,li2024efficient}.

\subsection{Non-Crossing Quantile Networks}
In our proposed distributional learning framework, we model the quantile curves $f_1,\ldots,f_K$ in the objective function \eqref{obj} by using deep neural networks. Specifically, let $\tau = (\tau_1, \ldots, \tau_K)$ denote a set of $K$ pre-specified non-decreasing quantile levels, typically chosen as $\tau_k =  {k}/(K+1)$ for $k=1,\ldots,K$. Our proposed NQ networks $f(x;\theta)$ output the vector
$f(x;\theta) = (f_1(x;\theta),\ldots,f_K(x;\theta))$, 
where $f(\cdot; \cdot)$ is a certain neural network model, $x$ is the input data, $\theta$ is the set of parameters in the network $f$, and $f_{k}(x;\theta)$ estimates the $\tau_k$th conditional quantile of $Y$ given $X=x$. However, without any constraints, 
the output may not be monotonic and satisfies ${f}_1(x;\theta)\le{f}_2(x;\theta)\le\ldots\le{f}_K(x;\theta)$.  
  See Figure~\ref{fig:demo} for a demonstration of the crossing issues.

\begin{figure}[t]
    \centering
    \includegraphics[width=0.6\linewidth]{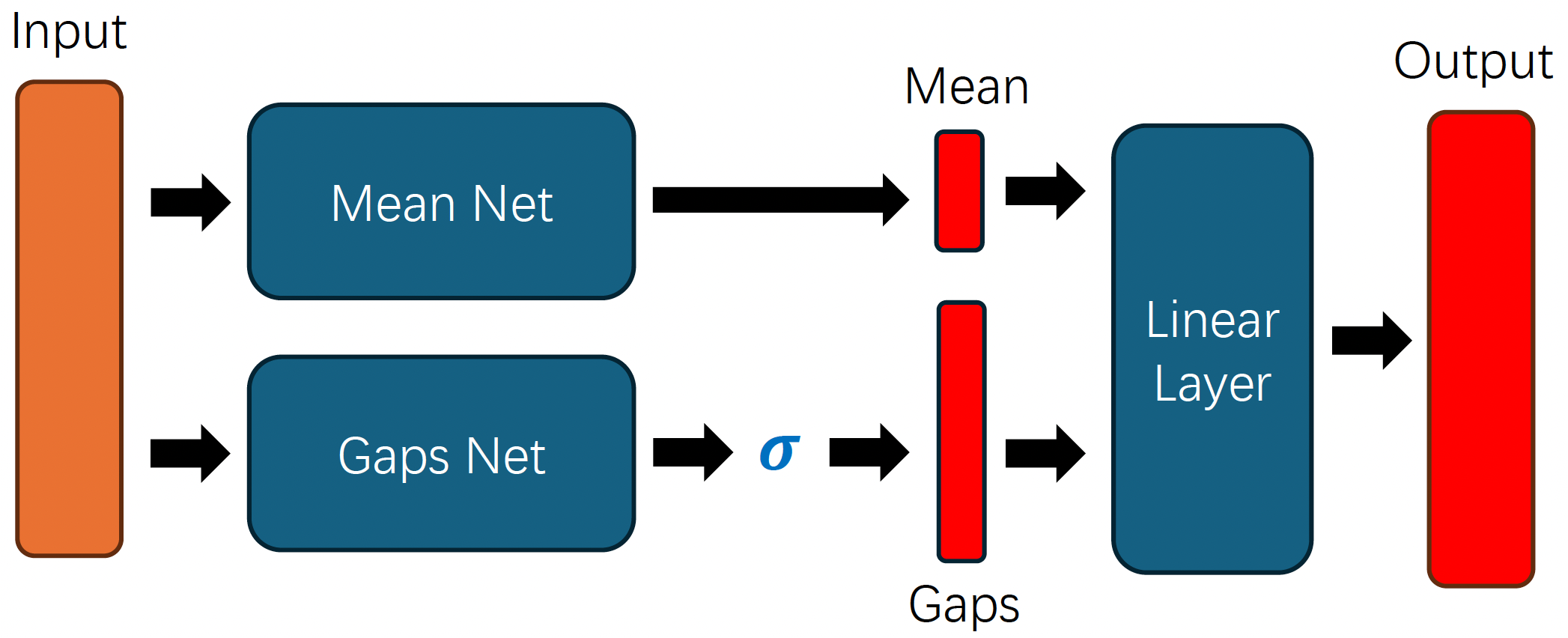}
    \caption{A graphical illustration of the Non-Crossing Quantile Network. The ``Mean Net" aims to learn the average of all quantiles and the ``Gaps Net" aims to learn the differences between adjacent quantiles. }
    \label{fig:ncq_illu}
   \vskip -0.5cm  
\end{figure}

To address the crossing issue, we propose the following design for the internal neural network architecture (see Figure~\ref{fig:ncq_illu} for a visualization). Specifically, the proposed NQ network computes through two different modules: the Mean network and the Gaps network. The Mean network focuses on predicting the average values of the target quantiles, while the vector-valued Gaps network estimates the non-negative differences between consecutive quantiles.
  More formally, let $v(\cdot;\theta)$ and $g(\cdot;\theta)$ denote the \textit{Mean} and  \textit{Pre-activated Gaps}, respectively, computed from networks parameterized by $\theta$. We use the concatenated vector $(v(\cdot;\theta),g(\cdot;\theta))$ to denote the real-valued output of the neural network. The \textit{Pre-activated Gaps} $g=(g_1,\ldots,g_K)$ can take negative values, we apply $\sigma(x) = ELU(x)+1$ activation on $g$ to get the  non-negative \textit{Gaps} $\sigma(g)$ for predicting the quantile gaps, where $ELU(x)$ is an exponential linear unit function, defined by $ELU(x) =I(x\geq0)\cdot x, + I(x<0)\cdot(\exp(x) - 1)$. The $ELU$-function rectifies the input to $(-1,+\infty)$ so that $\sigma(x)=ELU(x)+1$ is strictly positive. With the activated \textit{Gaps} $\sigma(g)=(\sigma(g_1),\ldots,\sigma(g_K))$ and the {\it Mean} $v$, the final output of NQ network $f=(f_1,\ldots,f_K)$ is computed by an additional linear layer as follows
\begin{align}\label{nq}
f_{k} &=v-\bar{g} +\sum_{i=1}^k \sigma(g_i),  \quad {\rm where}\quad \bar{g} = \frac{1}{K}\sum_{j=1}^K (K+1-j)\cdot \sigma(g_j).
\end{align}
It can be verified that $v={K}^{-1}\sum_{k=1}^K f_k$ and $\sigma(g_{k+1})= f_{k+1}-f_{k}$,  aligning with the definitions of  \textit{Mean} and \textit{Gaps}.  
It is important to note that the NQ network design is highly adaptable. The Mean and Gaps networks can be implemented as either independent or interdependent neural networks with varying architectures. Furthermore, alternative non-negative functions could be used for the activation function 
$\sigma(\cdot)$ to preserve monotonicity in the outputs. It is worth mentioning that using ReLU as $\sigma$ can be problematic, as it may lead to unreasonable zero estimations for the gaps between conditional quantiles due to that ReLU is not strictly positive.

\subsection{Training of NQ Networks}\label{sec_training}
In this subsection, we 
describe the neural network architectures and introduce the loss function for training the proposed non-crossing quantile functions.

For a given sequence of $K$ non-decreasing quantiles and the input dimension $d_0$, we set our NQ network to a ReLU-activated feedforward neural network $f:\mathbb{R}^{d_0}\to \mathbb{R}^{K+1}$ with input dimension $d_0$ and output dimension $K+1$. We pick the first entry of the $(K+1)$-dimensional vector output as the  {\it Mean} $v$ and apply $\sigma=ELU+1$  activation to the rest entries of output $(g_1,\ldots,g_K)$. Then the quantiles $f_1,\ldots,f_K$ can be calculated according to   \eqref{nq}. Additionally, we denote the number of hidden layers (depth), the maximum number of neurons in the hidden layers (width), the number of neurons, and the total number of parameters (size) including weights and biases in the NQ network by $\mathcal{D}, \mathcal{W}, \mathcal{U},$ and $ \mathcal{S}$,  respectively. Let
\begin{align}\label{NQ-networks}
\mathcal{F}_N:=\mathcal{F}_{\mathcal{D}, \mathcal{W}, \mathcal{U}, \mathcal{S},\mathcal{B}}
\end{align} 
denote the class of such NC networks with each entry of their output bounded by $\mathcal{B}$ and with architecture specified by $\mathcal{D}, \mathcal{W}, \mathcal{U}$ and $\mathcal{S}$. Here, the choices of hyper-parameters $\mathcal{D}, \mathcal{W}, \mathcal{U}, \mathcal{S}, \mathcal{B}$ and the network class $\mathcal{F}_N$ may depend on the sample size $N$, but we suppress the dependence of hyper-parameters $\mathcal{D}, \mathcal{W}, \mathcal{U}, \mathcal{S},\mathcal{B}$ on $N$ for simplicity of the presentation.

To train NQ networks, we use the quantile loss $\rho_\tau(u) = u\left( \tau - {\bf 1} \{u<0\} \right)$. Given a sample $\{(X_i, Y_i)_{i=1}^N\}$ with size $N$ and a sequence of quantile levels $\tau = (\tau_1, \ldots, \tau_K)$, we set the empirical risk for a NQ network $f$ by
\begin{equation}
    \mathcal{L}_N(f) = \frac{1}{N}\sum_{i=1}^N\frac{1}{K}\sum_{k=1}^{K}\rho_{\tau_k}(Y_i-f_{k}(X_i)),
    \label{eq: quantile_loss}
\end{equation}
where $f_{k}(X_i)$ is the $k$th qauntile output of the NQ network $f$ on sample $X_i$. Our target is to minimize the empirical risk over the class of NQ networks $\mathcal{F}_N$ to compute our estimator $\hat{f}_N$, i.e.,
\begin{equation}\label{erm}\hat{f}_N=\arg\min_{f\in\mathcal{F}_N}\mathcal{L}_N(f).
\end{equation}
We denote the risk by 
$\mathcal{L}(f) = \mathbb{E}_{X,Y}[{K}^{-1}\sum_{k=1}^{K}\rho_{\tau_k}(Y-f_{k}(X))].$
Obviously, the ground truth quantile curves $Q_Y:=(Q^{\tau_1}_Y,\ldots,Q^{\tau_K}_Y)$ are the minimizer of the risk $\mathcal{L}(\cdot)$. We call the ground truth $Q_Y$ as the target of the objective of NQ networks. It 
is expected that proper choices of $\mathcal{F}_N$ will lead to consistent estimates of the target $Q_Y$.

\section{Learning Guarantees}\label{sec_theory}

In this section, we establish the consistency of the proposed NQ network estimator and derive non-asymptotic upper bounds for its performance. The proofs of the lemmas and theorems are provided in the appendix.

Recall that the target quantile function, denoted by \( Q_Y \), minimizes the risk function \( \mathcal{L} \) and thus,  $\mathcal{L}(Q_Y)\le \mathcal{L}(f)$ holds for any estimator $f$. We define its excess risk by
\begin{align}\label{excess_risk}
    \mathcal{R}(f):=\mathcal{L}(f)-\mathcal{L}(Q_Y). 
\end{align}
 We aim to derive non-asymptotic upper bounds for $\mathcal{R}(\hat{f}_N)$ for $\hat{f}_N$ defined in (\ref{erm}). It is worth noting that under self-calibration conditions \citep{padilla2021adaptive}, the convergence of excess risk $\mathcal{R}(\hat{f}_N)$ toward zero implies the convergence of the estimator $\hat{f}_N$ towards the target $Q_Y$. We elucidate self-calibration results in the appendix. Let $\mathcal{R}_N(f):=\mathcal{L}_N(f)-\mathcal{L}_N(Q_Y)$ denote the empirical version of the excess risk, we first give preliminary bounds on the excess risks based on error decomposition.
\begin{lemma}\label{decomposition}
    The excess risk $\mathcal{R}(\hat{f}_N)$ of the empirical risk minimizer $\hat{f}_N$ defined in (\ref{erm}) satisfies
    $\mathbb{E}[\mathcal{R}(\hat{f}_N)]\le \mathbb{E}[\mathcal{R}(\hat{f}_N)-2\mathcal{R}_N(\hat{f}_N)] +2\inf_{f\in\mathcal{F}_N}\mathcal{R}(f),$
    where $\mathbb{E}[\mathcal{R}(\hat{f}_N)-2\mathcal{R}_N(\hat{f}_N)]$ is   the {\it variance} (or {\it stochastic error}) and $2\inf_{f\in\mathcal{F}_N}[\mathcal{R}(f)]$ is   the {\it bias} (or {\it approximation error}).
\end{lemma}

The \textit{bias}   measures the deviation between the regression functions (hypothesis class) $\mathcal{F}_N$ and the target function $Q_Y$. It reflects the expressiveness of the NQ network's hypothesis class, $\mathcal{F}_N$, and the properties of the target function $Q_Y$, such as its structure, continuity, and smoothness. Generally, richer classes of $\mathcal{F}_N$ and simpler target functions $Q_Y$ lead to reduced \textit{bias}. It is important to note that in most parametric methods, the model is assumed to be correctly specified, meaning the target $Q_Y$ belongs to the hypothesis class $\mathcal{F}_N$. Under such a realizability assumption, the \textit{bias} $\inf_{f \in \mathcal{F}_N} \mathcal{R}(f)$ is zero and the estimator's converges at a ``parametric'' rate.

The \textit{variance} measures the estimation error induced by the randomness of the samples. It is closely associated with the loss function, the sample size, and the properties of the hypothesis class $\mathcal{F}_N$. Typically, a smaller sample size $N$ and a more comprehensive hypothesis class $\mathcal{F}_N$ result in increased \textit{variance}. Variance analysis can be conducted using empirical process theories, which utilize neural network complexity analysis \citep{bartlett2019nearly}.

\subsection{Non-asymptotic error bounds without misspecification}

In this subsection, we operate under the premise that the model is correctly specified, indicated by \( Q_Y \in \mathcal{F}_N \).  
Under this assumption,, the \textit{bias} term \( \inf_{f \in \mathcal{F}_N} \mathcal{R}(f) \)  is effectively zero. This necessitates a focused analysis on the upper bounds of the \textit{variance}. Conducting this analysis enables us to establish an upper bound for the excess risk of the NQ-net estimator \( \hat{f}_N \).

\begin{theorem}\label{sto_error}
    Let the classes of neural networks $\mathcal{F}_N$ be defined as in Section \ref{sec_training} with depth $\mathcal{D}$, size $\mathcal{S}$ and bound $\mathcal{B}$, and the excess risk $\mathcal{R}$ be defined as in (\ref{excess_risk}), and suppose the target quantile curves $Q_Y=(Q_Y^{\tau_1},\ldots,Q_Y^{\tau_K})\in\mathcal{F}_N$. Then the NQ net estimator $\hat{f}_N=(\hat{f}^{(1)}_N.\ldots,\hat{f}^{(K)}_N)$ defined in (\ref{erm}) satisfies the monotonicity constraint
$\hat{f}^{(1)}_N< \cdots<\hat{f}^{(K)}_N$ and
\begin{align}\notag
    &\mathbb{E}[\mathcal{R}(\hat{f}_N)]\le C\cdot K\mathcal{B}^3\frac{\mathcal{S}\mathcal{D}\log(\mathcal{S})\log(N)}{N} 
\end{align}
for $N\ge C_0\mathcal{S}\mathcal{D}\log \mathcal{S}$,  where $C$ and $C_0>0$ are universal constants. 
\end{theorem}

The excess risk bound in Theorem \ref{sto_error} exhibits a decay of order \(O\left(\mathcal{S}\mathcal{D}\log(\mathcal{S})\log(N)/{N}\right)\). This rate decreases linearly with the sample size \(N\) and increases linearly with the network's complexity, which includes both size and depth. This convergence rate is notably superior to that derived from standard concentration inequalities for deep quantile networks, which typically exhibit rates of \(O\left((\mathcal{S}\mathcal{D}\log(\mathcal{S})\log(N)/{N})^{1/2}\right)\), as cited in recent studies \citep{padilla2022quantile, shen2024nonparametric} and other nonparametric methods \citep{takeuchi2006nonparametric, sangnier2016joint, mohri2018foundations}.
 Under self-calibration conditions \citep{padilla2021adaptive} (See Assumption A1 in the appendix), the convergence of the excess risk also implies  
\[
\sum_{k=1}^K\mathbb{E}|\hat{f}^{\tau_k}_{N}(X) - Q_Y^{\tau_k}(X)|^2 \leq C^\prime \cdot K \mathcal{B}^3 \frac{\mathcal{S}\mathcal{D}\log(\mathcal{S})\log(N)}{N},
\]
indicating a "parametric rate" of \(O(N^{-1})\) for the convergence of each  \(\hat{f}^{\tau_k}_{N}\).

\subsection{Non-asymptotic error bound with misspecification}

In nonparametric regression, if the target function is complex and falls outside the hypothesis class, the model is considered misspecified, leading to a nonzero bias or approximation error. 
The {\it bias} hinges on the representational capabilities of the hypothesis class     $\mathcal{F}_N$ and the characteristics of the target function $Q_Y$, such as its structure, continuity, and smoothness.
To quantify this error more precisely, we introduce the concept of H\"older smooth functions. This approach allows us to describe and measure the smoothness of the target function. We assume that each component of the target   $Q_Y=(Q_Y^{\tau_1}, \ldots, Q_Y^{\tau_K})$ is a H\"older smooth function. 

\begin{definition}\label{holder_function}
Given constants $B,\beta>0$, the H\"older class of functions $\mathcal{H}^\beta(\mathcal{X},B)$  is defined as
\begin{align}
\label{Hclass}
&\mathcal{H}^\beta(\mathcal{X},B) =
\Big\{f: \mathcal{X}\to\mathbb{R},\max_{\Vert\alpha\Vert_1\le s}\Vert\partial^\alpha f\Vert_\infty\le B,\max_{\Vert\alpha\Vert_1=s} \sup_{x\not=y}\frac{\vert\partial^\alpha f(x)-\partial^\alpha f(y)\vert}{\Vert x-y\Vert_2^r}\le B \Big\}, \nonumber
\end{align}
where $\partial^\alpha=\partial^{\alpha_1}\ldots\partial^{\alpha_d}$ with $\alpha=(\alpha_1,\ldots,\alpha_d)^\top\in\mathbb{N}_0^d$ and $\Vert\alpha\Vert_1=\sum_{i=1}^d\alpha_i$.
\end{definition}

The classes of Lipschitz continuous and continuously differentiable functions are  H\"older function classes. To analyze the {\it bias}, 
we make the following assumptions.

\begin{assumption}\label{assump1} 
(i) The target quantile curves $Q_Y=(Q_Y^{\tau_1},\ldots,Q_Y^{\tau_K})$ are $\beta$-H\"older smooth with constant $B$. (ii) The domain of the target function $Q_Y$ is $\mathcal{X}=[0,1]^d$. The probability distribution of covariate $X$ is absolutely continuous with respect to the Lebesgue measure.
\end{assumption}

Assumption \ref{assump1} (i) focuses on the smoothness of the target quantile curves $Q_Y=(Q_Y^{\tau_1},\ldots,Q_Y^{\tau_K})$, which is quantified by the smoothness parameter $\beta$. This parameter indicates the complexity involved in approximating the target function by the hypothesis class, directly influencing the approximation error \citep{schmidt2020nonparametric}. 
Assumption \ref{assump1} (ii) applies a standard condition that limits the covariate space to a bounded domain \citep{yarotsky2017error,guhring2020error}. While this domain is restricted, it can be significantly large, allowing for a broad range of applicability.

\begin{lemma}[Approximation Error bound]\label{app_error}
    Let the class of neural networks $\mathcal{F}_N$ be defined as in section \ref{sec_training} with depth $\mathcal{D}$, width $\mathcal{W}$ and bound $\mathcal{B}$ and let the excess risk $\mathcal{R}$ be defined as in (\ref{excess_risk}).
    Suppose Assumption \ref{assump1} holds with $B\le\mathcal{B}$. For NQ neural networks in $\mathcal{F}_N$, given any positive integers $U$ and  $M$, let the width $\mathcal{W}=38(K+1)(\lfloor\beta\rfloor+1)^2 d_1d_0^{\lfloor\beta\rfloor+1}U\log_2(8U)$ and depth $\mathcal{D}=21(\lfloor\beta\rfloor+1)^2d_0^{\lfloor\beta\rfloor+1}M\log_2(8M)$, then we have 
\begin{align}\notag
    \inf_{f\in\mathcal{F}_N}[\mathcal{R}(f)]&\le 18 (K+2)\mathcal{B}(\lfloor\beta\rfloor+1)^2 d_0^{\lfloor\beta\rfloor+1+(\beta\vee1)/2}(UM)^{-2\beta/d_0}+(K+2)\exp(-\mathcal{B}), 
\end{align}
where $d_0$ is the dimension of the input of NQ neural networks in $\mathcal{F}_N$.
\end{lemma}


The bias  in Lemma \ref{app_error} presents notable differences from traditional results found in the literature on quantile and robust regression \citep{lederer2020risk, padilla2021adaptive, padilla2022quantile, shen2024nonparametric}. First, the bias is proportional to \(K\), which is both the dimension of the NQ network output and the number of quantile curves being estimated. Second, an error term \(\exp(-\mathcal{B})\) emerges due to the use of a truncation technique to manage the unbounded preimage associated with the ELU activation function. This error term decays exponentially with the parameter \(\mathcal{B}\) and can be considered negligible in the error analysis, provided that \(\mathcal{B}\) increases appropriately.

The bias rate in Lemma \ref{app_error} can be further accelerated once the excess risk has a desirable local quadratic structure, as illustrated below.

\begin{assumption}[Local quadratic bound of the excess risk]
	\label{assump2}
There exist some constants $c^0=c^0(X,Y)>0$ and $\delta^0=\delta^0(X,Y)>0$ such that
	$$\mathcal{R}(f) -\mathcal{R}(Q_Y)\leq  \frac{c^0}{K}\sum_{k=1}^K \mathbb{E} \Vert f_{\tau_k}(X)-Q^{\tau_k}_Y(X) \Vert^2
 $$
	for any $f=(f_{\tau_1},\ldots,f_{\tau_K})$ satisfying $\Vert f_{\tau_k}-Q_Y^{\tau_k} \Vert_{L^\infty(\mathcal{X}^0)}\leq \delta^0$ for $k=1,\ldots,K$, where $\mathcal{X}^0$ is any subset of $\mathcal{X}$ such that $\mathbb{P}(X\in\mathcal{X}^0)=\mathbb{P}(X\in\mathcal{X})$.
\end{assumption}

The technical assumption of a local quadratic structure for the excess risk can be satisfied when the density function of the conditional distribution $Y$ given $X$ has an upper bound near $Q_Y^{\tau_k}(X)$ for $k=1,\ldots, K$. With Assumption \ref{assump2}, we have an improved approximation error bound.

\begin{lemma}\label{app_error1}
    Suppose the conditions in Lemma \ref{app_error}  and Assumption \ref{assump2} hold. For NQ neural networks in $\mathcal{F}_N$, given any positive integers $U$ and $M$, let the width $\mathcal{W}=38(K+1)(\lfloor\beta\rfloor+1)^2 d_1d_0^{\lfloor\beta\rfloor+1}U\log_2(8U)$ and depth $\mathcal{D}=21(\lfloor\beta\rfloor+1)^2d_0^{\lfloor\beta\rfloor+1}M\log_2(8M)$, then  we have 
\begin{align}\notag
    \inf_{f\in\mathcal{F}_N}[\mathcal{R}(f)]&\le C(K+2)^2\left[\mathcal{B}^2(\lfloor\beta\rfloor+1)^4d_0^{2\lfloor\beta\rfloor+(\beta\vee 1)}(UM)^{-4\beta/d_0} + \exp(-2\mathcal{B})\right],
\end{align}
where $C>0$ is a universal constant and $d_0$ is the input dimension of NQ neural networks in $\mathcal{F}_N$.
\end{lemma}

Based on the error decomposition in Lemma \ref{decomposition}, and our derived upper bounds for stochastic and approximation errors in Theorem \ref{sto_error} and Lemma \ref{app_error1}, we immediately obtain an upper bound for the excess risk $\mathcal{R}(\hat{f}_N)$ when the model is misspecified.

\begin{theorem}[Non-asymptotic Upper bounds]\label{nonasymp_error}
   Let the class of neural networks $\mathcal{F}_N$ be defined as in Section \ref{sec_training} with depth $\mathcal{D}$, width $\mathcal{W}$, size $\mathcal{S}$,  and bound $\mathcal{B}$ and let the excess risk $\mathcal{R}$ be defined as in (\ref{excess_risk}).  Suppose that Assumption \ref{assump1} holds with $B\le\mathcal{B}$ and Assumption \ref{assump2} holds. For NQ neural networks in $\mathcal{F}_N$, given any positive integers $U, M$, let the width $\mathcal{W}=38(K+1)(\lfloor\beta\rfloor+1)^2 d_1d_0^{\lfloor\beta\rfloor+1}U\log_2(8U)$ and depth $\mathcal{D}=21(\lfloor\beta\rfloor+1)^2d_0^{\lfloor\beta\rfloor+1}M\log_2(8M)$, then the NQ net estimator $\hat{f}_N=(\hat{f}^{(1)}_N.\ldots,\hat{f}^{(K)}_N)$ defined in (\ref{erm}) satisfies the monotonicity constraint
$\hat{f}^{(1)}_N< \cdots<\hat{f}^{(K)}_N$ and
{\small
\begin{align*}
\mathbb{E}[\mathcal{R}(\hat{f}_N)]&\le C_1(K+2)^2\left[\mathcal{B}^2(\lfloor\beta\rfloor+1)^4d_0^{2\lfloor\beta\rfloor+(\beta\vee 1)}(UM)^{-4\beta/d_0} + \exp(-2\mathcal{B})\right]+C_2\cdot \frac{K\mathcal{B}^3\mathcal{S}\mathcal{D}\log(\mathcal{S})}{(\log N)^{-1}N}
\end{align*}
}
for $N\ge c\cdot\mathcal{D}\mathcal{S}\log(\mathcal{S})$,  where $c,C_1,C_2>0$ are universal constants and $d_0$ is the input dimension of the target quantile functions $Q_Y$.
\end{theorem}


To optimize convergence rates relative to  \(N\), it is critical to balance the increasing stochastic error with network size against the decreasing bias with network width and depth. This balance should be tailored based on the ratio \(\beta/d_0\), where \(\beta\) denotes the smoothness of the target functions \(Q_Y\), and \(d_0\) represents their input dimension. Selecting the proper network architecture hinges on this ratio to effectively manage the trade-off between bias and stochastic error.

\begin{corollary}\label{cor1}
    Suppose that the conditions in Theorem \ref{nonasymp_error} hold. Let $U=1$, $M=N^{d_0/[2(d_0+2\beta)]}$ and $\mathcal{B}=\log(N)$. Then the NQ net estimator $\hat{f}_N=(\hat{f}^{(1)}_N.\ldots,\hat{f}^{(K)}_N)$ defined in (\ref{erm}) satisfies the monotonicity constraint
$\hat{f}^{(1)}_N< \cdots<\hat{f}^{(K)}_N$ and
\begin{align*}
  \mathbb{E}[\mathcal{R}(\hat{f}_N)]&\le
C\cdot K^2(\log N)^7 N^{-\frac{2\beta}{d_0+2\beta}}, 
\end{align*}
where $C>0$ is a constant depending only on $\beta$ and $d_0$.
\end{corollary}

Corollary \ref{cor1} indicates that the excess risk of the NQ network estimator can achieve the minimax optimal rate (up to logarithms) for nonparametric regression \citep{stone1982optimal}.  It is important to note that the choice of $U=1$ and $M=N^\gamma$ for NQ network architecture is for parameter efficiency rather than as a strict requirement as network size scales faster with increasing width than with depth (\(\mathcal{S} \approx \mathcal{D} \mathcal{W}^2\)) \citep{jiao2023deep}. Various network architectures, including those with varying widths and depths, can achieve the optimal rate as long as the total number of parameters in the network scales properly with the sample size.  

\subsection{Curse of dimensionality}\label{sec_cod}


Recall that the excess risk bound in Corollary \ref{cor1} scales at a rate of \(O(N^{-\frac{2\beta}{2\beta+d_0}})\), adjusted by a logarithmic factor of \(N\). In scenarios common to many machine learning tasks, where the input dimension \(d_0\) is large, the convergence rate of the NQ network estimator slows markedly. This is a manifestation of what is commonly referred to as the \emph{curse of dimensionality}. Such a slow convergence rate implies the need for a significantly larger sample size to achieve desired theoretical accuracy, often proving impractical in real-world settings.

In modern statistics and machine learning, many high-dimensional data tend to lie in the vicinity of a low-dimensional manifold \citep{pope2021intrinsic}. This fact provides a way to mitigate the curse of dimensionality in the sense that the estimator maintains a desirable convergence rate even with a very large $d_0$. As in manifold learning and semi-supervised learning \citep{fefferman2016testing}, we posit the assumption of approximate low-dimensional manifold support below.

\begin{assumption}
	\label{low-dim}
	The covariate $X$ is supported on $\mathcal{M}_\rho$, a $\rho$-neighborhood of $\mathcal{M}\subset[0,1]^{d_0}$, where $\mathcal{M}$ is a compact $d_\mathcal{M}$-dimensional Riemannian sub-manifold and
	$$\mathcal{M}_\rho=\{x\in[0,1]^{d_0}: \inf\{\Vert x-y\Vert_2: y\in\mathcal{M}\}\leq \rho\}, \ \rho \in (0, 1).$$
\end{assumption}

With an approximate low-dimensional manifold support, the approximation for ReLU networks is shown improved for nonparametric least squares regressions \citep{chen2019nonparametric,jiao2023deep}. Here we show the assumption also results in a faster rate for the distributional estimation using NQ networks. 

\begin{lemma}[Approximation Error bound]\label{app_error_lowdim}
    Suppose that Assumptions \ref{assump1}, \ref{assump2},  and \ref{low-dim} hold. Let the excess risk $\mathcal{R}$ be defined as in (\ref{excess_risk}) and $\mathcal{F}_N$ be defined as in Section \ref{sec_training} with depth $\mathcal{D}$ and width $\mathcal{W}$. Given $\delta\in(0,1)$, let  the effective dimension $d_0^*=O(d_\mathcal{M}{\log(d_0/\delta)}/{\delta^2})$ be an integer such that $d_\mathcal{M}\leq d_0^*<d_0$. For any positive integers $U$ and $M$, let $\mathcal{W}=38(\lfloor\beta\rfloor+1)^2 (K+1)(d_0^*)^{\lfloor\beta\rfloor+1}U\log_2(8U)$ and depth $\mathcal{D}=21(\lfloor\beta\rfloor+1)^2 (d_0^*)^{\lfloor\beta\rfloor+1}M\log_2(8M)$ for neural networks in $\mathcal{F}_N$, then we have 
    {\small
\begin{align*}
    \inf_{f\in\mathcal{F}_N}[\mathcal{R}(f)]&\le (18+C_2)^2(K+2)^2\mathcal{B}^2(\lfloor\beta\rfloor+1)^4(d_0^*)^{2\lfloor\beta\rfloor+(\beta\vee 1)}(UM)^{-4\beta/(d_0^*)}+(K+2)^2\exp(-2\mathcal{B}),
\end{align*}
}
when $\rho\leq C_2(UM)^{-2\beta/d_0^*}(\beta+1)^2(d_0^*)^{3\beta/2}\{(1-\delta)^{-1}\sqrt{{d_0}/{d_0^*}}+1\}^{-1}$ for some universal $C_2>0$.
\end{lemma}

    With Assumption \ref{low-dim}, the approximation error is improved in the exponent with the effective dimension $d_0^*$, which can be significantly smaller than the ambient dimension $d_0$ especially when the input dimension of data is high. 
    If the data is assumed to have a low-dimensional structure, the rate of convergence derived in Corollary \ref{cor1} can also be improved. 

\begin{corollary}\label{cor2}
    Suppose the conditions in Theorem \ref{nonasymp_error} and Lemma \ref{app_error_lowdim} hold. Let $U=1$ and $M=N^{d_0^*/[2(d_0^*+2\beta)]}$.Then the NQ net estimator $\hat{f}_N=(\hat{f}^{(1)}_N.\ldots,\hat{f}^{(K)}_N)$ defined in (\ref{erm}) satisfies the monotonicity constraint
$\hat{f}^{(1)}_N< \cdots<\hat{f}^{(K)}_N$ and
\begin{align*}
  \mathbb{E}[\mathcal{R}(\hat{f}_N)]&\le
C\cdot K^2(\log N)^7 N^{-\frac{2\beta}{d_0^*+2\beta}}, 
\end{align*}
where $C>0$ is a constant depending only on $\beta$, $d_0$ and $d_0^*$.
\end{corollary}

\section{Applications to Reinforcement Learning (RL)}\label{example}

The NQ network architecture is highly adaptable and can be applied effectively across a wide spectrum of distributional learning tasks including DRL  \citep{bellemare2017distributional}.

 Under the RL framework \citep{sutton2018reinforcement}, agents learn through sequential decision-making by interacting with an environment modeled as a Markov decision process \citep[MDP,][]{puterman1994markov}, characterized by  $(\mathcal{S}, \mathcal{A}, R, P, \gamma)$, where $\mathcal{S}$ and $\mathcal{A}$ are the state and action spaces, respectively, $R:\mathcal{S}\times\mathcal{A}\to\mathbb{R}$  
denotes the reward function, 
$P(s' \mid s, a)$ denotes the probability mass/density function of transitioning from one state $s$ to another $s'$ after taking action $a$, and $\gamma\in[0,1)$ is the discount factor. At each time $t$, 
the agent interacts with the environment by observing a state $S_t$, taking an action $A_t$, and receiving a reward $R(S_t,A_t)$ based on the action. The objective of RL is to learn an optimal policy $\pi^*$ that maximizes the expected discounted cumulative reward.
Mathematically, 
a policy $\pi(\cdot\mid s)$ maps each state $s\in\mathcal{S}$ to a distribution over the action space $\mathcal{A}$. For a fixed policy $\pi$ and a given initial state-action pair $(s,a)$, its return 
$$Z^\pi(s,a)=\sum_{t=0}^\infty\gamma^t R(S_t,A_t),$$
is a random variable representing the sum of discounted rewards observed along a trajectory $\{(S_t,A_t)\}_{t\ge 0}$ following $\pi$, conditional on that $S_0=s$ and $A_0=a$. The optimal policy can thus be defined as $\pi^*=\argmax_{\pi}J(\pi),$ where $J(\pi)=\mathbb{E} [\pi(a\mid S_0)Z^{\pi}(S_0,a)]$ measures the expected cumulative reward following $\pi$.


\subsection{Deep NQ network for DRL}

In DRL, the goal is to explicitly model the distribution of returns instead of its mean value. \cite{bellemare2017distributional} introduced the \textit{distributional} Bellman equation, which characterizes the entire distribution of the random return rather than its mean. This approach builds on previous work in risk-sensitive reinforcement learning 
\citep{heger1994consideration,morimura2010parametric,chow2015risk}. 
To begin with, we define the distributional Bellman operator $\mathcal{T}^\pi$, given by
\begin{align*}
(\mathcal{T}^\pi Z)(s,a) = R(s,a) + \gamma Z(s', a')\qquad{\rm for}\ s'\sim P(\cdot\mid s,a) \quad{\rm and}\quad a'\sim \pi(\cdot\mid s'),
\end{align*}
where 
$Z: \mathcal{S} \times \mathcal{A} \rightarrow \mathcal{D}(\mathbb{R})$ denotes a random function that maps a given state-action pair into a random variable in $\mathbb{R}$. 
The distributional Bellman equation is then given by
\begin{equation*}
    (\mathcal{T}^\pi Z^{\pi})(s,a) \stackrel{d}{=}Z^{\pi}(s,a),
\end{equation*}
where $ U\stackrel{d}{=} V$ denotes equality in probability. Using the Wasserstein metric as the distance measure, \cite{bellemare2017distributional} proved that the distributional Bellman operator is a contraction. By the fixed point theorem, $Z^\pi$ corresponds to the fixed point of Bellman operator $\mathcal{T}^\pi$ and can be computed by iteratively setting $Z \stackrel{d}{=}\mathcal{T}^\pi Z$ until convergence. This distributional Bellman operator $\mathcal{T}^\pi$ is defined with respect to a given target policy $\pi$. 
Similarly, for the optimal policy, its return satisfies the distributional Bellman optimality equation $\mathcal{T} Z^{\pi^*}\stackrel{d}{=}Z^{\pi^*}$ where the operator $\mathcal{T}$ is defined as
\begin{align}\label{opt-bellman}(\mathcal{T} Z)(s,a) {=} R(s,a) + \gamma Z(s', a'), ~~{\rm where}\ s'\sim P(\cdot\mid s,a)~{\rm and}~ a'=\argmax_{a\in\mathcal{A}} \mathbb{E} [Z(s',a)]. \end{align}
Here, the action used for the next state is the greedy action with respect to the mean of the next state-action value distribution, as the optimal policy is defined to maximize the expected return.


Quantile regression provides a compelling approach to implement DRL by estimating the quantiles of the random return \( Z^{\pi^*} \) to represent the entire distribution. A key observation is that aggregating all quantiles yields the expected return, which is critical for determining the optimal policy, as shown in the above equation. 
Motivated by this, we propose to use our non-crossing deep quantile regression method to simultaneously estimate the quantiles of \( Z^{\pi^*} \) at $K$ different quantile levels with a sufficiently large \( K \), and average them to compute the optimal policy. Algorithm \ref{alg-1} provides a comprehensive summary of this approach.


\begin{algorithm}[t]
\caption{Distributional RL with fitted NQ Iterations}\label{alg-1}
\begin{algorithmic}
\singlespacing\Require Initial quantile estimator $Z^{(0)}=(Z^{(0)}_1,\ldots,Z^{(0)}_{K})$ where each $\mathcal{Z}^{(0)}_k$ is a quantile function dependent upon the state-action pair and belongs to $\mathcal{F}^{(RL)}_N$. 
\For{iteration $m=0$ to $M-1$}
    \State Compute the expected return $Q^{(m)}=K^{-1}\sum_k Z_k^{(m)}$. 
    \State Compute the optimal policy $\pi_m$ as the greedy policy with respect to $Q^{(m)}$.
    \State Combine $\pi_m$ with certain exploration strategy (e.g., $\epsilon$-greedy) to generate a sequence of state-action-reward-next-state tuples $\{(S^{(m)}_i,A^{(m)}_i,R^{(m)}_i,S_i^{\prime(m)})\}_{i\in[N]}$ of size $N$.
    \State Compute $(\mathcal{T}Z^{(m)}_k)_i=R^{(m)}_i+\gamma Z_k^{(m)}(S_i'^{(m)},a')$ where $a'=\argmax_{a\in\mathcal{A}} \sum_{k=1}^K Z^{(m)}_k(S_i'^{(m)},a)$  \qquad\qquad for $i=1,\ldots,N$ and $k=1,\ldots,K$
    \State Apply the proposed NQ network to update the quantile function estimator
    $$Z^{(m+1)}\leftarrow\arg\min_{Z\in\mathcal{F}}\frac{1}{N}\sum_{i=1}^N\sum_{k=1}^K\sum_{j=1}^K\rho_{\tau_k}\left((\mathcal{T}Z^{(m)}_j)_i-Z_k(S^{(m)}_i,A^{(m)}_i)\right),$$
\EndFor
\State Compute the optimal policy $\pi_M$ as the greedy policy with respect to $Q^{(M)}=K^{-1}\sum_k Z_k^{(M)}$.
\State \textbf{Output:} 
The estimated optimal policy $\pi_M$.
\end{algorithmic}
\end{algorithm}

As illustrated in Algorithm \ref{alg-1}, different from the deep distributional learning methods presented in Section \ref{methodology}, quantile regression needs to be iteratively conducted to solve the Bellman optimality equation in DRL. 
In particular, during the $m$th iteration, we update the NQ network using newly generated data $\{(S^{(m)}_i,A^{(m)}_i,R^{(m)}_i,S_i^{\prime(m)})\}_{i\in[N]}$. The updated network then serves as the prediction target in the next iteration, in line with the classical fitted Q-iteration algorithm \citep[FQI,][]{ernst2005tree}. 
 Additionally, to model the distribution of the state-action value $Z(s, a)$, we modify the original NQ networks $\mathcal{F}_N$ defined in (\ref{NQ-networks}) for the value distribution estimation as follows: 
\begin{align}\label{NQ-networks_RL}
    \mathcal{F}_N^{{(RL)}}=\{f:\mathcal{S}\times\mathcal{A}\to\mathbb{R}: f(\cdot,a)\in\mathcal{F}_N {\rm \ for\ any\ } a\in\mathcal{A}\}.
\end{align}
By this definition, $\mathcal{F}^{(RL)}_N$ is a class of functions (may depend on the sample size $N$) that take state and action as input and output the corresponding values at $K$ different quantiles. For any function $f\in\mathcal{F}_N^{{(RL)}}$ and any action $a\in\mathcal{A}$, the function $f(\cdot,a)\in\mathcal{F}_N$ is a standard NQ network in (\ref{NQ-networks}) defined on the state space $\mathcal{X}$ with depth $\mathcal{D}$, width $\mathcal{W}$, number of neurons $\mathcal{U}$, size $\mathcal{S}$ and bounded by $\mathcal{B}$. Since the action space is finite, we can parallel the NQ networks $f(\cdot, a)\in\mathcal{F}_N$ for all $a\in\mathcal{A}$ to form a large network $f(\cdot, a)\in\mathcal{F}_N$ with width $\vert\mathcal{A}\vert \mathcal{W}$ and  depth $\mathcal{D}$. 


%

To conclude this section, we discuss a closely related work by \citet{zhou2020non}, who developed a similar deep neural network architecture for DRL, named NC-QR-DQN, to address the crossing issue. In particular,  NC-QR-DQN employs a feedforward network to extract features from the input state and maps the features to two scalars $\alpha,\beta$ by a ``Scale Factor Network", and to a $K$-dimensional softmax vector $(\phi_0,\ldots,\phi_{K-1})$ by a ``Quantile logit Network". Then, each of the $k$-th quantile is modeled by $\beta+\alpha\times\sum_{t=0}^k\phi_t$ where the non-crossing property is guaranteed by the non-negativity of $\phi_t,t=0,\ldots, K-1$ (softmax) and that of $\alpha$ (ReLU). Essentially. Here $\beta$ models the lowest quantile, and $\{\alpha\times \phi_t: t=0,\ldots, K-1\}$ model the gaps between adjacent quantiles. However, as mentioned earlier, the estimated quantile gaps from NC-QR-DQR may be zero, leading to (partially) overleaped quantiles since the employed ReLU activation function is not strictly positive.

\subsection{Theoretical results}


In this section, we establish statistical guarantees for DRL  using the proposed deep NQ networks.  Similar to the analysis in Section \ref{sec_theory}, we set the state space $\mathcal{S}$ to $[0,1]^{d_0}$. 

Before introducing our technical assumptions, we first highlight the assumptions that we do not make. Unlike the existing literature which typically requires the data to be i.i.d. \citep[see e.g.,][]{chen2019information,fan2020theoretical,li2021tightening,uehara2021finite}, here we do not impose the i.i.d. assumption, as it is often violated in MDPs due to the temporal dependence between the observations \citep{hao2021bootstrapping}. Nor do we require certain stationarity, ergodicity, or mixing conditions. These conditions are again frequently imposed \citep[see e.g.,][]{antos2007value,
shi2022statistical,ramprasad2023online,zhou2024estimating}, but are likely violated in episodic tasks with a finite termination time. We next present our imposed assumptions to establish our theories. 
\begin{assumption}\label{assump_RL0}
     There exist some $p>1$ and $0<C_{p,R}<\infty$ such that $(\mathbb{E}\vert R(s, a)\vert^p)^{1/p} < C_{p,R}$ for any $s\in\mathcal{S}$ and $a\in\mathcal{A}$.
\end{assumption}
Assumption \ref{assump_RL0} only requires the reward function to have bounded absolute moments of order $p>1$, no matter how close $p$ is to $1$. In contrast, existing works typically require the reward to be either bounded \citep[see e.g.,][]{chen2019information,fan2020theoretical,shi2022statistical,
ramprasad2023online} or sub-Gaussian \citep[see e.g.,][]{rowland2023statistical}. As commented in the introduction, such a relaxation characterizes the statistical benefits of DRL in the presence of heavy-tailed rewards, which have been empirically demonstrated in the DRL literature \citep{dabney2018distributional,rowland2023statistical}. Other studies have either discussed them heuristically \citep{rowland2023statistical} or employed similar conditions 
for policy evaluation \citep{xu2022quantile,zhu2024robust}.




\begin{assumption}\label{assump_RL1}
    For any $f\in\mathcal{F}^{(RL)}_N$ and any $a, a'\in\mathcal{A}$, the function $R_{\tau}(\cdot, a)+ \gamma f(\cdot, a')$ is $\beta$-H\"older smooth with constant $B$, where $R_{\tau}(s, a)$ denotes the $\tau$-th conditional quantile of the reward given the state $s$ and action $a$. 
\end{assumption}

Assumption \ref{assump_RL1} ensures that the target function in each iteration can be approximated by networks. It is closely related to the completeness assumption that is commonly assumed in the RL literature 
\citep{chen2019information,fan2020theoretical,uehara2021finite}. It is automatically satisfied when  $R_\tau$ and $f$ are both sufficiently smooth. 

\begin{assumption}\label{assump_RL2}
	There exist constants $c,C >0$ such that for any $\vert \delta\vert\le C$ and $m=0,\ldots,M-1$,
	$$\left\vert P_{m}\left(\tilde{Z}^{(m)}_\tau(s+\delta,a)\bigg|s,a\right)-P_{m}\left(\tilde{Z}^{(m)}(s,a)\bigg|s,a\right)\right\vert\ge c\vert \delta\vert,$$
    for all $\tau\in(0,1)$, $s\in\mathcal{S}$, and $a\in\mathcal{A}$ up to a negligible set. Moreover, $ P_{m}(\cdot|s,a)$ denotes the conditional distribution function of $\tilde{Z}^{(m)}:=\mathcal{T}Z^{(m)}$ given state $s$ and action $a$ and $\tilde{Z}^{(m)}_\tau(s',a')$ denotes the $\tau$th conditional quantile of $\tilde{Z}^{(m)}$ given state $s'$ and action $a'$.
\end{assumption}

Similar to the calibration condition in Assumption A1 in appendix, we posit the Assumption \ref{assump_RL2} which links the learning risk of each iteration in Algorithm \ref{alg-1} with the convergence of quantiles estimations. 

\begin{assumption}\label{assump_RL3}
    Given the sample size $N$,  let the NQ networks in $\mathcal{F}_N^{{(RL)}}$ have depth $\mathcal{D}=21(\lfloor\beta\rfloor+1)^2 (d_0)^{\lfloor\beta\rfloor+1}N^{d_0/[d_0+4\beta]}\log_2(8N^{d_0/[d_0+4\beta]})$, width $\mathcal{W}=114(\lfloor\beta\rfloor+1)^2 (K+1)(d_0)^{\lfloor\beta\rfloor+1}$, and bound $\mathcal{B}=\log(N)$. 
\end{assumption}
Given the sample size $N$, we set the proper NQ network architecture with respect to $N$ to achieve a fast learning rate of the quantiles. Then,  we are ready to obtain our main theorem which guarantees the learning efficiency of the fitted NQ in Algorithm \ref{alg-1}.

\begin{theorem}\label{theorem_rl}
Suppose that Assumptions \ref{assump_RL0}, \ref{assump_RL1},  \ref{assump_RL2},  and \ref{assump_RL3} hold. 
Let $\pi_M$ denote the greedy policy of $Z^{(M)}$. Then the expected cumulative reward following $\pi_M$ satisfies
\begin{align}\label{theorem_RL} 
    J(\pi^*)-J(\pi_M)
  \le \frac{2c_{M}\gamma}{(1-\gamma)^2} \vert \mathcal{A}\vert (\log N)^4 N^{-2\beta/(4\beta+d_0)} +\frac{8\gamma^{M+1}}{(1-\gamma)^2} C_{p,R}+\frac{C_1\times C_{p,R}}{(1-\gamma)K^{(p-1)/p}},
\end{align}
where $C_1>0$ is a universal constant and $c_{M}>0$ is the concentration coefficient \citep{antos2007value,chen2019information,fan2020theoretical} depending on the distribution of data (see Definition S3 in the appendix).
\end{theorem}

By Theorem \ref{theorem_rl}, the statistical convergence rate for the regret learned by the NQ network is the sum of an estimation error, an algorithmic error, and an approximation error.  The approximation error occurs since we use the average of $K$ quantiles to approximate the d mean. The algorithmic error converges to zero at a linear rate with respect to the number of fitted iterations $M$, whereas the estimation error intrinsically relates to distributional learning with quantile regression using the NQ network. When the number of iterations $M\ge C[\log\vert \mathcal{A}\vert^{-1}+(2\beta/(4\beta+d_0))\log(N)]$ for some large constant $C>0$, the estimation error will dominate the algorithmic error. By ignoring the algorithmic error, we see that the prediction error for the expected action-value converges at the rate $\vert\mathcal{A}\vert N^{-2\beta/(4\beta+d_0)}$ (up to logarithm of $N$), which scales linearly in the cardinality of the action space and approach to zero in a nonparametric rate with respect to the sample size $N$.

It is worth noting that Theorem \ref{theorem_rl} is a novel learning guarantee for DRL in several aspects. To the best of our knowledge, Theorem \ref{theorem_rl} is the first result in the RL literature that establishes the statistical properties of DRL using deep learning without imposing the restrictive i.i.d., stationary or mixing assumptions on the training data. Nor do we require the reward to be bounded, sub-Gaussian or to possess high-order moments. In addition, the estimation error in Theorem \ref{theorem_rl} can be further improved if the distribution of state-action pair data $(S, A)$ has a low-dimensional structure as in Section \ref{sec_cod}.

\section{Numerical Study}\label{sec_numerical}

In this section, we implement the NQ network on non-parametric quantile regression and compare it with popular deep quantile regression methods.  
We implement five popular deep quantile regression methods as follows: \\ (I) Deep quantile regression \citep{padilla2022quantile,shen2021deep}, denoted by \textit{DQR}. As a benchmark for quantile regression methods. It estimates the conditional quantile curves using ReLU neural networks without non-crossing constraints. \\ (II) Deep quantile regression with non-crossing constraints \citep{padilla2022quantile}, denoted by \textit{DQR*}. It is an extension of the \textit{DQR} to multiple quantile estimation with non-crossing constraints. Given a finite set of quantiles, \textit{DQR*} estimates the conditional quantile curve at the smallest level and the gaps between quantile curves by a non-negative transformation $\log(1+\exp(\cdot))$ on the output of ReLU networks to ensure the monotonicity. \\ (III) Deep quantile regression process using Rectified Quadratic Unit (ReQU) networks \citep{shen2024nonparametric}, denoted by \textit{DQRP}, which nonparametrically estimates the quantile process with a non-crossing penalty. The \textit{DQRP} penalizes the non-positivity of the partial derivative of the process estimator with respect to the quantile level to encourage monotonicity. We implement \textit{DQRP} with tuning parameter $\lambda=\log(n)$ and uniformly distributed $\xi\sim Unif(0,1)$ according to \cite{shen2024nonparametric}. \\ (IV) Non-crossing Quantile Regression Deep-Q-Network \citep{zhou2020non}, denoted by \textit{NC-QR-DQN}, which is proposed for Distributional RL by using non-crossing quantile regression based on the QR-DQN architecture.  NC-QR-DQN employs a feedforward network to extract features, then maps them to two scalars (a slope and an intercept) by a ``Scale Factor Network", and maps to a softmax vector by a ``Quantile logit Network". The quantiles are modeled by the linear transformations of cumulative sums of the softmax values with the two scalars as slope and intercept. The non-crossing property is guaranteed by the non-negativity of the ReLU-activated slope. The intercept models the smallest quantile and scaled cumulative sums of the softmax values model the gaps between adjacent quantiles.\\ (V) Non-crossing quantile network in this paper, denoted by \textit{NQ-Net}. Given a set of quantiles, \textit{NQ-Net} estimates the mean of the conditional quantile curves and estimates the gaps between quantile curves by a non-negative activation $ELU+1$ to ensure the estimated quantile curves are non-crossing. We set the ``Mean Net" and ``Gaps Net" to have the same architecture, and they are paralleled to form the NQ network.

 For a fair comparison, we implement all these methods using rectangle neural networks with the same width and depth. In particular, we use networks with 3 hidden layers and a width of $[128,128,128]$ for estimating univariate target functions, and 3 hidden layers and a width of $[256,256,256]$ for multivariate target functions. We implement all these methods via \textit{Pytorch} and use \textit{Adam} 
 optimization algorithm with default hyperparameters and learning rate 0.001.
 
\subsection{Training and Testing}

We generate the training data $(X_i^{train},Y_i^{train})_{i=1}^N$ with sample size $N$ from the joint distribution $(X,Y)$. We also generate the validation data with size $N/4$ for the early stopping of the training. We set the batch size by 128 and the maximum training epochs by 1,000 during the training for all methods. For each simulated model, we apply \textit{DQR}, \textit{DQR*} and \textit{NQ-Net} to estimate the quantile curves at 19 different levels $(\tau_1,\ldots,\tau_{19})=(0.05,0.1,\ldots,0.9,0.95)$, and we apply \textit{DQRP} to estimate the quantile process for $\tau\in(0,1)$. 

For each simulated model, we generate testing data \((X_t^{test}, Y_t^{test})_{t=1}^T\) with a sample size of \(T = 10^5\), using the data distribution \((X, Y)\). The ground truth for the \(\tau\)th conditional quantile function of \((Y \mid X)\) is denoted by \(Q_Y^{\tau}(\cdot)\). 
Estimates obtained from various methods are represented as \(\hat{f}_N^{\tau_k}\) for the estimated conditional quantile at level \(\tau_k\) for \(k = 1, \ldots, 19\). To assess performance, we compute the \(L_1\) and \(L_2\) distances between the estimates \(\hat{f}_N^\tau\) and the ground truth \(Q_Y^\tau(\cdot)\) at each quantile level \(\tau \in (0.05, 0.1, \ldots, 0.9, 0.95)\).
We report the mean and standard deviation of the $L_1$  and $L_2$ distances over $R = 100$ replications.


\subsection{Univariate models}

In our simulations, we generate \(X\) uniformly from the interval \([0,1]\). We explore the following settings for the function \(Q_Y^\tau\), which are used to compute the \(\tau\)-th conditional quantile of the response \(Y\) given \(X = x\):
(i) \textbf{Linear Model}: 
    $
    Q_Y^\tau(x) = 2x + F_t^{-1}(\tau),
    $
    where \(F_t(\cdot)\) is the cumulative distribution function (CDF) of the Student's t-distribution with 2 degrees of freedom.
(ii) \textbf{Wave Model}:
    $
    Q_Y^\tau(x) = 2x \sin(4\pi x) + \exp(4x - 2) \Phi^{-1}(\tau),
    $
    with \(\Phi(\cdot)\) representing the CDF of the standard normal distribution.
(iii) \textbf{Angle Model}:
    $
    Q_Y^\tau(x) = 4(1 - |x - 0.5|) + |\sin(\pi x)| \Phi^{-1}(\tau).
    $
These models allow us to evaluate the quantile estimation across various functional relationships between \(X\) and \(Y\).

\begin{figure}[t]
	\centering
	\includegraphics[width=\textwidth]{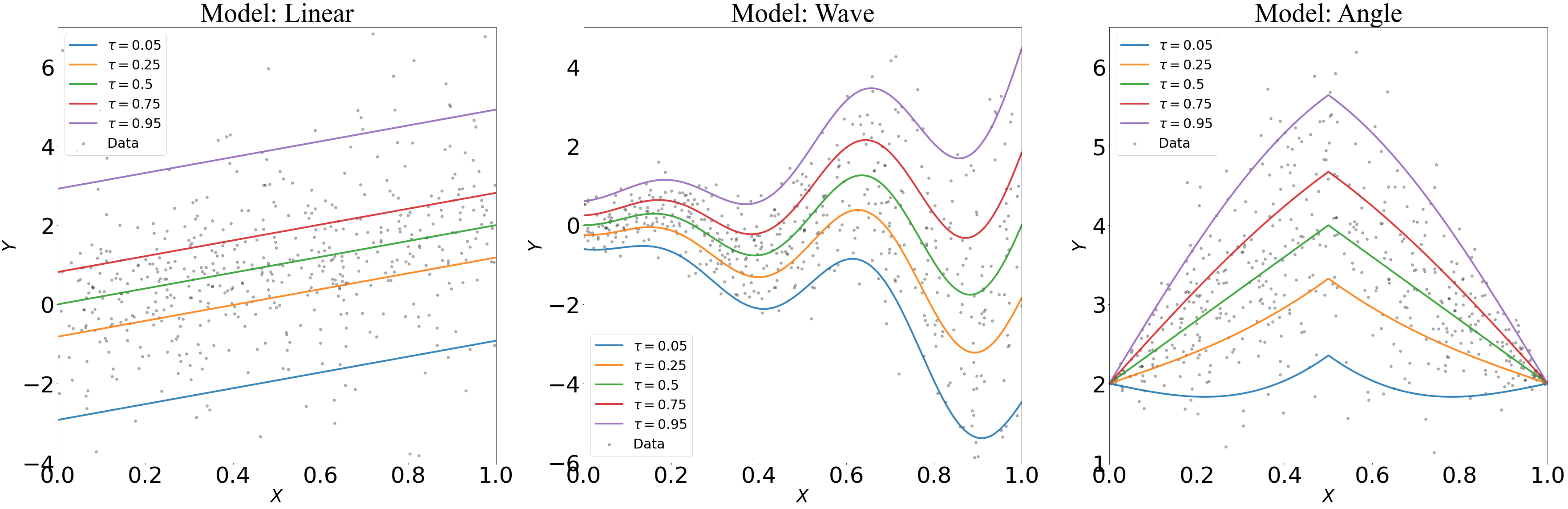}
	\caption{The simulated univariate models. The sample data with size $N=512$ is depicted as grey dots. Five conditional quantile curves at levels $\tau=$0.05 (blue), 0.25 (orange), 0.5 (green), 0.75 (red), and 0.95 (purple) are depicted as solid curves.}
	\label{fig:univariate_models}
\end{figure}

\begin{figure}[t]
	\centering\includegraphics[width=\textwidth]{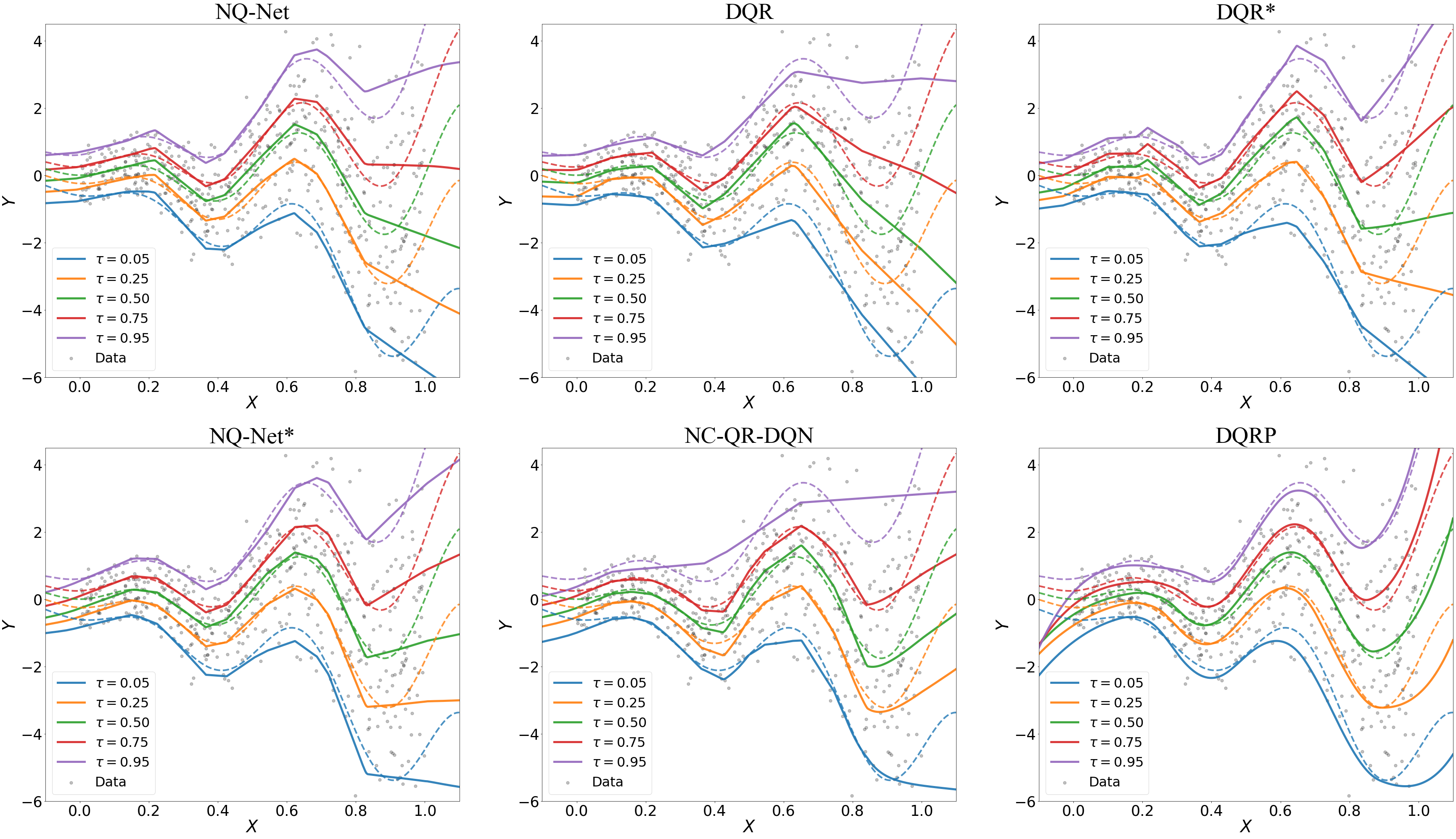}
	\caption{An instance of the fitted quantile curves under the ``Wave" model when $N=512$. The training data is depicted as grey dots. The target (estimated) quantile curves are depicted as dashed (solid) curves at levels $\tau=$0.05 (blue), 0.25 (orange), 0.5 (green), 0.75 (red), 0.95 (purple).}
	\label{fig:wave_estimates}
\end{figure}

The ``Linear" model serves as a baseline, and we expect all methods to perform well under this setting. The ``Wave" model introduces a nonlinear and smooth dynamic, while the ``Angle" model presents a nonlinear, continuous, yet non-differentiable challenge. Both the ``Wave" and ``Angle" models exhibit heteroscedasticity. Figure \ref{fig:univariate_models}  displays these univariate data generation models along with their corresponding conditional quantile curves at levels $\tau=0.05, 0.25, 0.50, 0.75, 0.95$. An example of estimated quantiles at these levels under the ``Wave" model is shown in Figure \ref{fig:wave_estimates}. The \textit{NQ-Net*} is a variant of the \textit{NQ-Net}, employing $ReLU$ activation instead of $ELU+1$.

In Tables \ref{tab:linear_512} and \ref{tab:wave_512}, we present the simulation results for ``Linear" and ``Wave" models with sample size $N=512$, which collect the $L_1$ and $L_2^2$ prediction errors for NQ-net, DQR, DQR*, NC-QR-DQN and DQRP at 19 quantile levels. The NQ network in general performs the best among deep non-crossing quantile estimations including DQR*, NC-QR-DQN, and DQRP, and it outperforms the benchmark method DQR on some occasions. Additional simulation results with sample size $N=2048$ and the ``Angle" model are presented in appendix.

 \begin{table}[H]
\centering
\caption{\footnotesize Summary statistics for the ``Linear" model with training sample size $N= 512$ and replications $R = 100$. The averaged $L_1$ and $L_2^2$ test errors with the corresponding standard deviation (in parentheses) are reported for the estimators trained by different methods.}
\label{tab:linear_512}
\resizebox{\textwidth}{!}{%
\begin{tabular}{c|ccccc|ccccc}
\hline
\multirow{2}{*}{$\tau$} & \multicolumn{5}{c|}{$L_1$}                                                                 & \multicolumn{5}{c}{$L_2^2$}                                                                \\
                        & NQ-Net                & DQR                   & DQR*         & NC-QR-DQN    & DQRP         & NQ-Net                & DQR                   & DQR*         & NC-QR-DQN    & DQRP         \\ \hline
0.05                    & \textbf{0.296(0.175)} & 0.363(0.166)          & 0.604(0.205) & 0.506(0.239) & 0.143(0.206) & \textbf{0.143(0.206)} & 0.216(0.234)          & 0.492(0.264) & 0.423(0.439) & 0.497(0.193) \\
0.1                     & \textbf{0.171(0.082)} & 0.194(0.094)          & 0.251(0.081) & 0.318(0.215) & 0.050(0.059) & \textbf{0.050(0.059)} & 0.067(0.069)          & 0.099(0.061) & 0.205(0.286) & 0.495(0.187) \\
0.15                    & \textbf{0.136(0.060)} & 0.142(0.066)          & 0.230(0.085) & 0.222(0.145) & 0.031(0.030) & \textbf{0.031(0.030)} & 0.035(0.036)          & 0.087(0.059) & 0.098(0.133) & 0.473(0.168) \\
0.2                     & 0.117(0.049)          & \textbf{0.115(0.050)} & 0.192(0.063) & 0.162(0.091) & 0.023(0.021) & \textbf{0.023(0.021)} & 0.023(0.025)          & 0.060(0.035) & 0.049(0.059) & 0.452(0.156) \\
0.25                    & 0.105(0.044)          & \textbf{0.104(0.045)} & 0.158(0.054) & 0.131(0.062) & 0.019(0.016) & \textbf{0.019(0.016)} & 0.019(0.016)          & 0.040(0.024) & 0.030(0.028) & 0.436(0.15)  \\
0.3                     & 0.099(0.041)          & \textbf{0.095(0.041)} & 0.135(0.048) & 0.116(0.048) & 0.017(0.015) & 0.017(0.015)          & \textbf{0.016(0.014)} & 0.029(0.018) & 0.023(0.018) & 0.426(0.147) \\
0.35                    & 0.093(0.039)          & \textbf{0.087(0.038)} & 0.114(0.045) & 0.111(0.043) & 0.015(0.014) & 0.015(0.014)          & \textbf{0.013(0.011)} & 0.022(0.015) & 0.020(0.015) & 0.419(0.146) \\
0.4                     & 0.088(0.039)          & \textbf{0.082(0.039)} & 0.102(0.042) & 0.111(0.041) & 0.014(0.013) & 0.014(0.013)          & \textbf{0.012(0.012)} & 0.017(0.014) & 0.020(0.014) & 0.415(0.146) \\
0.45                    & 0.086(0.036)          & \textbf{0.078(0.034)} & 0.095(0.038) & 0.115(0.042) & 0.014(0.012) & 0.014(0.012)          & \textbf{0.011(0.010)} & 0.015(0.011) & 0.021(0.015) & 0.413(0.147) \\
0.5                     & 0.085(0.035)          & \textbf{0.078(0.036)} & 0.085(0.034) & 0.120(0.044) & 0.013(0.011) & 0.013(0.011)          & \textbf{0.011(0.011)} & 0.012(0.009) & 0.023(0.016) & 0.412(0.147) \\
0.55                    & 0.085(0.037)          & \textbf{0.082(0.038)} & 0.085(0.036) & 0.124(0.045) & 0.013(0.012) & 0.013(0.012)          & \textbf{0.012(0.012)} & 0.013(0.011) & 0.024(0.018) & 0.411(0.147) \\
0.6                     & 0.089(0.037)          & \textbf{0.086(0.037)} & 0.092(0.035) & 0.130(0.046) & 0.014(0.012) & 0.014(0.012)          & \textbf{0.013(0.012)} & 0.014(0.012) & 0.027(0.019) & 0.413(0.148) \\
0.65                    & 0.091(0.038)          & \textbf{0.089(0.041)} & 0.101(0.039) & 0.138(0.050) & 0.015(0.013) & 0.015(0.013)          & \textbf{0.014(0.014)} & 0.017(0.014) & 0.031(0.023) & 0.417(0.151) \\
0.7                     & 0.096(0.041)          & \textbf{0.094(0.045)} & 0.112(0.046) & 0.154(0.064) & 0.016(0.014) & \textbf{0.016(0.014)} & 0.016(0.015)          & 0.022(0.017) & 0.041(0.034) & 0.426(0.156) \\
0.75                    & 0.107(0.044)          & \textbf{0.106(0.049)} & 0.132(0.052) & 0.182(0.009) & 0.019(0.015) & \textbf{0.019(0.015)} & 0.020(0.017)          & 0.029(0.022) & 0.059(0.054) & 0.443(0.167) \\
0.8                     & 0.123(0.050)          & \textbf{0.120(0.054)} & 0.156(0.063) & 0.221(0.121) & 0.025(0.018) & 0.025(0.018)          & \textbf{0.024(0.020)} & 0.040(0.029) & 0.090(0.085) & 0.468(0.186) \\
0.85                    & 0.145(0.063)          & \textbf{0.143(0.068)} & 0.180(0.073) & 0.261(0.147) & 0.034(0.028) & \textbf{0.034(0.028)} & 0.035(0.033)          & 0.053(0.037) & 0.125(0.124) & 0.503(0.214) \\
0.9                     & \textbf{0.185(0.085)} & 0.200(0.093)          & 0.220(0.085) & 0.285(0.148) & 0.053(0.043) & \textbf{0.053(0.043)} & 0.063(0.057)          & 0.074(0.050) & 0.143(0.146) & 0.539(0.245) \\
0.95                    & \textbf{0.296(0.168)} & 0.346(0.165)          & 0.298(0.157) & 0.529(0.236) & 0.132(0.124) & \textbf{0.132(0.124)} & 0.189(0.181)          & 0.138(0.126) & 0.384(0.301) & 0.553(0.238) \\ \hline
\end{tabular}%
}
\end{table}

\begin{table}[H]
\centering
\caption{\footnotesize Summary statistics for the ``Wave" model with training sample size $N= 512$ and replications $R = 100$. The averaged $L_1$ and $L_2^2$ test errors with the corresponding standard deviation (in parentheses) are reported for the estimators trained by different methods.}
\label{tab:wave_512}
\resizebox{\textwidth}{!}{%
\begin{tabular}{c|ccccc|ccccc}
\hline
\multirow{2}{*}{$\tau$} & \multicolumn{5}{c|}{$L_1$}                                                        & \multicolumn{5}{c}{$L_2^2$}                                                       \\
                        & NQ-Net                & DQR          & DQR*         & NC-QR-DQN    & DQRP         & NQ-Net                & DQR          & DQR*         & NC-QR-DQN    & DQRP         \\ \hline
0.05                    & \textbf{0.221(0.064)} & 0.284(0.095) & 0.416(0.122) & 0.625(0.654) & 0.541(0.119) & \textbf{0.133(0.157)} & 0.191(0.139) & 0.321(0.209) & 1.127(1.874) & 0.497(0.193) \\
0.1                     & \textbf{0.202(0.050)} & 0.257(0.086) & 0.320(0.115) & 0.468(0.515) & 0.533(0.122) & \textbf{0.111(0.112)} & 0.163(0.119) & 0.227(0.178) & 0.676(1.151) & 0.495(0.187) \\
0.15                    & \textbf{0.190(0.046)} & 0.239(0.083) & 0.279(0.103) & 0.405(0.402) & 0.519(0.124) & \textbf{0.095(0.063)} & 0.146(0.112) & 0.183(0.144) & 0.481(0.748) & 0.473(0.168) \\
0.2                     & \textbf{0.186(0.046)} & 0.234(0.083) & 0.255(0.092) & 0.358(0.314) & 0.507(0.127) & \textbf{0.089(0.053)} & 0.140(0.105) & 0.154(0.121) & 0.354(0.494) & 0.452(0.156) \\
0.25                    & \textbf{0.182(0.044)} & 0.231(0.083) & 0.232(0.079) & 0.316(0.241) & 0.499(0.130) & \textbf{0.085(0.048)} & 0.137(0.102) & 0.128(0.095) & 0.266(0.324) & 0.436(0.15)  \\
0.3                     & \textbf{0.182(0.045)} & 0.227(0.083) & 0.218(0.072) & 0.274(0.175) & 0.493(0.132) & \textbf{0.084(0.047)} & 0.136(0.101) & 0.113(0.078) & 0.193(0.210) & 0.426(0.147) \\
0.35                    & \textbf{0.181(0.044)} & 0.226(0.082) & 0.206(0.070) & 0.236(0.122) & 0.490(0.133) & \textbf{0.085(0.046)} & 0.135(0.100) & 0.103(0.076) & 0.137(0.136) & 0.419(0.146) \\
0.4                     & \textbf{0.181(0.044)} & 0.226(0.082) & 0.202(0.067) & 0.207(0.077) & 0.488(0.134) & \textbf{0.084(0.047)} & 0.137(0.097) & 0.099(0.074) & 0.099(0.073) & 0.415(0.146) \\
0.45                    & \textbf{0.181(0.042)} & 0.226(0.081) & 0.198(0.063) & 0.195(0.055) & 0.487(0.135) & \textbf{0.084(0.048)} & 0.138(0.095) & 0.096(0.065) & 0.089(0.052) & 0.413(0.147) \\
0.5                     & \textbf{0.180(0.043)} & 0.228(0.081) & 0.194(0.063) & 0.191(0.049) & 0.487(0.134) & \textbf{0.083(0.048)} & 0.142(0.096) & 0.094(0.070) & 0.086(0.052) & 0.412(0.147) \\
0.55                    & \textbf{0.182(0.045)} & 0.228(0.081) & 0.192(0.063) & 0.197(0.054) & 0.488(0.134) & \textbf{0.086(0.051)} & 0.141(0.095) & 0.093(0.070) & 0.093(0.061) & 0.411(0.147) \\
0.6                     & \textbf{0.182(0.044)} & 0.228(0.082) & 0.195(0.060) & 0.213(0.084) & 0.489(0.133) & \textbf{0.086(0.050)} & 0.141(0.097) & 0.095(0.069) & 0.114(0.097) & 0.413(0.148) \\
0.65                    & \textbf{0.184(0.046)} & 0.231(0.081) & 0.195(0.062) & 0.240(0.132) & 0.492(0.133) & \textbf{0.088(0.052)} & 0.144(0.098) & 0.096(0.077) & 0.148(0.161) & 0.417(0.151) \\
0.7                     & \textbf{0.187(0.045)} & 0.235(0.081) & 0.200(0.063) & 0.275(0.190) & 0.496(0.133) & \textbf{0.091(0.054)} & 0.148(0.101) & 0.101(0.082) & 0.201(0.256) & 0.426(0.156) \\
0.75                    & \textbf{0.190(0.044)} & 0.238(0.081) & 0.203(0.060) & 0.313(0.258) & 0.503(0.135) & \textbf{0.094(0.055)} & 0.152(0.100) & 0.103(0.077) & 0.273(0.392) & 0.443(0.167) \\
0.8                     & \textbf{0.191(0.044)} & 0.244(0.081) & 0.210(0.067) & 0.357(0.332) & 0.515(0.139) & \textbf{0.097(0.057)} & 0.157(0.100) & 0.108(0.086) & 0.373(0.582) & 0.468(0.186) \\
0.85                    & \textbf{0.198(0.047)} & 0.256(0.084) & 0.214(0.070) & 0.414(0.418) & 0.530(0.146) & \textbf{0.105(0.061)} & 0.171(0.111) & 0.115(0.100) & 0.519(0.853) & 0.503(0.214) \\
0.9                     & \textbf{0.206(0.052)} & 0.270(0.084) & 0.230(0.083) & 0.498(0.520) & 0.546(0.153) & \textbf{0.117(0.070)} & 0.186(0.113) & 0.128(0.116) & 0.758(1.265) & 0.539(0.245) \\
0.95                    & \textbf{0.228(0.063)} & 0.301(0.084) & 0.256(0.096) & 0.799(0.608) & 0.553(0.141) & \textbf{0.150(0.093)} & 0.212(0.117) & 0.153(0.150) & 1.462(1.922) & 0.553(0.238) \\ \hline
\end{tabular}%
}
\end{table}

\subsection{Multivariate models}

We compare NQ-Net against various deep quantile regression methods using several multivariate models:
(i) \textbf{Multivariate Linear Model:}
    $
    Q_Y^\tau(x) = 2A^\top x + F_t^{-1}(\tau),
    $ 
    where \(A\) is defined below.
 (ii) \textbf{Single Index Model:}
    $
    Q_Y^\tau(x) = \exp(0.1 \times A^\top x) + |\sin(\pi B^\top x)| \Phi^{-1}(\tau),
    $ 
    with vectors \(A\) and \(B\) as specified below.
(iii) \textbf{Additive Model:}
    $
    Q_Y^\tau(x) = 3x_1 + 4(x_2 - 0.5)^2 + 2\sin(\pi x_3) - 5|x_4 - 0.5| + \exp\{0.1(B^\top x - 0.5)\} \Phi^{-1}(\tau).
    $ 
The vectors \(A\) and \(B\) are, respectively,  given by
$
A = (1.012, -0.965, -0.785, 1.336, 0, 0.378, 0.599, 1.292)^\top
$ and 
$
B = (1.002, 0, -0.497, 3.993, 0, 0, 0, 0)^\top.
$ 
Detailed simulation results for these multivariate models are provided in the appendix. In summary, NQ-Net generally outperforms other deep non-crossing quantile estimation methods, including DQR*, NC-QR-DQN, DQRP, and the benchmark DQR.

\subsection{Evaluation on Atari 2600}
In this section, we evaluate the empirical performance of the NQ network on six selected Atari game environments \citep{bellemare2013arcade}. In  Atari environments, reinforcement learning agents act like game players and the goal is to learn an optimal policy as a function of the snapshots of the game interface.  
We compare the performance of \textit{NQ-Net} against the state-of-the-art model \textit{NC-QR-DQN} \citep{zhou2020non}. We implement both models in PyTorch by utilizing the same image-embedding network architecture and downstream networks with similar scales. Specifically, we employ ReLU activation for the ``Gaps net" in our model, and denote the network by \textit{NQ-Net*}.  As in most game settings, the differences between quantiles are near zero, ReLU activation in \textit{NQ-Net*} may facilitate the optimization.

We set the number of quantiles, $K$, to 200 and train both models on a sample of 200 million frames. In line with modern reinforcement learning practices, we utilize a replay buffer and a double-Q network. Both models employ a linear $\epsilon$-greedy exploration strategy, starting with $\epsilon = 1$ at 0.5 million frames and gradually decreasing to $\epsilon = 0.01$ by 1 million frames. The learning rate for both models was set to $5 \times 10^{-5}$. For both models, we use the quantile Huber loss $ \rho^{\kappa}_\tau(u)$ with $\kappa=1$ which is a smoothed variation of the quantile loss in \eqref{eq: quantile_loss} where $\rho_{\tau}^{\kappa}(u) = u\left( \tau - {\bf 1} \{u<0\} \right){\mathcal{L}_{\kappa}(u)}/{\kappa}$ and $\mathcal{L}_{\kappa}(u)=I(|u| \leq \kappa)\cdot u^2/2 +I(|u|>\kappa)\cdot\kappa \left( |u| - \frac{1}{2} u \right)$.   \vspace*{-0.3cm} 
 \begin{figure}[H]
    \centering
\includegraphics[width=1\textwidth]{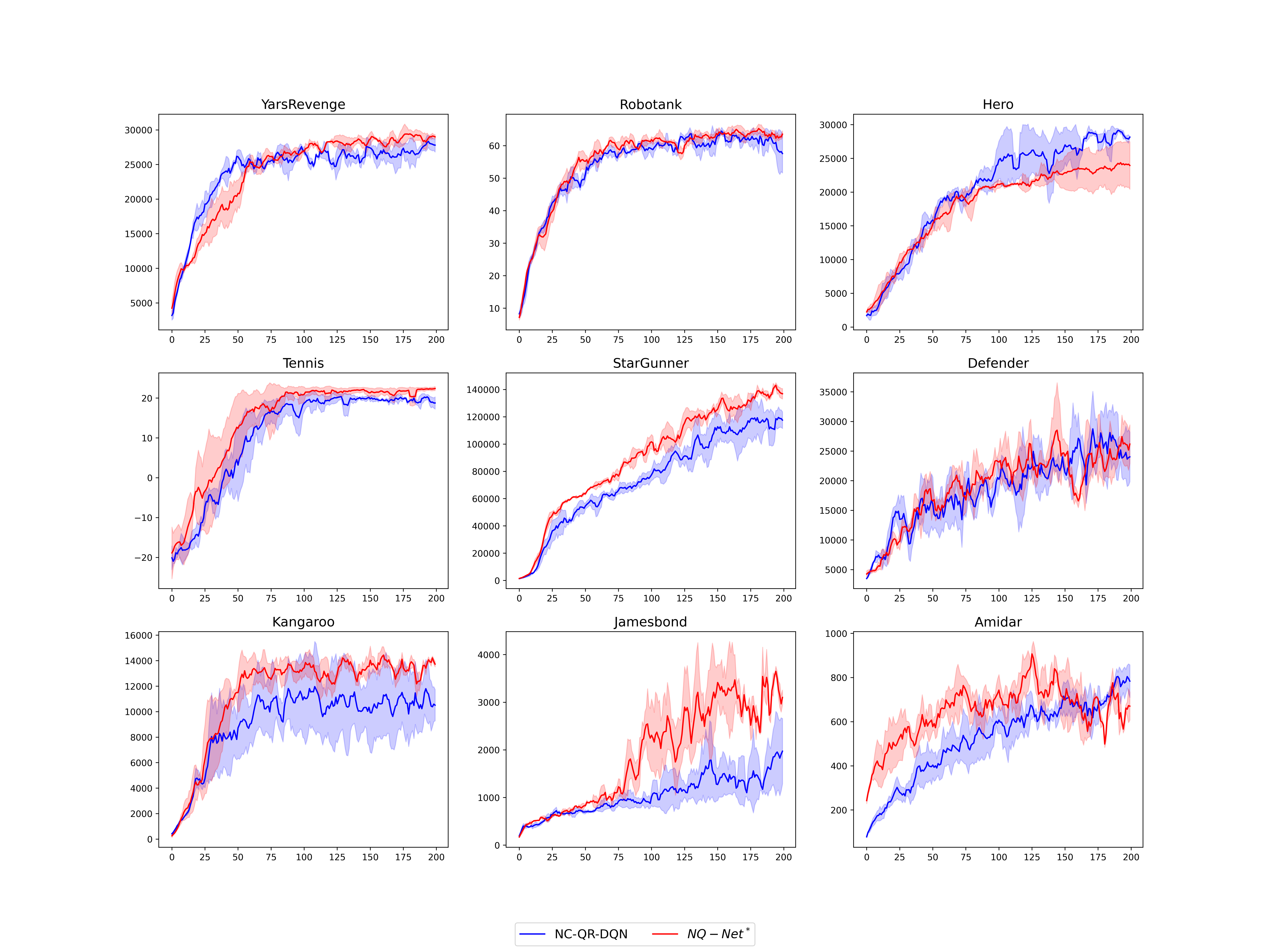}
    \caption{Comparison of testing scores for \textit{NQ-Net*} and \textit{NC-QR-DQN} along the training process.}
    \label{fig:game}
   \vskip -0.5cm 
\end{figure}

\begin{figure}[H]
    \centering
    \includegraphics[width=0.5\linewidth]{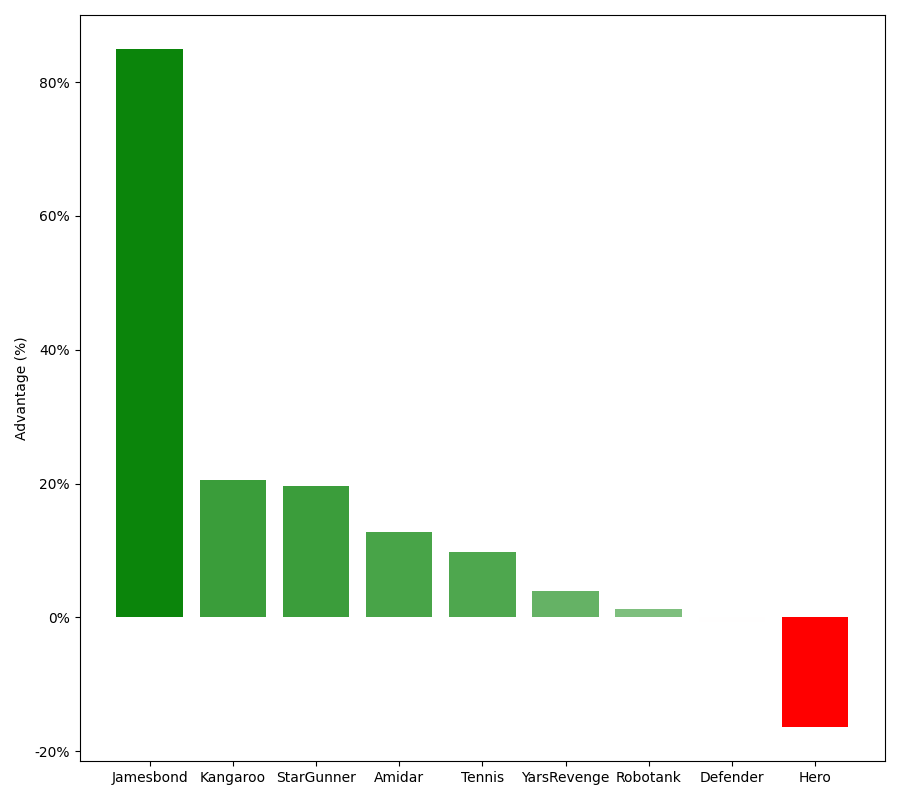}
    \caption{The advantage of the best-performing \textit{NQ-Net*} model over \textit{NC-QR-DQN}. The advantage (\%) on the y-axis is defined as $(\text{Score}_{NQ}-\text{Score}_{NC})/\text{Score}_{NC}$, where $\text{Score}_{NQ}$ and $\text{Score}_{NC}$ denote the highest testing scores achieved by the trained \textit{NQ-Net*} and \textit{NC-QR-DQN} models, respectively.} 
    \label{fig:max}
   \vskip -0.5cm 
\end{figure}

We conduct the experiments using NVIDIA V100 and A100 GPUs, with the memory usage of both models remaining under 2 gigabytes. The testing scores of \textit{NQ-Net*} and \textit{NC-QR-DQN} at various stages of training are shown in Figure \ref{fig:game}. The curves depict averages over 3 seeds and are smoothed using a window size of 5. The \textit{NQ-Net*} network outperforms \textit{NC-QR-DQN} in most environments and demonstrates competitive performance. This advantage is further highlighted when comparing the best-performing models, as illustrated in Figure \ref{fig:max}.

\section{Conclusion}\label{conclusion}

In this work, we propose a deep distributional learning framework utilizing the NQ network to ensure that the quantile estimators remain non-crossing. We establish upper bounds for the prediction error of these quantile estimators, which achieve the minimax optimal rate of convergence. Additionally, we demonstrate that the NQ network can adapt to the low-dimensional structure of the data, effectively mitigating the curse of dimensionality. We apply the NQ network to DRL, providing novel theoretical guarantees. Notably, our analysis accommodates heavy-tailed rewards and dependent data, relaxing the commonly imposed restrictive conditions in the reinforcement learning literature and making our approach more applicable to real-world scenarios. While the design of the NQ network is highly flexible, we do not explore all potential architectures, such as different activation functions or variations for Mean and Gaps nets. We leave this exploration for future work to develop more efficient networks. Furthermore, the adaptability of the NQ network to various distributional learning tasks motivates us to investigate its efficiency in causal inference and beyond.


\spacingset{1.6}
\bibliographystyle{apalike}
\bibliography{dnet}

\clearpage
{\noindent \bf \LARGE Appendix}
\appendix

\section{Implementations in Numerical Study}
We implement popular deep quantile regression methods including the vanilla deep quantile regression, deep quantile regression at multiple levels with non-crossing constraints, deep quantile regression process with non-crossing penalty, and our proposed NQ network. We summarize implementation details as follows.

\begin{itemize}
	\item Deep quantile regression \citep{padilla2022quantile,shen2021deep}, denoted by \textit{DQR}, which estimates the conditional quantile curves using ReLU neural networks without non-crossing constraints. We implement \textit{DQR} as a benchmark for quantile regression methods.

	\item  Deep quantile regression with non-crossing constraints \citep{padilla2022quantile}, denoted by \textit{DQR*}, which is an extension of the \textit{DQR} to multiple quantile estimation with non-crossing constraints. Given a finite set of quantiles, \textit{DQR*} estimates the conditional quantile curve at the smallest level and the gaps between quantile curves by a non-negative transformation $\log(1+\exp(\cdot))$ on the output of ReLU networks to ensure the monotonicity.
	
	\item  Deep quantile regression process using Rectified Quadratic Unit (ReQU) networks \citep{shen2024nonparametric}, denoted by \textit{DQRP}, which nonparametrically estimates the quantile process with a non-crossing penalty. The \textit{DQRP} penalizes the non-positivity of the partial derivative of the process estimator with respect to the quantile level to encourage monotonicity. We implement \textit{DQRP} with tuning parameter $\lambda=\log(n)$ and uniformly distributed $\xi\sim Unif(0,1)$ according to \cite{shen2024nonparametric}.
	
	\item  Non-crossing Quantile Regression Deep-Q-Network \citep{zhou2020non}, denoted by \textit{NC-QR-DQN}, which is proposed for Distributional RL by using non-crossing quantile regression based on the QR-DQN architecture.  NC-QR-DQN employs a feedforward network to extract features, then maps them to two scalars (a slope and an intercept) by a ``Scale Factor Network", and maps to a softmax vector by a ``Quantile logit Network". The quantiles are modeled by the linear transformations of cumulative sums of the softmax values with the two scalars as slope and intercept. The non-crossing property is guaranteed by the non-negativity of the ReLU-activated slope. The intercept models the smallest quantile and scaled cumulative sums of the softmax values model the gaps between adjacent quantiles.

	\item  Non-crossing quantile network in this paper, denoted by \textit{NQ-Net}. Given a set of quantiles, \textit{NQ-Net} estimates the mean of the conditional quantile curves and estimates the gaps between quantile curves by a non-negative activation $ELU+1$ to ensure the estimated quantile curves are non-crossing. We set the ``Mean Net" and ``Gaps Net" to have the same architecture, and they are paralleled to form the NQ network. We denote \textit{NQ-Net*} by a variant of the \textit{NQ-Net} employing $ReLU$ activation instead of $ELU+1$.
	
\end{itemize}

For a fair comparison, we implement all these methods using rectangle neural networks with the same width and depth. In particular, we use networks with 3 hidden layers and a width of $[128,128,128]$ for estimating univariate target functions, and 3 hidden layers and a width of $[256,256,256]$ for multivariate target functions. We implement all these methods in Python via \textit{Pytorch} and use \textit{Adam} \citep{kingma2014adam} optimization algorithm with learning rate 0.001 and default momentum parameters $\beta=(0.9,0.99)$.

\section{Additional results of Numerical Study}\label{appendix:simulation}

\begin{figure}[H]
	\centering
	\includegraphics[width=\textwidth]{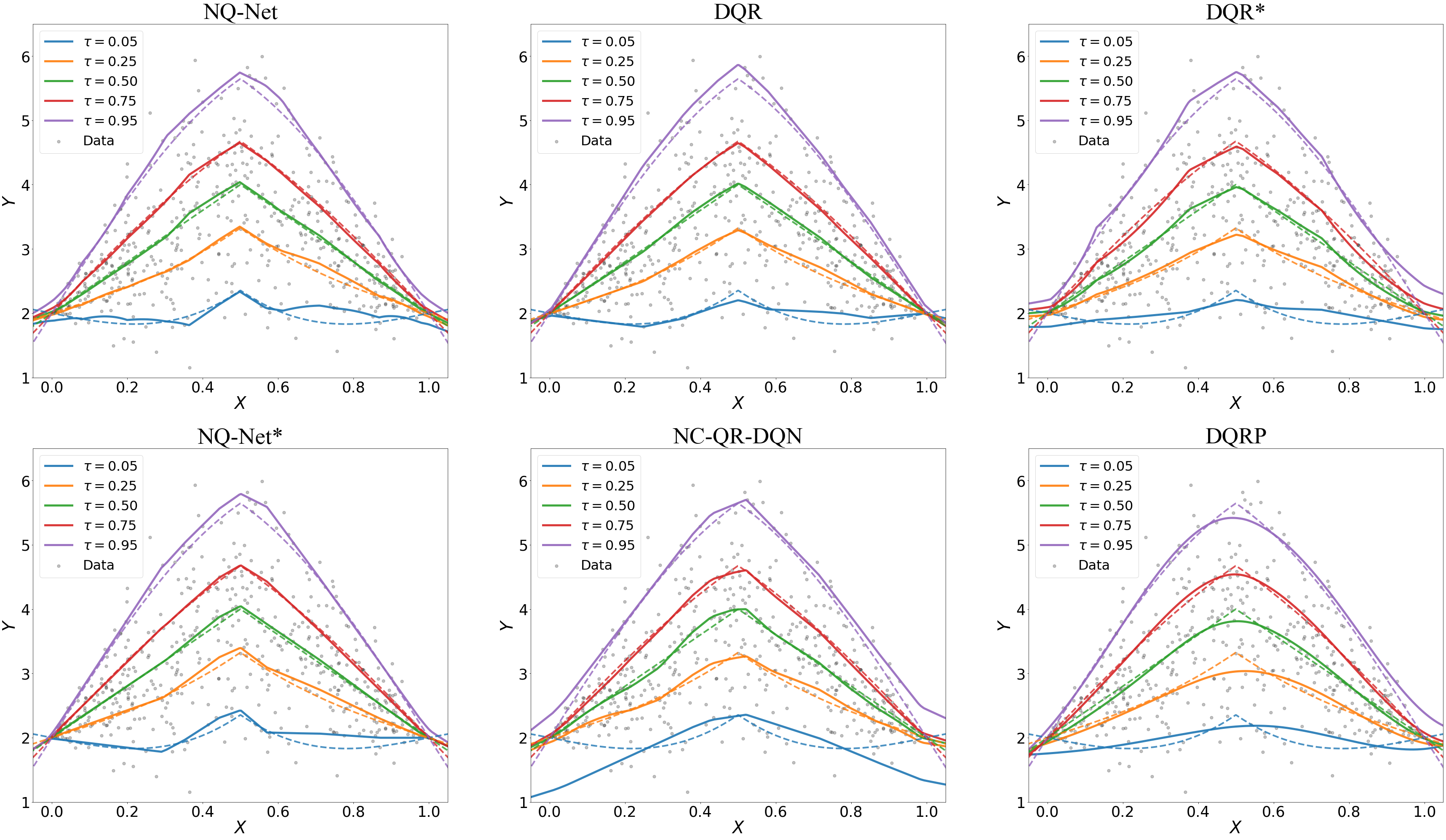}
	\caption{The fitted quantile curves under the ``Angle" model when $N=512$.
		The training data is depicted as grey dots. The target (estimated) quantile curves are depicted as dashed (solid) curves at levels $\tau=$0.05 (blue), 0.25 (orange), 0.5 (green), 0.75 (red), and 0.95 (purple).}
	\label{fig:angle_estimates}
\end{figure}

\begin{table}[H]
	\caption{\small 
		Summary statistics for the ``Angle" model with training sample size $N=512$ and the number of replications $R = 100$. The averaged $L_1$ and $L_2^2$ test errors with the corresponding standard deviation (in parentheses) are reported for the estimators trained by different methods.}
	\label{tab:angle_512}
	\setlength{\tabcolsep}{3pt}
	\renewcommand{\arraystretch}{1.2}
	\resizebox{\textwidth}{!}{%
		\begin{tabular}{c|ccccc|ccccc}
			\hline
			\multirow{2}{*}{$\tau$} & \multicolumn{5}{c|}{$L_1$}                                                        & \multicolumn{5}{c}{$L_2^2$}                                                       \\
			& NQ-Net       & DQR                   & DQR*         & NC-QR-DQN    & DQRP         & NQ-Net       & DQR                   & DQR*         & NC-QR-DQN    & DQRP         \\ \hline
			0.05                    & 0.131(0.036) & \textbf{0.109(0.059)} & 0.172(0.073) & 0.398(0.369) & 0.208(0.072) & 0.027(0.014) & \textbf{0.024(0.026)} & 0.049(0.044) & 0.365(0.562) & 0.062(0.040) \\
			0.1                     & 0.111(0.031) & \textbf{0.089(0.049)} & 0.118(0.022) & 0.269(0.311) & 0.133(0.034) & 0.020(0.011) & \textbf{0.016(0.018)} & 0.020(0.007) & 0.209(0.351) & 0.027(0.013) \\
			0.15                    & 0.097(0.029) & \textbf{0.080(0.043)} & 0.106(0.024) & 0.218(0.251) & 0.116(0.026) & 0.016(0.010) & \textbf{0.013(0.013)} & 0.017(0.008) & 0.137(0.231) & 0.022(0.010) \\
			0.2                     & 0.086(0.026) & \textbf{0.072(0.035)} & 0.092(0.022) & 0.186(0.199) & 0.113(0.026) & 0.013(0.008) & \textbf{0.010(0.010)} & 0.013(0.006) & 0.093(0.153) & 0.022(0.011) \\
			0.25                    & 0.078(0.024) & \textbf{0.066(0.030)} & 0.081(0.022) & 0.158(0.156) & 0.112(0.027) & 0.011(0.007) & \textbf{0.009(0.008)} & 0.011(0.006) & 0.063(0.099) & 0.022(0.011) \\
			0.3                     & 0.071(0.022) & \textbf{0.064(0.026)} & 0.074(0.021) & 0.133(0.117) & 0.109(0.028) & 0.009(0.006) & \textbf{0.008(0.007)} & 0.010(0.006) & 0.041(0.061) & 0.022(0.012) \\
			0.35                    & 0.065(0.021) & \textbf{0.059(0.024)} & 0.068(0.020) & 0.111(0.081) & 0.106(0.029) & 0.008(0.005) & \textbf{0.007(0.006)} & 0.009(0.005) & 0.026(0.034) & 0.020(0.012) \\
			0.4                     & 0.059(0.019) & \textbf{0.055(0.020)} & 0.065(0.020) & 0.090(0.050) & 0.101(0.029) & 0.007(0.005) & \textbf{0.006(0.004)} & 0.008(0.005) & 0.016(0.016) & 0.019(0.011) \\
			0.45                    & 0.056(0.019) & \textbf{0.053(0.021)} & 0.064(0.020) & 0.072(0.028) & 0.098(0.029) & 0.006(0.004) & \textbf{0.005(0.004)} & 0.008(0.005) & 0.010(0.007) & 0.017(0.011) \\
			0.5                     & 0.056(0.020) & \textbf{0.051(0.020)} & 0.067(0.021) & 0.066(0.023) & 0.096(0.030) & 0.006(0.004) & \textbf{0.005(0.004)} & 0.008(0.005) & 0.008(0.005) & 0.016(0.010) \\
			0.55                    & 0.056(0.021) & \textbf{0.053(0.021)} & 0.071(0.023) & 0.076(0.029) & 0.096(0.032) & 0.006(0.004) & \textbf{0.005(0.004)} & 0.009(0.005) & 0.010(0.008) & 0.016(0.010) \\
			0.6                     & 0.059(0.021) & \textbf{0.055(0.022)} & 0.076(0.023) & 0.095(0.050) & 0.099(0.035) & 0.006(0.004) & \textbf{0.006(0.005)} & 0.010(0.006) & 0.017(0.017) & 0.016(0.011) \\
			0.65                    & 0.063(0.021) & \textbf{0.057(0.022)} & 0.081(0.022) & 0.117(0.080) & 0.105(0.040) & 0.007(0.004) & \textbf{0.006(0.005)} & 0.011(0.006) & 0.027(0.035) & 0.018(0.012) \\
			0.7                     & 0.068(0.020) & \textbf{0.061(0.023)} & 0.090(0.023) & 0.141(0.115) & 0.112(0.045) & 0.008(0.004) & \textbf{0.007(0.005)} & 0.013(0.006) & 0.043(0.062) & 0.020(0.014) \\
			0.75                    & 0.074(0.020) & \textbf{0.065(0.024)} & 0.097(0.024) & 0.163(0.155) & 0.120(0.050) & 0.009(0.005) & \textbf{0.008(0.006)} & 0.015(0.007) & 0.065(0.100) & 0.022(0.017) \\
			0.8                     & 0.081(0.021) & \textbf{0.069(0.027)} & 0.106(0.025) & 0.191(0.199) & 0.126(0.055) & 0.011(0.005) & \textbf{0.009(0.007)} & 0.017(0.008) & 0.096(0.155) & 0.025(0.020) \\
			0.85                    & 0.090(0.022) & \textbf{0.076(0.029)} & 0.117(0.026) & 0.226(0.249) & 0.131(0.057) & 0.013(0.006) & \textbf{0.010(0.008)} & 0.021(0.009) & 0.141(0.233) & 0.026(0.022) \\
			0.9                     & 0.103(0.025) & \textbf{0.086(0.038)} & 0.130(0.028) & 0.285(0.305) & 0.137(0.059) & 0.016(0.007) & \textbf{0.014(0.013)} & 0.026(0.011) & 0.218(0.351) & 0.030(0.025) \\
			0.95                    & 0.124(0.030) & \textbf{0.114(0.054)} & 0.153(0.034) & 0.630(0.298) & 0.179(0.078) & 0.024(0.011) & \textbf{0.024(0.024)} & 0.036(0.015) & 0.691(0.496) & 0.055(0.049) \\ \hline
		\end{tabular}%
	}
\end{table}

\begin{table}[H]
	\caption{\small Summary statistics for the ``Wave" model with training sample size $N= 2048$ and the number of replications $R = 100$. The averaged $L_1$ and $L_2^2$ test errors with the corresponding standard deviation (in parentheses) are reported for the estimators trained by different methods.}
	\label{tab:wave_2048}
	\setlength{\tabcolsep}{3pt}
	\renewcommand{\arraystretch}{1.2}
	\resizebox{\textwidth}{!}{%
		\begin{tabular}{c|ccccc|ccccc}
			\hline
			\multirow{2}{*}{$\tau$} & \multicolumn{5}{c|}{$L_1$}                                                        & \multicolumn{5}{c}{$L_2^2$}                                                                \\
			& NQ-Net                & DQR          & DQR*         & NC-QR-DQN    & DQRP         & NQ-Net                & DQR          & DQR*                  & NC-QR-DQN    & DQRP         \\ \hline
			0.05                    & \textbf{0.135(0.037)} & 0.151(0.041) & 0.319(0.091) & 0.558(0.695) & 0.265(0.168) & \textbf{0.054(0.079)} & 0.062(0.043) & 0.190(0.107)          & 1.070(1.949) & 0.164(0.187) \\
			0.1                     & \textbf{0.120(0.031)} & 0.139(0.040) & 0.218(0.061) & 0.416(0.550) & 0.252(0.173) & \textbf{0.043(0.047)} & 0.055(0.042) & 0.118(0.081)          & 0.651(1.192) & 0.158(0.195) \\
			0.15                    & \textbf{0.114(0.026)} & 0.134(0.041) & 0.177(0.043) & 0.349(0.440) & 0.233(0.172) & \textbf{0.036(0.028)} & 0.054(0.045) & 0.084(0.060)          & 0.446(0.778) & 0.142(0.187) \\
			0.2                     & \textbf{0.111(0.025)} & 0.134(0.041) & 0.152(0.038) & 0.293(0.354) & 0.220(0.170) & \textbf{0.034(0.022)} & 0.053(0.043) & 0.064(0.049)          & 0.302(0.519) & 0.130(0.179) \\
			0.25                    & \textbf{0.110(0.026)} & 0.132(0.044) & 0.137(0.038) & 0.242(0.280) & 0.213(0.169) & \textbf{0.032(0.020)} & 0.052(0.044) & 0.050(0.042)          & 0.191(0.341) & 0.122(0.174) \\
			0.3                     & \textbf{0.108(0.026)} & 0.131(0.044) & 0.125(0.034) & 0.204(0.211) & 0.209(0.168) & \textbf{0.031(0.020)} & 0.052(0.046) & 0.038(0.025)          & 0.124(0.211) & 0.118(0.170) \\
			0.35                    & \textbf{0.107(0.025)} & 0.131(0.045) & 0.121(0.034) & 0.170(0.148) & 0.206(0.167) & \textbf{0.031(0.019)} & 0.052(0.046) & 0.033(0.022)          & 0.078(0.119) & 0.116(0.169) \\
			0.4                     & \textbf{0.108(0.027)} & 0.131(0.045) & 0.112(0.031) & 0.141(0.091) & 0.205(0.167) & 0.031(0.019)          & 0.051(0.044) & \textbf{0.030(0.019)} & 0.048(0.058) & 0.114(0.167) \\
			0.45                    & \textbf{0.107(0.025)} & 0.131(0.045) & 0.110(0.029) & 0.113(0.044) & 0.205(0.167) & 0.030(0.019)          & 0.052(0.044) & \textbf{0.027(0.017)} & 0.029(0.024) & 0.113(0.167) \\
			0.5                     & \textbf{0.107(0.025)} & 0.130(0.045) & 0.107(0.026) & 0.102(0.025) & 0.206(0.166) & 0.030(0.019)          & 0.052(0.045) & \textbf{0.025(0.015)} & 0.024(0.016) & 0.113(0.166) \\
			0.55                    & \textbf{0.108(0.025)} & 0.131(0.046) & 0.111(0.030) & 0.113(0.043) & 0.208(0.166) & 0.031(0.019)          & 0.052(0.046) & \textbf{0.027(0.017)} & 0.030(0.027) & 0.114(0.165) \\
			0.6                     & \textbf{0.108(0.025)} & 0.132(0.045) & 0.111(0.030) & 0.138(0.087) & 0.212(0.165) & 0.031(0.019)          & 0.053(0.045) & \textbf{0.026(0.017)} & 0.048(0.062) & 0.115(0.165) \\
			0.65                    & \textbf{0.110(0.025)} & 0.132(0.046) & 0.112(0.031) & 0.168(0.146) & 0.215(0.164) & 0.032(0.019)          & 0.053(0.045) & \textbf{0.027(0.016)} & 0.078(0.123) & 0.117(0.165) \\
			0.7                     & \textbf{0.111(0.027)} & 0.134(0.046) & 0.114(0.030) & 0.200(0.211) & 0.220(0.164) & 0.033(0.020)          & 0.055(0.047) & \textbf{0.029(0.017)} & 0.124(0.215) & 0.121(0.167) \\
			0.75                    & \textbf{0.114(0.027)} & 0.135(0.047) & 0.114(0.030) & 0.238(0.280) & 0.227(0.163) & 0.034(0.020)          & 0.056(0.047) & \textbf{0.027(0.017)} & 0.191(0.345) & 0.127(0.172) \\
			0.8                     & \textbf{0.116(0.027)} & 0.140(0.048) & 0.121(0.032) & 0.282(0.356) & 0.237(0.164) & 0.035(0.020)          & 0.058(0.046) & \textbf{0.031(0.019)} & 0.286(0.528) & 0.137(0.179) \\
			0.85                    & \textbf{0.120(0.027)} & 0.145(0.050) & 0.124(0.036) & 0.342(0.441) & 0.251(0.164) & 0.038(0.021)          & 0.062(0.049) & \textbf{0.034(0.022)} & 0.427(0.791) & 0.151(0.189) \\
			0.9                     & \textbf{0.127(0.028)} & 0.152(0.052) & 0.129(0.035) & 0.433(0.541) & 0.266(0.163) & 0.042(0.023)          & 0.067(0.050) & \textbf{0.037(0.024)} & 0.663(1.193) & 0.164(0.196) \\
			0.95                    & \textbf{0.139(0.034)} & 0.168(0.058) & 0.143(0.035) & 0.692(0.636) & 0.280(0.155) & 0.052(0.030)          & 0.080(0.059) & \textbf{0.045(0.028)} & 1.235(1.879) & 0.172(0.181) \\ \hline
		\end{tabular}%
	}
\end{table}

\begin{table}[H]
	\caption{\small Summary statistics for the univariate ``Linear" model with training sample size $N=2048$ and the number of replications $R = 100$. The averaged $L_1$ and $L_2^2$ test errors with the corresponding standard deviation (in parentheses) are reported for the estimators trained by different methods.}
	\label{tab:linear_2048}
	\setlength{\tabcolsep}{3pt}
	\renewcommand{\arraystretch}{1.2}
	\resizebox{\textwidth}{!}{%
		\begin{tabular}{c|ccccc|ccccc}
			\hline
			\multirow{2}{*}{$\tau$} & \multicolumn{5}{c|}{$L_1$}                                                                 & \multicolumn{5}{c}{$L_2^2$}                                                                \\
			& NQ-Net                & DQR                   & DQR*         & NC-QR-DQN    & DQRP         & NQ-Net                & DQR                   & DQR*         & NC-QR-DQN    & DQRP         \\ \hline
			0.05                    & \textbf{0.158(0.103)} & 0.207(0.109)          & 0.586(0.175) & 1.136(1.155) & 0.306(0.146) & \textbf{0.041(0.051)} & 0.076(0.084)          & 0.457(0.227) & 2.643(3.799) & 0.176(0.160) \\
			0.1                     & \textbf{0.097(0.046)} & 0.111(0.048)          & 0.232(0.056) & 0.663(0.791) & 0.240(0.125) & \textbf{0.015(0.014)} & 0.020(0.016)          & 0.079(0.031) & 1.066(1.612) & 0.105(0.095) \\
			0.15                    & \textbf{0.074(0.026)} & 0.078(0.032)          & 0.207(0.072) & 0.463(0.597) & 0.141(0.068) & \textbf{0.009(0.005)} & 0.010(0.007)          & 0.070(0.041) & 0.570(0.879) & 0.037(0.031) \\
			0.2                     & 0.065(0.023)          & \textbf{0.061(0.025)} & 0.170(0.063) & 0.355(0.457) & 0.093(0.041) & 0.007(0.004)          & \textbf{0.006(0.005)} & 0.049(0.030) & 0.335(0.516) & 0.015(0.012) \\
			0.25                    & 0.060(0.022)          & \textbf{0.056(0.024)} & 0.134(0.046) & 0.280(0.348) & 0.081(0.034) & 0.006(0.004)          & \textbf{0.005(0.004)} & 0.030(0.017) & 0.200(0.307) & 0.011(0.008) \\
			0.3                     & 0.056(0.021)          & \textbf{0.052(0.021)} & 0.109(0.037) & 0.219(0.259) & 0.073(0.027) & 0.005(0.003)          & \textbf{0.004(0.003)} & 0.019(0.011) & 0.117(0.176) & 0.009(0.006) \\
			0.35                    & 0.055(0.021)          & \textbf{0.050(0.019)} & 0.089(0.031) & 0.168(0.181) & 0.064(0.023) & 0.005(0.003)          & \textbf{0.004(0.003)} & 0.013(0.007) & 0.063(0.092) & 0.007(0.005) \\
			0.4                     & 0.050(0.019)          & \textbf{0.046(0.018)} & 0.073(0.023) & 0.122(0.112) & 0.062(0.025) & 0.004(0.003)          & \textbf{0.003(0.002)} & 0.008(0.005) & 0.029(0.040) & 0.006(0.005) \\
			0.45                    & 0.048(0.016)          & \textbf{0.044(0.016)} & 0.059(0.023) & 0.079(0.049) & 0.067(0.030) & 0.004(0.002)          & \textbf{0.003(0.002)} & 0.006(0.004) & 0.010(0.011) & 0.007(0.006) \\
			0.5                     & 0.051(0.019)          & \textbf{0.044(0.017)} & 0.052(0.022) & 0.050(0.018) & 0.076(0.034) & 0.004(0.003)          & \textbf{0.003(0.002)} & 0.004(0.003) & 0.004(0.003) & 0.009(0.008) \\
			0.55                    & 0.051(0.016)          & \textbf{0.046(0.018)} & 0.054(0.024) & 0.077(0.046) & 0.081(0.037) & 0.004(0.003)          & \textbf{0.004(0.003)} & 0.005(0.004) & 0.010(0.010) & 0.010(0.009) \\
			0.6                     & 0.054(0.019)          & \textbf{0.047(0.019)} & 0.060(0.024) & 0.121(0.108) & 0.080(0.039) & 0.005(0.003)          & \textbf{0.004(0.003)} & 0.006(0.004) & 0.028(0.038) & 0.010(0.009) \\
			0.65                    & 0.054(0.020)          & \textbf{0.049(0.019)} & 0.071(0.027) & 0.167(0.178) & 0.077(0.037) & 0.005(0.003)          & \textbf{0.004(0.003)} & 0.008(0.006) & 0.061(0.089) & 0.010(0.009) \\
			0.7                     & 0.060(0.020)          & \textbf{0.050(0.019)} & 0.090(0.030) & 0.219(0.254) & 0.082(0.038) & 0.006(0.003)          & \textbf{0.004(0.003)} & 0.013(0.008) & 0.114(0.172) & 0.012(0.010) \\
			0.75                    & 0.060(0.022)          & \textbf{0.053(0.020)} & 0.109(0.037) & 0.280(0.343) & 0.111(0.056) & 0.006(0.004)          & \textbf{0.005(0.003)} & 0.019(0.012) & 0.197(0.301) & 0.021(0.020) \\
			0.8                     & 0.065(0.024)          & \textbf{0.060(0.026)} & 0.131(0.043) & 0.357(0.451) & 0.166(0.082) & 0.007(0.005)          & \textbf{0.006(0.005)} & 0.027(0.015) & 0.330(0.509) & 0.042(0.038) \\
			0.85                    & \textbf{0.075(0.032)} & 0.075(0.034)          & 0.153(0.052) & 0.458(0.595) & 0.225(0.107) & \textbf{0.009(0.008)} & 0.009(0.009)          & 0.036(0.021) & 0.563(0.869) & 0.074(0.063) \\
			0.9                     & \textbf{0.097(0.047)} & 0.112(0.054)          & 0.176(0.057) & 0.643(0.798) & 0.226(0.119) & \textbf{0.015(0.015)} & 0.021(0.021)          & 0.047(0.027) & 1.050(1.604) & 0.081(0.076) \\
			0.95                    & \textbf{0.144(0.098)} & 0.191(0.104)          & 0.188(0.087) & 1.058(1.201) & 0.322(0.156) & \textbf{0.036(0.055)} & 0.065(0.064)          & 0.057(0.057) & 2.556(3.825) & 0.149(0.122) \\ \hline
		\end{tabular}%
	}
\end{table}

\begin{table}[H]
	\caption{\small Summary statistics for the ``Angle" model with training sample size $N=2048$ and the number of replications $R = 100$. The averaged $L_1$ and $L_2^2$ test errors with the corresponding standard deviation (in parentheses) are reported for the estimators trained by different methods.}
	\label{tab:angle_2048}
	\setlength{\tabcolsep}{3pt}
	\renewcommand{\arraystretch}{1.2}
	\resizebox{\textwidth}{!}{%
		\begin{tabular}{c|ccclc|ccclc}
			\hline
			\multirow{2}{*}{$\tau$} & \multicolumn{5}{c|}{$L_1$}                                                        & \multicolumn{5}{c}{$L_2^2$}                                                                \\
			& NQ-Net       & DQR                   & DQR*         & NC-QR-DQN    & DQRP         & NQ-Net                & DQR                   & DQR*         & NC-QR-DQN    & DQRP         \\ \hline
			0.05                    & 0.070(0.018) & \textbf{0.052(0.029)} & 0.114(0.019) & 0.378(0.393) & 0.131(0.033) & 0.008(0.004)          & \textbf{0.006(0.011)} & 0.020(0.007) & 0.369(0.578) & 0.025(0.012) \\
			0.1                     & 0.056(0.013) & \textbf{0.044(0.024)} & 0.095(0.019) & 0.251(0.329) & 0.097(0.021) & 0.005(0.002)          & \textbf{0.004(0.006)} & 0.013(0.004) & 0.212(0.359) & 0.016(0.008) \\
			0.15                    & 0.048(0.013) & \textbf{0.038(0.019)} & 0.079(0.018) & 0.203(0.267) & 0.091(0.02)  & 0.004(0.002)          & \textbf{0.003(0.004)} & 0.009(0.004) & 0.139(0.235) & 0.015(0.007) \\
			0.2                     & 0.044(0.012) & \textbf{0.036(0.019)} & 0.068(0.016) & 0.170(0.214) & 0.088(0.019) & \textbf{0.003(0.002)} & 0.003(0.004)          & 0.007(0.003) & 0.093(0.156) & 0.015(0.007) \\
			0.25                    & 0.040(0.012) & \textbf{0.033(0.017)} & 0.059(0.015) & 0.142(0.169) & 0.086(0.019) & 0.003(0.002)          & \textbf{0.002(0.003)} & 0.006(0.003) & 0.061(0.100) & 0.014(0.007) \\
			0.3                     & 0.037(0.012) & \textbf{0.031(0.015)} & 0.052(0.013) & 0.116(0.129) & 0.082(0.019) & \textbf{0.002(0.001)} & 0.002(0.002)          & 0.005(0.002) & 0.038(0.061) & 0.013(0.006) \\
			0.35                    & 0.034(0.011) & \textbf{0.030(0.013)} & 0.046(0.011) & 0.094(0.091) & 0.078(0.019) & \textbf{0.002(0.001)} & 0.002(0.001)          & 0.004(0.002) & 0.022(0.033) & 0.012(0.006) \\
			0.4                     & 0.032(0.012) & \textbf{0.030(0.012)} & 0.042(0.011) & 0.072(0.055) & 0.075(0.019) & \textbf{0.002(0.001)} & 0.002(0.001)          & 0.003(0.002) & 0.011(0.015) & 0.011(0.006) \\
			0.45                    & 0.031(0.012) & \textbf{0.030(0.013)} & 0.039(0.010) & 0.052(0.024) & 0.072(0.020) & \textbf{0.002(0.001)} & 0.002(0.001)          & 0.003(0.001) & 0.005(0.004) & 0.010(0.005) \\
			0.5                     & 0.031(0.012) & \textbf{0.030(0.012)} & 0.039(0.010) & 0.039(0.013) & 0.069(0.020) & \textbf{0.002(0.001)} & 0.002(0.001)          & 0.003(0.001) & 0.003(0.002) & 0.009(0.005) \\
			0.55                    & 0.032(0.011) & \textbf{0.029(0.011)} & 0.041(0.010) & 0.050(0.022) & 0.067(0.020) & \textbf{0.002(0.001)} & 0.002(0.001)          & 0.003(0.001) & 0.005(0.004) & 0.008(0.004) \\
			0.6                     & 0.034(0.010) & \textbf{0.030(0.012)} & 0.044(0.011) & 0.071(0.053) & 0.067(0.022) & \textbf{0.002(0.001)} & 0.002(0.001)          & 0.003(0.002) & 0.010(0.014) & 0.008(0.004) \\
			0.65                    & 0.036(0.010) & \textbf{0.031(0.013)} & 0.048(0.012) & 0.092(0.089) & 0.068(0.024) & \textbf{0.002(0.001)} & 0.002(0.001)          & 0.004(0.002) & 0.021(0.032) & 0.008(0.005) \\
			0.7                     & 0.039(0.010) & \textbf{0.034(0.012)} & 0.053(0.012) & 0.115(0.127) & 0.071(0.027) & 0.003(0.001)          & \textbf{0.002(0.002)} & 0.004(0.002) & 0.037(0.059) & 0.008(0.006) \\
			0.75                    & 0.043(0.011) & \textbf{0.035(0.012)} & 0.059(0.014) & 0.139(0.168) & 0.075(0.032) & 0.003(0.002)          & \textbf{0.002(0.001)} & 0.005(0.003) & 0.059(0.098) & 0.009(0.007) \\
			0.8                     & 0.047(0.011) & \textbf{0.038(0.015)} & 0.065(0.016) & 0.168(0.213) & 0.080(0.037) & \textbf{0.003(0.002)} & 0.003(0.003)          & 0.007(0.003) & 0.091(0.152) & 0.011(0.009) \\
			0.85                    & 0.052(0.014) & \textbf{0.041(0.016)} & 0.073(0.019) & 0.205(0.263) & 0.085(0.040) & 0.004(0.002)          & \textbf{0.003(0.002)} & 0.008(0.004) & 0.137(0.231) & 0.012(0.010) \\
			0.9                     & 0.059(0.015) & \textbf{0.045(0.018)} & 0.083(0.023) & 0.249(0.328) & 0.088(0.038) & 0.005(0.003)          & \textbf{0.004(0.004)} & 0.011(0.006) & 0.209(0.354) & 0.013(0.011) \\
			0.95                    & 0.072(0.019) & \textbf{0.056(0.025)} & 0.100(0.028) & 0.497(0.352) & 0.119(0.046) & 0.009(0.004)          & \textbf{0.006(0.008)} & 0.016(0.008) & 0.503(0.536) & 0.023(0.020) \\ \hline
		\end{tabular}%
	}
\end{table}

\begin{table}[H]
	\caption{\small Summary statistics for the multivariate ``Linear" model with training sample size $N=512$ and the number of replications $R = 100$. The averaged $L_1$ and $L_2^2$ test errors with the corresponding standard deviation (in parentheses) are reported for the estimators trained by different methods.}
	\label{tab:multi_linear_512}
	\setlength{\tabcolsep}{3pt}
	\renewcommand{\arraystretch}{1.2}
	\resizebox{\textwidth}{!}{%
		\begin{tabular}{c|ccccc|ccccc}
			\hline
			\multirow{2}{*}{$\tau$} & \multicolumn{5}{c|}{$L_1$}                                                                 & \multicolumn{5}{c}{$L_2^2$}                                                                \\
			& NQ-Net                & DQR                   & DQR*         & NC-QR-DQN    & DQRP         & NQ-Net                & DQR                   & DQR*         & NC-QR-DQN    & DQRP         \\ \hline
			0.05                    & \textbf{0.603(0.145)} & 0.651(0.132)          & 0.618(0.068) & 1.373(0.942) & 1.435(0.205) & \textbf{0.565(0.242)} & 0.643(0.232)          & 0.612(0.134) & 3.086(3.368) & 2.864(1.136) \\
			0.1                     & \textbf{0.461(0.070)} & 0.486(0.084)          & 0.489(0.079) & 0.973(0.551) & 0.814(0.149) & \textbf{0.346(0.105)} & 0.377(0.127)          & 0.394(0.127) & 1.492(1.343) & 1.288(1.021) \\
			0.15                    & \textbf{0.407(0.058)} & 0.420(0.070)          & 0.455(0.062) & 0.925(0.304) & 0.669(0.140) & \textbf{0.273(0.081)} & 0.287(0.092)          & 0.341(0.091) & 1.271(0.585) & 0.974(0.852) \\
			0.2                     & \textbf{0.376(0.055)} & 0.386(0.061)          & 0.468(0.066) & 0.859(0.171) & 0.627(0.125) & \textbf{0.235(0.071)} & 0.245(0.075)          & 0.361(0.102) & 1.108(0.317) & 0.848(0.685) \\
			0.25                    & \textbf{0.354(0.052)} & 0.363(0.058)          & 0.480(0.069) & 0.786(0.110) & 0.609(0.108) & \textbf{0.209(0.063)} & 0.219(0.068)          & 0.380(0.109) & 0.961(0.268) & 0.773(0.545) \\
			0.3                     & \textbf{0.337(0.052)} & 0.349(0.057)          & 0.484(0.071) & 0.724(0.111) & 0.597(0.093) & \textbf{0.190(0.060)} & 0.203(0.064)          & 0.388(0.114) & 0.850(0.289) & 0.719(0.436) \\
			0.35                    & \textbf{0.324(0.052)} & 0.338(0.055)          & 0.484(0.072) & 0.675(0.139) & 0.588(0.082) & \textbf{0.177(0.057)} & 0.191(0.060)          & 0.391(0.117) & 0.774(0.317) & 0.681(0.355) \\
			0.4                     & \textbf{0.314(0.052)} & 0.329(0.052)          & 0.485(0.072) & 0.639(0.166) & 0.581(0.074) & \textbf{0.167(0.055)} & 0.181(0.056)          & 0.394(0.120) & 0.721(0.333) & 0.659(0.298) \\
			0.45                    & \textbf{0.309(0.052)} & 0.324(0.053)          & 0.484(0.072) & 0.618(0.183) & 0.579(0.070) & \textbf{0.162(0.055)} & 0.177(0.056)          & 0.395(0.120) & 0.690(0.342) & 0.656(0.267) \\
			0.5                     & \textbf{0.307(0.052)} & 0.322(0.054)          & 0.486(0.073) & 0.608(0.187) & 0.584(0.070) & \textbf{0.160(0.054)} & 0.174(0.057)          & 0.399(0.123) & 0.675(0.340) & 0.674(0.265) \\
			0.55                    & \textbf{0.309(0.052)} & 0.321(0.054)          & 0.491(0.072) & 0.611(0.175) & 0.597(0.075) & \textbf{0.162(0.055)} & 0.173(0.058)          & 0.408(0.122) & 0.673(0.327) & 0.720(0.295) \\
			0.6                     & \textbf{0.314(0.053)} & 0.322(0.056)          & 0.500(0.075) & 0.628(0.151) & 0.622(0.085) & \textbf{0.167(0.057)} & 0.173(0.059)          & 0.423(0.129) & 0.690(0.305) & 0.798(0.361) \\
			0.65                    & \textbf{0.322(0.054)} & 0.326(0.056)          & 0.508(0.079) & 0.660(0.122) & 0.658(0.100) & \textbf{0.176(0.060)} & 0.177(0.059)          & 0.436(0.138) & 0.732(0.278) & 0.915(0.468) \\
			0.7                     & 0.334(0.054)          & \textbf{0.331(0.056)} & 0.520(0.082) & 0.707(0.099) & 0.706(0.119) & 0.188(0.063)          & \textbf{0.182(0.060)} & 0.456(0.145) & 0.801(0.246) & 1.077(0.622) \\
			0.75                    & 0.349(0.058)          & \textbf{0.339(0.058)} & 0.530(0.083) & 0.768(0.111) & 0.764(0.142) & 0.205(0.071)          & \textbf{0.190(0.062)} & 0.474(0.150) & 0.905(0.231) & 1.287(0.834) \\
			0.8                     & 0.370(0.060)          & \textbf{0.355(0.060)} & 0.548(0.088) & 0.839(0.182) & 0.829(0.167) & 0.230(0.079)          & \textbf{0.205(0.066)} & 0.502(0.164) & 1.043(0.308) & 1.542(1.117) \\
			0.85                    & 0.399(0.064)          & \textbf{0.377(0.062)} & 0.568(0.092) & 0.882(0.328) & 0.899(0.190) & 0.265(0.090)          & \textbf{0.228(0.072)} & 0.534(0.173) & 1.157(0.634) & 1.828(1.476) \\
			0.9                     & 0.443(0.073)          & \textbf{0.417(0.078)} & 0.612(0.106) & 0.993(0.554) & 0.987(0.199) & 0.320(0.111)          & \textbf{0.277(0.105)} & 0.608(0.202) & 1.536(1.369) & 2.139(1.888) \\
			0.95                    & \textbf{0.553(0.130)} & 0.572(0.124)          & 0.744(0.172) & 1.549(0.900) & 1.277(0.200) & \textbf{0.476(0.205)} & 0.508(0.213)          & 0.855(0.350) & 3.639(3.294) & 2.898(2.207) \\ \hline
		\end{tabular}%
	}
\end{table}

\begin{table}[H]
	\caption{\small Summary statistics for the additive model with training sample size $N=512$ and the number of replications $R = 100$. The averaged $L_1$ and $L_2^2$ test errors with the corresponding standard deviation (in parentheses) are reported for the estimators trained by different methods.}
	\label{tab:additive_512}
	\setlength{\tabcolsep}{3pt}
	\renewcommand{\arraystretch}{1.2}
	\resizebox{\textwidth}{!}{%
		\begin{tabular}{c|ccccc|ccccc}
			\hline
			\multirow{2}{*}{$\tau$} & \multicolumn{5}{c|}{$L_1$}                                                                          & \multicolumn{5}{c}{$L_2^2$}                                                                         \\
			& NQ-Net                & DQR                   & DQR*                  & NC-QR-DQN    & DQRP         & NQ-Net                & DQR                   & DQR*                  & NC-QR-DQN    & DQRP         \\ \hline
			0.05                    & 0.760(0.089)          & 0.695(0.102)          & \textbf{0.680(0.073)} & 0.915(0.527) & 0.821(0.097) & 0.913(0.191)          & 0.754(0.203)          & \textbf{0.715(0.142)} & 1.429(1.482) & 1.112(0.240) \\
			0.1                     & 0.684(0.071)          & 0.666(0.088)          & \textbf{0.517(0.052)} & 0.926(0.336) & 0.695(0.066) & 0.750(0.144)          & 0.698(0.171)          & \textbf{0.431(0.086)} & 1.348(0.824) & 0.839(0.162) \\
			0.15                    & 0.638(0.061)          & 0.638(0.080)          & \textbf{0.546(0.075)} & 0.894(0.226) & 0.647(0.055) & 0.657(0.118)          & 0.646(0.152)          & \textbf{0.480(0.125)} & 1.234(0.510) & 0.734(0.130) \\
			0.2                     & 0.602(0.055)          & 0.613(0.070)          & \textbf{0.568(0.085)} & 0.840(0.159) & 0.622(0.051) & 0.589(0.103)          & 0.601(0.132)          & \textbf{0.516(0.144)} & 1.097(0.344) & 0.677(0.112) \\
			0.25                    & 0.577(0.050)          & 0.594(0.063)          & \textbf{0.569(0.082)} & 0.777(0.117) & 0.607(0.049) & 0.541(0.091)          & 0.566(0.114)          & \textbf{0.516(0.141)} & 0.954(0.251) & 0.642(0.101) \\
			0.3                     & \textbf{0.560(0.047)} & 0.579(0.058)          & 0.562(0.075)          & 0.726(0.090) & 0.597(0.049) & 0.509(0.084)          & 0.540(0.104)          & \textbf{0.505(0.127)} & 0.845(0.196) & 0.618(0.095) \\
			0.35                    & \textbf{0.548(0.045)} & 0.567(0.051)          & 0.555(0.069)          & 0.687(0.075) & 0.590(0.048) & \textbf{0.487(0.079)} & 0.518(0.090)          & 0.494(0.115)          & 0.764(0.164) & 0.602(0.092) \\
			0.4                     & \textbf{0.539(0.044)} & 0.558(0.047)          & 0.549(0.059)          & 0.659(0.069) & 0.586(0.048) & \textbf{0.472(0.076)} & 0.503(0.082)          & 0.484(0.100)          & 0.707(0.147) & 0.593(0.091) \\
			0.45                    & \textbf{0.534(0.043)} & 0.553(0.043)          & 0.545(0.051)          & 0.643(0.069) & 0.585(0.048) & \textbf{0.463(0.074)} & 0.494(0.074)          & 0.480(0.087)          & 0.676(0.141) & 0.591(0.091) \\
			0.5                     & \textbf{0.532(0.043)} & 0.552(0.042)          & 0.548(0.044)          & 0.637(0.068) & 0.586(0.048) & \textbf{0.459(0.073)} & 0.492(0.073)          & 0.484(0.076)          & 0.664(0.138) & 0.595(0.093) \\
			0.55                    & \textbf{0.533(0.044)} & 0.555(0.044)          & 0.556(0.041)          & 0.640(0.066) & 0.591(0.048) & \textbf{0.461(0.075)} & 0.497(0.076)          & 0.499(0.073)          & 0.671(0.137) & 0.606(0.096) \\
			0.6                     & \textbf{0.537(0.044)} & 0.560(0.048)          & 0.568(0.043)          & 0.653(0.065) & 0.599(0.049) & \textbf{0.467(0.077)} & 0.506(0.084)          & 0.520(0.078)          & 0.697(0.139) & 0.626(0.102) \\
			0.65                    & \textbf{0.545(0.046)} & 0.569(0.053)          & 0.585(0.051)          & 0.678(0.070) & 0.612(0.052) & \textbf{0.481(0.080)} & 0.521(0.094)          & 0.553(0.094)          & 0.745(0.152) & 0.656(0.111) \\
			0.7                     & \textbf{0.556(0.048)} & 0.580(0.060)          & 0.607(0.058)          & 0.716(0.086) & 0.629(0.055) & \textbf{0.500(0.085)} & 0.541(0.107)          & 0.593(0.110)          & 0.822(0.182) & 0.698(0.126) \\
			0.75                    & \textbf{0.572(0.051)} & 0.596(0.067)          & 0.636(0.073)          & 0.761(0.116) & 0.653(0.060) & \textbf{0.528(0.093)} & 0.569(0.122)          & 0.648(0.140)          & 0.917(0.239) & 0.755(0.146) \\
			0.8                     & \textbf{0.594(0.056)} & 0.614(0.074)          & 0.669(0.085)          & 0.813(0.163) & 0.682(0.066) & \textbf{0.570(0.106)} & 0.601(0.135)          & 0.713(0.165)          & 1.031(0.342) & 0.828(0.172) \\
			0.85                    & \textbf{0.625(0.062)} & 0.636(0.083)          & 0.704(0.098)          & 0.845(0.238) & 0.719(0.072) & \textbf{0.629(0.120)} & 0.639(0.153)          & 0.784(0.196)          & 1.107(0.533) & 0.922(0.202) \\
			0.9                     & 0.671(0.072)          & \textbf{0.663(0.092)} & 0.743(0.113)          & 0.821(0.369) & 0.770(0.077) & 0.720(0.147)          & \textbf{0.689(0.170)} & 0.867(0.230)          & 1.103(0.902) & 1.049(0.231) \\
			0.95                    & 0.741(0.088)          & \textbf{0.696(0.103)} & 0.788(0.135)          & 1.157(0.452) & 0.861(0.088) & 0.865(0.188)          & \textbf{0.750(0.197)} & 0.963(0.283)          & 2.081(1.324) & 1.275(0.260) \\ \hline
		\end{tabular}%
	}
\end{table}

\begin{table}[H]
	\caption{\small Summary statistics for the single index model with training sample size $N=512$ and the number of replications $R = 100$. The averaged $L_1$ and $L_2^2$ test errors with the corresponding standard deviation (in parentheses) are reported for the estimators trained by different methods.}
	\label{tab:sim_512}
	\setlength{\tabcolsep}{3pt}
	\renewcommand{\arraystretch}{1.2}
	\resizebox{\textwidth}{!}{%
		\begin{tabular}{c|ccccc|ccccc}
			\hline
			\multirow{2}{*}{$\tau$} & \multicolumn{5}{c|}{$L_1$}                                                        & \multicolumn{5}{c}{$L_2^2$}                                                       \\
			& NQ-Net       & DQR          & DQR*                  & NC-QR-DQN    & DQRP         & NQ-Net       & DQR          & DQR*                  & NC-QR-DQN    & DQRP         \\ \hline
			0.05                    & 0.621(0.046) & 0.556(0.057) & \textbf{0.473(0.019)} & 0.671(0.273) & 0.598(0.043) & 0.577(0.091) & 0.455(0.103) & \textbf{0.350(0.047)} & 0.715(0.495) & 0.509(0.074) \\
			0.1                     & 0.542(0.046) & 0.499(0.057) & \textbf{0.384(0.020)} & 0.549(0.200) & 0.459(0.032) & 0.452(0.079) & 0.380(0.094) & \textbf{0.220(0.023)} & 0.459(0.300) & 0.315(0.049) \\
			0.15                    & 0.484(0.045) & 0.450(0.053) & \textbf{0.342(0.028)} & 0.522(0.122) & 0.382(0.029) & 0.369(0.071) & 0.316(0.078) & \textbf{0.173(0.030)} & 0.401(0.157) & 0.229(0.042) \\
			0.2                     & 0.437(0.042) & 0.411(0.052) & \textbf{0.316(0.033)} & 0.489(0.071) & 0.331(0.029) & 0.308(0.062) & 0.270(0.070) & \textbf{0.151(0.034)} & 0.352(0.082) & 0.181(0.037) \\
			0.25                    & 0.400(0.041) & 0.377(0.053) & \textbf{0.295(0.035)} & 0.441(0.051) & 0.295(0.029) & 0.263(0.056) & 0.231(0.066) & \textbf{0.135(0.034)} & 0.293(0.061) & 0.151(0.034) \\
			0.3                     & 0.369(0.038) & 0.346(0.050) & \textbf{0.277(0.036)} & 0.392(0.048) & 0.271(0.028) & 0.228(0.049) & 0.199(0.058) & \textbf{0.121(0.032)} & 0.239(0.060) & 0.132(0.031) \\
			0.35                    & 0.346(0.037) & 0.321(0.048) & \textbf{0.260(0.035)} & 0.348(0.056) & 0.256(0.028) & 0.203(0.046) & 0.173(0.052) & \textbf{0.109(0.030)} & 0.196(0.062) & 0.120(0.029) \\
			0.4                     & 0.330(0.036) & 0.303(0.047) & \textbf{0.247(0.034)} & 0.314(0.067) & 0.248(0.028) & 0.186(0.043) & 0.156(0.049) & \textbf{0.101(0.029)} & 0.167(0.065) & 0.114(0.028) \\
			0.45                    & 0.320(0.035) & 0.294(0.044) & \textbf{0.240(0.033)} & 0.292(0.076) & 0.247(0.029) & 0.174(0.040) & 0.146(0.045) & \textbf{0.097(0.027)} & 0.149(0.067) & 0.113(0.028) \\
			0.5                     & 0.316(0.034) & 0.291(0.042) & \textbf{0.240(0.032)} & 0.285(0.077) & 0.251(0.029) & 0.169(0.038) & 0.143(0.041) & \textbf{0.097(0.026)} & 0.143(0.067) & 0.117(0.029) \\
			0.55                    & 0.318(0.033) & 0.298(0.041) & \textbf{0.247(0.032)} & 0.292(0.070) & 0.262(0.029) & 0.169(0.036) & 0.148(0.040) & \textbf{0.103(0.027)} & 0.147(0.065) & 0.126(0.031) \\
			0.6                     & 0.327(0.033) & 0.314(0.039) & \textbf{0.261(0.032)} & 0.312(0.058) & 0.277(0.029) & 0.177(0.036) & 0.161(0.039) & \textbf{0.115(0.028)} & 0.162(0.060) & 0.140(0.033) \\
			0.65                    & 0.344(0.034) & 0.336(0.038) & \textbf{0.282(0.032)} & 0.344(0.045) & 0.299(0.029) & 0.193(0.038) & 0.182(0.040) & \textbf{0.132(0.030)} & 0.189(0.055) & 0.160(0.037) \\
			0.7                     & 0.366(0.035) & 0.366(0.039) & \textbf{0.310(0.033)} & 0.386(0.038) & 0.327(0.029) & 0.215(0.040) & 0.212(0.044) & \textbf{0.156(0.033)} & 0.229(0.050) & 0.189(0.042) \\
			0.75                    & 0.394(0.037) & 0.401(0.041) & \textbf{0.343(0.033)} & 0.435(0.044) & 0.361(0.030) & 0.245(0.045) & 0.251(0.048) & \textbf{0.188(0.037)} & 0.281(0.051) & 0.226(0.049) \\
			0.8                     & 0.430(0.038) & 0.442(0.043) & \textbf{0.383(0.035)} & 0.483(0.068) & 0.404(0.032) & 0.287(0.050) & 0.300(0.055) & \textbf{0.230(0.041)} & 0.341(0.076) & 0.277(0.058) \\
			0.85                    & 0.474(0.041) & 0.493(0.046) & \textbf{0.431(0.036)} & 0.514(0.121) & 0.456(0.033) & 0.344(0.058) & 0.369(0.065) & \textbf{0.285(0.048)} & 0.388(0.153) & 0.345(0.067) \\
			0.9                     & 0.528(0.043) & 0.553(0.048) & \textbf{0.490(0.037)} & 0.574(0.179) & 0.524(0.033) & 0.420(0.068) & 0.459(0.076) & \textbf{0.362(0.055)} & 0.499(0.265) & 0.444(0.076) \\
			0.95                    & 0.602(0.044) & 0.637(0.049) & \textbf{0.571(0.036)} & 0.722(0.240) & 0.632(0.034) & 0.535(0.078) & 0.601(0.090) & \textbf{0.482(0.062)} & 0.792(0.447) & 0.617(0.082) \\ \hline
		\end{tabular}%
	}
\end{table}

\begin{table}[H]
	\caption{\small Summary statistics for the multivariate linear model with training sample size $N=2048$ and the number of replications $R = 100$. The averaged $L_1$ and $L_2^2$ test errors with the corresponding standard deviation (in parentheses) are reported for the estimators trained by different methods.}
	\label{tab:multi_linear_2048}
	\setlength{\tabcolsep}{3pt}
	\renewcommand{\arraystretch}{1.2}
	\resizebox{\textwidth}{!}{%
		\begin{tabular}{c|ccccc|ccccc}
			\hline
			\multirow{2}{*}{$\tau$} & \multicolumn{5}{c|}{$L_1$}                                                                 & \multicolumn{5}{c}{$L_2^2$}                                                                \\
			& NQ-Net                & DQR                   & DQR*         & NC-QR-DQN    & DQRP         & NQ-Net                & DQR                   & DQR*         & NC-QR-DQN    & DQRP         \\ \hline
			0.05                    & \textbf{0.290(0.073)} & 0.343(0.061)          & 0.500(0.051) & 1.154(0.978) & 1.040(0.156) & \textbf{0.136(0.063)} & 0.187(0.061)          & 0.404(0.083) & 2.466(3.361) & 1.540(0.440) \\
			0.1                     & \textbf{0.237(0.030)} & 0.253(0.038)          & 0.325(0.046) & 0.738(0.632) & 0.563(0.097) & \textbf{0.092(0.023)} & 0.105(0.030)          & 0.175(0.049) & 1.030(1.416) & 0.650(0.320) \\
			0.15                    & \textbf{0.212(0.024)} & 0.220(0.029)          & 0.252(0.029) & 0.642(0.413) & 0.467(0.079) & \textbf{0.074(0.016)} & 0.080(0.020)          & 0.105(0.023) & 0.685(0.714) & 0.477(0.214) \\
			0.2                     & \textbf{0.196(0.022)} & 0.201(0.026)          & 0.235(0.026) & 0.572(0.276) & 0.429(0.063) & \textbf{0.064(0.014)} & 0.067(0.017)          & 0.092(0.020) & 0.509(0.381) & 0.385(0.135) \\
			0.25                    & \textbf{0.184(0.020)} & 0.188(0.023)          & 0.233(0.026) & 0.509(0.181) & 0.403(0.052) & \textbf{0.057(0.012)} & 0.059(0.014)          & 0.091(0.019) & 0.393(0.202) & 0.321(0.091) \\
			0.3                     & \textbf{0.175(0.019)} & 0.178(0.021)          & 0.231(0.026) & 0.454(0.110) & 0.380(0.047) & \textbf{0.052(0.011)} & 0.054(0.012)          & 0.089(0.019) & 0.316(0.107) & 0.276(0.067) \\
			0.35                    & \textbf{0.168(0.020)} & 0.172(0.021)          & 0.232(0.026) & 0.408(0.067) & 0.362(0.042) & \textbf{0.048(0.011)} & 0.050(0.012)          & 0.090(0.019) & 0.265(0.078) & 0.243(0.053) \\
			0.4                     & \textbf{0.163(0.019)} & 0.167(0.021)          & 0.232(0.025) & 0.372(0.068) & 0.348(0.038) & \textbf{0.045(0.010)} & 0.048(0.011)          & 0.090(0.019) & 0.233(0.090) & 0.223(0.044) \\
			0.45                    & \textbf{0.160(0.019)} & 0.163(0.020)          & 0.232(0.026) & 0.347(0.092) & 0.340(0.035) & \textbf{0.044(0.009)} & 0.046(0.011)          & 0.090(0.020) & 0.215(0.104) & 0.212(0.041) \\
			0.5                     & \textbf{0.159(0.019)} & 0.161(0.020)          & 0.231(0.025) & 0.336(0.102) & 0.339(0.035) & \textbf{0.043(0.010)} & 0.044(0.011)          & 0.089(0.018) & 0.207(0.107) & 0.211(0.044) \\
			0.55                    & \textbf{0.159(0.019)} & 0.160(0.020)          & 0.232(0.026) & 0.341(0.087) & 0.345(0.040) & \textbf{0.043(0.009)} & 0.044(0.011)          & 0.091(0.019) & 0.207(0.098) & 0.221(0.055) \\
			0.6                     & 0.162(0.020)          & \textbf{0.160(0.021)} & 0.236(0.025) & 0.360(0.063) & 0.360(0.050) & 0.045(0.010)          & \textbf{0.043(0.011)} & 0.094(0.018) & 0.218(0.079) & 0.243(0.071) \\
			0.65                    & 0.166(0.019)          & \textbf{0.161(0.021)} & 0.238(0.026) & 0.389(0.071) & 0.384(0.062) & 0.047(0.010)          & \textbf{0.044(0.011)} & 0.095(0.020) & 0.241(0.071) & 0.278(0.093) \\
			0.7                     & 0.173(0.021)          & \textbf{0.163(0.021)} & 0.244(0.029) & 0.427(0.125) & 0.415(0.074) & 0.050(0.011)          & \textbf{0.045(0.011)} & 0.099(0.022) & 0.282(0.117) & 0.327(0.120) \\
			0.75                    & 0.179(0.022)          & \textbf{0.168(0.023)} & 0.245(0.029) & 0.476(0.201) & 0.452(0.085) & 0.054(0.013)          & \textbf{0.047(0.013)} & 0.100(0.023) & 0.352(0.225) & 0.390(0.149) \\
			0.8                     & 0.190(0.023)          & \textbf{0.176(0.024)} & 0.249(0.029) & 0.535(0.300) & 0.489(0.092) & 0.061(0.014)          & \textbf{0.051(0.014)} & 0.104(0.023) & 0.460(0.414) & 0.461(0.178) \\
			0.85                    & 0.205(0.026)          & \textbf{0.191(0.029)} & 0.259(0.033) & 0.596(0.442) & 0.526(0.089) & 0.070(0.017)          & \textbf{0.059(0.018)} & 0.112(0.027) & 0.628(0.757) & 0.533(0.197) \\
			0.9                     & 0.227(0.033)          & \textbf{0.216(0.037)} & 0.277(0.043) & 0.708(0.655) & 0.592(0.072) & 0.085(0.023)          & \textbf{0.075(0.026)} & 0.127(0.037) & 0.996(1.453) & 0.637(0.185) \\
			0.95                    & \textbf{0.266(0.062)} & 0.311(0.061)          & 0.321(0.081) & 1.266(0.955) & 0.990(0.172) & \textbf{0.115(0.048)} & 0.152(0.060)          & 0.168(0.077) & 2.718(3.301) & 1.393(0.371) \\ \hline
		\end{tabular}%
	}
\end{table}

\begin{table}[H]
	\caption{\small Summary statistics for the single index model with training sample size $N=2048$ and the number of replications $R = 100$. The averaged $L_1$ and $L_2^2$ test errors with the corresponding standard deviation (in parentheses) are reported for the estimators trained by different methods.}
	\label{tab:sim_2048}
	\setlength{\tabcolsep}{3pt}
	\renewcommand{\arraystretch}{1.2}
	\resizebox{\textwidth}{!}{%
		\begin{tabular}{c|ccccc|ccccc}
			\hline
			\multirow{2}{*}{$\tau$} & \multicolumn{5}{c|}{$L_1$}                                                                 & \multicolumn{5}{c}{$L_2^2$}                                                                \\
			& NQ-Net       & DQR                   & DQR*                  & NC-QR-DQN    & DQRP         & NQ-Net       & DQR                   & DQR*                  & NC-QR-DQN    & DQRP         \\ \hline
			0.05                    & 0.470(0.009) & \textbf{0.444(0.006)} & 0.445(0.008)          & 0.608(0.260) & 0.527(0.025) & 0.319(0.012) & \textbf{0.283(0.014)} & 0.291(0.025)          & 0.595(0.456) & 0.392(0.046) \\
			0.1                     & 0.385(0.011) & 0.355(0.009)          & \textbf{0.348(0.005)} & 0.471(0.204) & 0.410(0.020) & 0.212(0.013) & \textbf{0.171(0.009)} & 0.176(0.010)          & 0.350(0.284) & 0.245(0.034) \\
			0.15                    & 0.330(0.013) & 0.304(0.010)          & \textbf{0.290(0.008)} & 0.409(0.151) & 0.338(0.018) & 0.158(0.014) & 0.125(0.010)          & \textbf{0.116(0.005)} & 0.248(0.179) & 0.171(0.024) \\
			0.2                     & 0.289(0.015) & 0.265(0.011)          & \textbf{0.249(0.011)} & 0.366(0.108) & 0.285(0.016) & 0.124(0.015) & 0.097(0.010)          & \textbf{0.085(0.007)} & 0.195(0.106) & 0.125(0.017) \\
			0.25                    & 0.252(0.017) & 0.231(0.011)          & \textbf{0.215(0.012)} & 0.319(0.075) & 0.243(0.015) & 0.097(0.014) & 0.076(0.009)          & \textbf{0.064(0.007)} & 0.149(0.060) & 0.094(0.013) \\
			0.3                     & 0.222(0.018) & 0.199(0.011)          & \textbf{0.185(0.013)} & 0.273(0.050) & 0.209(0.014) & 0.078(0.014) & 0.059(0.007)          & \textbf{0.049(0.007)} & 0.111(0.033) & 0.072(0.010) \\
			0.35                    & 0.194(0.019) & 0.168(0.011)          & \textbf{0.155(0.012)} & 0.232(0.031) & 0.183(0.013) & 0.062(0.012) & 0.044(0.007)          & \textbf{0.036(0.006)} & 0.082(0.020) & 0.058(0.009) \\
			0.4                     & 0.172(0.019) & 0.140(0.012)          & \textbf{0.129(0.013)} & 0.196(0.028) & 0.165(0.013) & 0.051(0.012) & 0.032(0.006)          & \textbf{0.026(0.005)} & 0.062(0.020) & 0.049(0.008) \\
			0.45                    & 0.157(0.019) & 0.119(0.013)          & \textbf{0.110(0.013)} & 0.169(0.043) & 0.156(0.014) & 0.043(0.011) & 0.024(0.006)          & \textbf{0.020(0.005)} & 0.050(0.024) & 0.044(0.008) \\
			0.5                     & 0.152(0.019) & 0.112(0.014)          & \textbf{0.104(0.013)} & 0.158(0.053) & 0.155(0.014) & 0.040(0.010) & 0.022(0.005)          & \textbf{0.019(0.005)} & 0.047(0.026) & 0.044(0.008) \\
			0.55                    & 0.156(0.018) & 0.122(0.014)          & \textbf{0.112(0.013)} & 0.169(0.042) & 0.164(0.014) & 0.042(0.009) & 0.026(0.006)          & \textbf{0.022(0.005)} & 0.050(0.024) & 0.048(0.008) \\
			0.6                     & 0.173(0.017) & 0.146(0.013)          & \textbf{0.134(0.013)} & 0.194(0.024) & 0.180(0.014) & 0.049(0.010) & 0.035(0.006)          & \textbf{0.029(0.006)} & 0.061(0.018) & 0.057(0.009) \\
			0.65                    & 0.195(0.016) & 0.176(0.013)          & \textbf{0.162(0.013)} & 0.229(0.025) & 0.203(0.014) & 0.060(0.010) & 0.049(0.008)          & \textbf{0.041(0.007)} & 0.081(0.015) & 0.070(0.010) \\
			0.7                     & 0.220(0.016) & 0.209(0.012)          & \textbf{0.195(0.013)} & 0.271(0.044) & 0.232(0.015) & 0.075(0.011) & 0.066(0.009)          & \textbf{0.057(0.008)} & 0.110(0.027) & 0.089(0.012) \\
			0.75                    & 0.251(0.014) & 0.246(0.012)          & \textbf{0.231(0.013)} & 0.318(0.069) & 0.268(0.015) & 0.094(0.011) & 0.090(0.010)          & \textbf{0.078(0.009)} & 0.149(0.053) & 0.115(0.015) \\
			0.8                     & 0.287(0.014) & 0.286(0.012)          & \textbf{0.270(0.012)} & 0.370(0.099) & 0.310(0.016) & 0.120(0.013) & 0.119(0.011)          & \textbf{0.104(0.010)} & 0.201(0.095) & 0.152(0.019) \\
			0.85                    & 0.328(0.012) & 0.332(0.012)          & \textbf{0.315(0.011)} & 0.421(0.140) & 0.362(0.017) & 0.154(0.013) & 0.159(0.013)          & \textbf{0.141(0.010)} & 0.261(0.165) & 0.203(0.023) \\
			0.9                     & 0.381(0.010) & 0.390(0.011)          & \textbf{0.373(0.009)} & 0.487(0.191) & 0.431(0.018) & 0.207(0.012) & 0.218(0.013)          & \textbf{0.197(0.010)} & 0.368(0.266) & 0.280(0.027) \\
			0.95                    & 0.467(0.008) & 0.480(0.010)          & \textbf{0.460(0.007)} & 0.646(0.237) & 0.540(0.026) & 0.317(0.012) & 0.335(0.016)          & \textbf{0.309(0.013)} & 0.661(0.414) & 0.423(0.039) \\ \hline
		\end{tabular}%
	}
\end{table}

\section{Self-calibration Results}\label{sec_selfcali}

The empirical risk minimization typically results in estimators $\hat{f}_N$ whose risk $\mathcal{R}(\hat{f}_N)$ is close to the optimal risk $\mathcal{R}(Q_Y)$ in expectation or with high probability. However, small excess risk in general only implies that the empirical risk minimizer $\hat{f}_N$ is close to the target $Q_Y$ in a weak sense 
\cite[Remark 3.18][]{steinwart2007compare}. 
In the following, we bridge the gap between the excess risk and the mean integrated error of the estimated conditional quantile functions $\hat{f}_N$. To this end, we need the following condition on the conditional distribution of $Y$ given $X$.
\begin{customassum}{A1} \label{assump4}
	There exist constants $C >0$ and $c>0$ such that for any $\vert \delta\vert\le C$,
	$$\vert P_{Y|X}(Q^\tau_Y(x+\delta))-P_{Y|X}(Q^\tau_Y(x))\vert\ge c\vert \delta\vert,$$
	for all $\tau\in(0,1)$ and $x\in\mathcal{X}$ up to a negligible set, where $ P_{Y|X}(\cdot)$ denotes the conditional distribution function of $Y$ given $X=x$.
\end{customassum}

Assumption \ref{assump4} is a mild condition on the distribution of $Y$ as it holds if $Y$ has a density that is bounded away from zero on any compact interval \citep{padilla2021adaptive}. In particular, no moment assumptions are made on the distribution of $Y$. This condition is weaker than Condition 2 in \cite{he1994convergence} where the density function of response is required to be lower-bounded everywhere by some positive constant. Assumption \ref{assump4} is also weaker than Condition D.1 in \cite{belloni2011}, which requires the conditional density of $Y$ given $X=x$ to be continuously differentiable and bounded away from zero uniformly for all quantiles in $(0,1)$ and all $x$ in the support $\mathcal{X}$.

Assumption \ref{assump4} leads to a bound on the mean integrated error of the estimated quantile functions based on the error bound of excess risk, which is the following self-calibration results. Here we omit the proof of the Lemma and refer to that of Lemma 7 in \cite{shen2024nonparametric}.

\begin{customthm}{S1}[Self-calibration]\label{calib}
	Suppose that Assumption \ref{assump4} holds. For any $\tau\in(0,1)$ and $f\in\mathcal{F}_N$, let $f_{\tau}$ be an component of the output of $f$. Denote
	\begin{align*}
		\Delta^2(f_{\tau}(x),Q^\tau_{Y}(x))=&\mathbb{E}[ \min\{\vert Q^{\tau}_Y(X)-f_\tau(X)\vert,\vert Q_{Y}^\tau(X)-f_\tau(X)\vert^2\}],
	\end{align*}
	where $X$ is the predictor vector. Then we have
	$$\sum_{k=1}^K\Delta^2(f_{\tau_k}(x),Q_Y^{\tau_k}(x))/K\le \max\{2/C,4/(Cc)\} \mathcal{R}(f),$$
	where $C,c>0$ are defined in Assumption \ref{assump1}. Especially, if $\Vert f_{\tau}(x)-Q_Y^{\tau}(x)\Vert_\infty \le C$ for any $\tau\in(0,1)$, then
	$$\sum_{k=1}^K\mathbb{E}\vert f_{\tau_k}(X)-Q_Y^{\tau_k}(X)\vert^2/K\le \frac{2}{c} \mathcal{R}(f).$$
	Further if $f_{\tau},Q_Y^{\tau}$ are bounded by $\mathcal{B}$ with $C\ge 2\mathcal{B}$ for all $\tau\in(0,1)$, then $\Vert f_{\tau}-Q_Y{\tau}\Vert_\infty \le C$.
\end{customthm}

With Lemma \ref{calib}, the upper bounds obtained for excess risk $\mathcal{R}(\hat{f}_N)$ can subsequently be the upper bounds of the prediction error of the conditional quantiles $\mathbb{E}\Vert \hat{f}_N^{\tau_k}(X)-Q_Y^{\tau_k}(X)\Vert^2, k=1,\ldots, K$. The $L_2$ error of the estimated conditional quantile can scale as the nonparametric rate $O_p(N^{-2\beta/(2\beta+d_0)})$ with respect to the sample size $N$ and input dimension $d_0$ and smoothness of the target feature extraction mapping. Our established statistical learning theories ensure the consistency and efficiency of our proposed NQ net estimators, which also can lay a theoretical foundation for deriving learning guarantees of many existing non-crossing deep quantile estimators \citep{bondell2010noncrossing,padilla2022quantile,yan2023ensemble}.

\section{Proofs of Theorems and Lemmas}\label{appendix:proof}

\subsection{Proof of Lemma \ref{decomposition}.}

For the empirical risk minimizer $\hat{f}_N$ and any $f\in\mathcal{F}_N$, we have
\begin{align*}
	\mathbb{E} [\mathcal{R}(\hat{f}_{N})]
	\le &  \mathbb{E} [ \mathcal{R}(\hat{f}_{N}) + 2(\mathcal{R}_N (f)-\mathcal{R}_N (\hat{f}_{N}))]\\
	= &  \mathbb{E} [ \mathcal{R}(\hat{f}_{N}) -2\mathcal{R}_N (\hat{f}_{N})+ 2\mathcal{R}_N (f)]\\
	= &  \mathbb{E} [ \mathcal{R}(\hat{f}_{N}) -2\mathcal{R}_N (\hat{f}_{N})]+ 2\mathbb{E} [\mathcal{R}_N (f)]\\
	= &  \mathbb{E} [ \mathcal{R}(\hat{f}_{N}) -2\mathcal{R}_N (\hat{f}_{N})]+ 2\mathcal{R} (f),
\end{align*}
where the first inequality follows from the definition of $\hat{f}_{N}$ as the minimizer of $\mathcal{R}_N(f)$ over $\mathcal{F}_{N}$, the last equality is due to the function $f$ is independent with the sample. In the above inequality, since $f$  is arbitrary in $\mathcal{F}_N$, we have
\begin{align*}
	\mathbb{E} [\mathcal{R}(\hat{f}_{N})]
	\le & \mathbb{E} [ \mathcal{R}(\hat{f}_{N}) -2\mathcal{R}_N (\hat{f}_{N})]+ 2\inf_{f\in\mathcal{F}_N}\mathcal{R} (f).
\end{align*}

\subsection{Proof of Theorem \ref{sto_error}}

When the model is correctly specified, there is no bias or approximation error, and we know that
$\mathbb{E} [\mathcal{R}(\hat{f}_{N})]\le \mathbb{E} [ \mathcal{R}(\hat{f}_{N})-2\mathcal{R}_N(\hat{f}_{N})]$. Then it is sufficient to give upper bounds of $\mathbb{E} [ \mathcal{R}(\hat{f}_{N})-2\mathcal{R}_N(\hat{f}_{N})]$. 
{
	
Firstly, by the definition of risk \(\mathcal{R}\), it is the average of the quantile losses at different quantile levels. We decompose the quantity \(\mathbb{E} [ \mathcal{R}(\hat{f}_{N}) - 2\mathcal{R}_N(\hat{f}_{N})]\) into summations of the losses at different quantile levels and bound each part of the summation. 
	
	To this end, let \(S = \{Z_i = (X_i, Y_i)\}_{i=1}^N\) be a sample from the distribution \(Z = (X, Y)\) used to estimate \(\hat{f}_N\). With a bit of abuse of notation, let \(Z = (X, Y)\) be another sample independent of \(S\). Given any \(f=(f_1,\ldots,f_K)\in\mathcal{F}\) (which may be random) and \(Z_i = (X_i, Y_i)\), for \(k = 1, \ldots, K\), we define the operator \(g_k\) by
	\[
	g_k(f, Z_i) := \mathbb{E}\left\{\rho_{\tau_k}(Y_i - f_k(X_i)) - \rho_{\tau_k}(Y_i - Q^{\tau_k}_Y(X_i)) \mid X_i, f \right\},
	\]
	where \(f_{\tau_k}\) is the \(k\)th component of the output of \(f \in \mathcal{F}_N\) and \(Q^{\tau_k}_Y\) is the target conditional quantile curve at level \(\tau_k\).  We write \(g_k(f, Z_i)\) instead of \(g_k(f, X_i)\) to emphasize that  \(g_k\) is an operator whose result depends on the conditional distribution of \(Y_i\) given \(X_i\) and $f$. It is worth noting that given two certain random variables $U$ and $V$, the $g_k(f,(X_i,U))$ and $g_k(f,(X_i,V))$ are both functions of \(f\) and \(X_i\), but they can be different from each other. For example, 
	\[
	g_k(f, Z_i) = g_k(f_k, (X_i, Y_i)) \neq g_k(f_k, (X_i, Y)) 
	\]
	because \(g_k(f, (X_i, Y)) = \mathbb{E}\left\{\rho_{\tau_k}(Y - f_k(X_i)) - \rho_{\tau_k}(Y - Q^{\tau_k}_Y(X_i)) \mid X_i, f\right\}\) is a conditional expectation with respect to \((X_i, Y)\) (given a fixed function $f$), which has a different joint distribution than \((X_i, Y_i)\). Now \(\hat{f}_N\) is a random element depending on the sample \(S = \{Z_i = (X_i, Y_i)\}_{i=1}^N\), it follows that \(g_k(\hat{f}_N, Z_i) \neq g_k(\hat{f}_N, Z)\). We also have 
	\[
	\mathcal{R}(\hat{f}_N) = \frac{1}{K} \sum_{k=1}^K \mathbb{E}_{X}[g_k(\hat{f}_N, Z)] =\frac{1}{K} \sum_{k=1}^K \mathbb{E}_{Z}[g_k(\hat{f}_N, Z)] 
	\]
	based on the independence between \(\hat{f}_N\) and \(Z'\), and 
	\[
	\mathbb{E}_S\left[\mathcal{R}_N(\hat{f}_N)\right] = \mathbb{E}_S\left[\frac{1}{KN} \sum_{k=1}^K \sum_{i=1}^N g_k(\hat{f}_N, Z_i) \right]
	\]
	based on the definition of \(g_k\) and the law of total expectation. Now we have,
	\begin{align}\label{ineq1}
		\mathbb{E}_S[\mathcal{R}(\hat{f}_{N})-2\mathcal{R}_N(\hat{f}_{N})]=\frac{1}{K}\sum_{k=1}^K\mathbb{E}_{S}\left[\mathbb{E}_{Z}[g_k(\hat{f}_N,Z)]-2\frac{1}{N}\sum_{i=1}^N [g_k(\hat{f}_N,Z_i)]\right].
	\end{align}
}
Based on equality (\ref{ineq1}), we intend to give upper bounds of
$$\mathbb{E}_{S}\left[\mathbb{E}_{Z}[g_k(\hat{f}_N,Z)]-2\frac{1}{N}\sum_{i=1}^N [g_k(\hat{f}_N,Z_i)]\right]\qquad{\rm for}\ k=1,\ldots,K.$$
Note that for any random variable $W$, we have $$\mathbb{E}[W]\le \mathbb{E}[W\times I(W>0)]\le \int_{t=0}^\infty \mathbb{P}(W>t).$$ In light of this, we derive upper bounds of 
\begin{align}\label{prob}
	\mathbb{P}\left(\mathbb{E}_{Z}[g_k(\hat{f}_N, Z)]-2\frac{1}{N}\sum_{i=1}^N [g_k(\hat{f}_N, Z_i)]>t\right)
\end{align}
for $t>0$ and $k=1,\ldots,K$. Since $\hat{f}_N\in\mathcal{F}_N$, we know
\begin{align*}
	&\mathbb{P}\Big(\mathbb{E}_{Z}[g_k(\hat{f}_N,Z)]-2\frac{1}{N}\sum_{i=1}^N [g_k(\hat{f}_N, Z_i)]>t\Big)\\
	&\qquad\qquad\le \mathbb{P}\left(\exists f\in\mathcal{F}_N: \mathbb{E}_{Z}[g_k(f, Z)]-2\frac{1}{N}\sum_{i=1}^N [g_k(f, Z_i)]>t\right).
\end{align*}
It then suffices to prove the uniform upper bound for the tail probability. By symmetrization, conditioning, and covering techniques, we can prove the following result for $k=1,\ldots, K$
\begin{align} \notag
&\mathbb{P}\left(\exists f\in\mathcal{F}_N: \mathbb{E}\big\{g_k(f,Z)\big\}-\frac{1}{N}\sum_{i=1}^{N}\big[g_k(f,Z_i)\big]>\epsilon\bigg(\alpha+\beta+\mathbb{E}\big\{g( f, Z)\big\}\bigg)\right)\\ \label{tail1}&\qquad\qquad\qquad\qquad\qquad\qquad\le 14 \mathcal{N}_{N}\left(\frac{\epsilon \beta}{7}, \mathcal{F}^{(k)}_N, \Vert\cdot\Vert_\infty \right) \exp \left(-\frac{3\epsilon^{2}(1-\epsilon) \alpha N}{40(1+\epsilon) \mathcal{B}^2}\right),
\end{align}
where $\alpha,\beta>0$, $0<\epsilon\le 1/2$, $\mathcal{F}^{(k)}_N$ denotes the function class of the $k$th component of the output of $f\in\mathcal{F}_N$. Here $\mathcal{N}_N(\delta,\mathcal{F},\Vert\cdot\Vert_{\infty})$ denotes the covering number of function class $\mathcal{F}$ with radius $\delta$ under the norm $\Vert\cdot\Vert_{\infty}$ (See Definition \ref{cove_number} in Appendix). To facilitate the presentation, we defer the proof of (\ref{tail1}) to the end of this subsection.
We take $\epsilon=1/2$ and $\alpha=\beta=t/2$ in (\ref{tail1}), then (\ref{prob}) can be bounded by
\begin{align} \notag
&\mathbb{P}\left(\mathbb{E}_{Z }[g_k(\hat{f}_N, Z )]-2\frac{1}{N}\sum_{i=1}^N [g_k(\hat{f}_N, Z_i)]>t\right)\le 14 \mathcal{N}_{N}\left(\frac{t}{28}, \mathcal{F}^{(k)}_N, \Vert\cdot\Vert_\infty \right) \exp \left(-\frac{tN}{320 \mathcal{B}^2}\right).
\end{align}
Then for $k=1,\ldots,K$ and $a_N>1/N$,  we have
\begin{align*}
&\mathbb{E}_S\Big(\mathbb{E}_{Z }[g_k(\hat{f}_N, Z )]-2\frac{1}{N}\sum_{i=1}^N [g_k(\hat{f}_N, Z_i)]\Big)\\
\le & \int_0^\infty \mathbb{P}\Big( \mathbb{E}_{Z }[g_k(\hat{f}_N, Z )]-2\frac{1}{N}\sum_{i=1}^N [g_k(\hat{f}_N, Z_i)]>t\Big) dt\\
\le & \int_0^{a_N} 1 dt+ \int_{a_n}^\infty 14 \mathcal{N}_{N}\left(\frac{t}{28}, \mathcal{F}^{(k)}_N, \Vert\cdot\Vert_\infty \right) \exp \left(-\frac{tN}{320 \mathcal{B}^2}\right) dt\\
\le& a_N + 14 \mathcal{N}_{N}\left(\frac{1}{28N}, \mathcal{F}^{(k)}_N, \Vert\cdot\Vert_\infty \right) \int_{a_N}^\infty \exp \left(-\frac{tN}{320 \mathcal{B}^2}\right) dt\\
=& a_N+ 14 \mathcal{N}_{N}\left(\frac{1}{28N}, \mathcal{F}^{(k)}_N, \Vert\cdot\Vert_\infty \right) \exp \left(-\frac{a_NN}{320 \mathcal{B}^2}\right)\frac{320 \mathcal{B}^2}{N} dt.
\end{align*}
Choosing $a_N=\log \{14\mathcal{N}_N(1/(28N),\mathcal{F}^{(k)}_N,\Vert\cdot\Vert_\infty)\} \cdot 320\mathcal{B}^2/(N)$, we get
\begin{align*} \notag\mathbb{E}_{S}\left[\mathbb{E}_{Z }[g_k(\hat{f}_N,Z )]-2\frac{1}{N}\sum_{i=1}^N [g_k(\hat{f}_N,Z_i)]\right]\le \frac{320\log [14e \mathcal{N}_N(1/(28N),\mathcal{F}^{(k)}_N,\Vert\cdot\Vert_\infty)] \mathcal{B}^2}{N}.
\end{align*}
Combining (\ref{ineq1}), we have proved that 
$$\mathbb{E}[\mathcal{R}(\hat{f}_{N})-2\mathcal{R}_N(\hat{f}_{N})]\le \frac{320\log [14e\Pi_{k=1}^K \mathcal{N}_N(1/(28N),\mathcal{F}^{(k)}_N,\Vert\cdot\Vert_\infty)] \mathcal{B}^2}{N}.$$
Next, we bound of the covering numbers $\mathcal{N}_N(1/N,\mathcal{F}^{(k)}_{N},\Vert\cdot\Vert_\infty)$. For each $k=1,\ldots,K$ we know from Lemma \ref{thm:covering_number} that
\begin{align*}
\log\mathcal{N}_N(\delta,\mathcal{F}_N^{(k)},\Vert\cdot\Vert_\infty)\le \text{Pdim}(\mathcal{F}_N^{(k)})\log\bigg(\frac{eN\mathcal{B}}{\delta \cdot \text{Pdim}(\mathcal{F}_N^{(k)})}\bigg)
\end{align*}
for $N\geq \text{Pdim}(\mathcal{F}_N^{(k)})$, where $\text{Pdim}(\mathcal{F}_N^{(k)})$ is the pseudo-dimension of $\mathcal{F}_N^{(k)}$ stated in Definition \ref{def:pseudo_dimension}. Note that the networks in $\mathcal{F}_N^{(k)}$ are ReLU (piecewise linear) neural networks with output activated by ELU, by applying Theorems 3 and 6 of \cite{bartlett2019nearly}, we know that there exist two universal constants $c$ and $C$ such that
\begin{equation}\label{bartlett_pdim}
c\cdot \mathcal{S}\mathcal{D} \log(\mathcal{S}/\mathcal{D})\leq \text{Pdim}(\mathcal{F}_N^{(k)})\leq C\cdot \mathcal{S}\mathcal{D}\log(\mathcal{S}),
\end{equation}
where $\mathcal{S}$ and $\mathcal{D}$ are the size and depth of the given $\mathcal{F}_N^{(k)}$, respectively. Combining the above results, we finally get the upper bound for the excess risk
$$\mathbb{E}[\mathcal{R}(\hat{f}_{N})-2\mathcal{R}_N(\hat{f}_{N})]\le CK\mathcal{B}^3\frac{\mathcal{S}\mathcal{D}\log(\mathcal{S})\log(N)}{N},$$
for $N\ge C_1\mathcal{S}\mathcal{D}\log \mathcal{S}$.

Lastly, we present the proof of the main inequality (\ref{tail1}) bounding the tail probability.

{\noindent \bf Proof of inequality (\ref{tail1})}

{

We follow the proof of Theorem 11.4 in \cite{gyorfi2002distribution}. Recall that we have introduced the notations of $S=\{Z_{i}\}_{i=1}^N$ with $Z,Z_1,\ldots,Z_N$ being independent and identically distributed, and the operator $g_k(Z,f)$ for any $Z$ and $f$ (which may be random). For ease of presentation, we denote $g_f(Z):=g_k(Z,f)=\mathbb{E}\left\{\rho_{\tau_k}(Y - f(X)) - \rho_{\tau_k}(Y - Q^{\tau_k}_Y(X)) \mid X, f \right\}$. 
}
Now we rewrite the interested probability as follows
\begin{align}\label{main}
\mathbb{P}\left\{\exists f \in \mathcal{F}_N: \mathbb{E} g_{f}(Z)-\frac{1}{N} \sum_{i=1}^{N} g_{f}\left(Z_{i}\right) \geq \epsilon\left(\alpha+\beta+\mathbb{E} g_{f}(Z)\right)\right\}
\end{align}

The proof will proceed in several steps.

{\noindent STEP 1. Symmetrization by a ghost sample.}

Replace the expectation on the left-hand side of the inequality in (\ref{main}) by the empirical mean based on the ghost sample $S^{\prime}=\{Z_{i}^\prime\}_{i=1}^N$ of i.i.d. random
variables distributed as $Z$ and independent of $S$. Consider a function $\hat{f}_N \in \mathcal{F}_N$ depending upon $S$ such that
$$
\mathbb{E}\left\{g_{\hat{f}_N}(Z) \mid S \right\}-\frac{1}{N} \sum_{i=1}^{N} g_{\hat{f}_N}\left(Z_{i}\right) \geq \epsilon(\alpha+\beta)+\epsilon \mathbb{E}\left\{g_{\hat{f}_N}(Z) \mid S \right\}
$$
if such a function exists in $\mathcal{F}_N$, otherwise choose an arbitrary function in $\mathcal{F}_N$.
{ Based on the definition of $Q_y^{\tau_k}$, it is the minimizer of $\mathbb{E}[\rho_{\tau_k}(Y-f(X))\mid X, f]$ over fixed measurable functions $f$. Thus, $g_f(Z)\ge 0$ for any function $f$ that is fixed and independent of $Z$. In addition,  $|f_k(X)| \leq \mathcal{B}$ for any $f=(f_1,\ldots,f_K)\in\mathcal{F}_N$ and $|Q_Y^{\tau_k}(X)| \leq \mathcal{B}$, together with the Lipschitz property of check loss $\rho_{\tau_k}$ imply $g_f(Z)\le 2\mathcal{B}$ for $f=(f_1,\ldots,f_K)\in\mathcal{F}_N$. Now given the sample $S$ independent of $Z$, it can be checked that $0\le g_{\hat{f}_N}(Z)\le 2\mathcal{B}$ and 
$$
\operatorname{Var}\left\{g_{\hat{f}_N}(Z) \mid S \right\} \leq  2\mathcal{B}\times \mathbb{E}\left\{g_{\hat{f}_N}(Z) \mid  S \right\},
$$
by the definition of $g_k$ and the independence between $\hat{f}_N$ and $Z$.} Using Chebyshev's inequality, we have
$$
\begin{aligned}
& \mathbb{P}\left\{\mathbb{E}\left\{g_{\hat{f}_N}(Z) \mid   S \right\}-\frac{1}{N} \sum_{i=1}^{N} g_{\hat{f}_N}\left(Z_{i}^{\prime}\right)\right.\left.>\frac{\epsilon}{2}(\alpha+\beta)+\frac{\epsilon}{2} \mathbb{E}\left\{g_{\hat{f}_N}(Z) \mid   S \right\} \mid   S \right\} \\
& \leq \frac{\operatorname{Var}\left\{g_{\hat{f}_N}(Z) \mid   S \right\}}{N \cdot\left(\frac{\epsilon}{2}(\alpha+\beta)+\frac{\epsilon}{2} \mathbb{E}\left\{g_{\hat{f}_N}(Z) \mid   S \right\}\right)^{2}} \\
& \leq \frac{2\mathcal{B}\times \mathbb{E}\left\{g_{\hat{f}_N}(Z) \mid   S \right\}}{N \cdot\left(\frac{\epsilon}{2}(\alpha+\beta)+\frac{\epsilon}{2} \mathbb{E}\left\{g_{\hat{f}_N}(Z) \mid   S \right\}\right)^{2}} \\
& \leq \frac{8\mathcal{B}}{\epsilon^{2}(\alpha+\beta) N},
\end{aligned}
$$

where the last inequality follows from
$$
h(x)=\frac{x}{(a+x)^{2}} \leq h(a)=\frac{1}{4 a}
$$
for all $x \geq 0$ and all $a>0$. Thus, for $n>\frac{64\mathcal{B}}{\epsilon^{2}(\alpha+\beta)}$,

\begin{align}\label{bound1}
\mathbb{P}\left\{\mathbb{E}\left\{g_{\hat{f}_N}(Z) \mid   S \right\}-\frac{1}{N} \sum_{i=1}^{N} g_{\hat{f}_N}\left(Z_{i}^{\prime}\right) \leq \frac{\epsilon}{2}(\alpha+\beta)+\frac{\epsilon}{2} \mathbb{E}\left\{g_{\hat{f}_N}(Z) \mid   S \right\} \mid   S \right\}\geq \frac{7}{8}
\end{align}

Hence

$$
\begin{aligned}
& \mathbb{P}\left\{\exists f \in \mathcal{F}_N: \frac{1}{N} \sum_{i=1}^{N} g_{f}\left(Z_{i}^{\prime}\right)-\frac{1}{N} \sum_{i=1}^{N} g_{f}\left(Z_{i}\right) \geq \frac{\epsilon}{2}(\alpha+\beta)+\frac{\epsilon}{2} \mathbb{E} g_{f}(Z)\right\} \\
& \geq \mathbb{P}\left\{\frac{1}{N} \sum_{i=1}^{N} g_{\hat{f}_N}\left(Z_{i}^{\prime}\right)-\frac{1}{N} \sum_{i=1}^{N} g_{\hat{f}_N}\left(Z_{i}\right) \geq \frac{\epsilon}{2}(\alpha+\beta)+\frac{\epsilon}{2} \mathbb{E}\left\{g_{\hat{f}_N}(Z) \mid   S \right\}\right\}
\end{aligned}
$$

$$
\begin{aligned}
& \geq \mathbb{P}\left\{\mathbb{E}\left\{g_{\hat{f}_N}(Z) \mid   S \right\}-\frac{1}{N} \sum_{i=1}^{N} g_{\hat{f}_N}\left(Z_{i}\right) \geq \epsilon(\alpha+\beta)+\epsilon \mathbb{E}\left\{g_{\hat{f}_N}(Z) \mid   S \right\}\right., \\
& \qquad\qquad\left.\mathbb{E}\left\{g_{\hat{f}_N}(Z) \mid   S \right\}-\frac{1}{N} \sum_{i=1}^{N} g_{\hat{f}_N}\left(Z_{i}^{\prime}\right) \leq \frac{\epsilon}{2}(\alpha+\beta)+\frac{\epsilon}{2} \mathbb{E}\left\{g_{\hat{f}_N}(Z) \mid   S \right\}\right\} \\
& =\mathbb{E}\left\{I_{\left\{\mathbb{E}\left\{g_{\hat{f}_N}(Z) \mid   S \right\}-\frac{1}{N} \sum_{i=1}^{N} g_{\hat{f}_N}\left(Z_{i}\right) \geq \epsilon(\alpha+\beta)+\epsilon \mathbb{E}\left\{g_{\hat{f}_N}(Z) \mid   S \right\}\right\}}\right. \\
& \qquad\qquad \left.\times \mathbb{E}\left\{I_{\left\{\mathbb{E}\left\{g_{\hat{f}_N}(Z) \mid   S \right\}-\frac{1}{N} \sum_{i=1}^{N} g_{\hat{f}_N}\left(Z_{i}^{\prime}\right) \leq \frac{\epsilon}{2}(\alpha+\beta)+\frac{\epsilon}{2} \mathbb{E}\left\{g_{\hat{f}_N}(Z) \mid   S \right\}\right\}} \mid   S \right\}\right\} \\
& =\mathbb{E}\left\{I _ { \{ \cdots \} } \mathbf { P } \left\{\mathbb{E}\left\{g_{\hat{f}_N}(Z) \mid   S \right\}-\frac{1}{N} \sum_{i=1}^{N} g_{\hat{f}_N}\left(Z_{i}^{\prime}\right)\right.\right. \left.\left.\leq \frac{\epsilon}{2}(\alpha+\beta)+\frac{\epsilon}{2} \mathbb{E}\left\{g_{\hat{f}_N}(Z) \mid   S \right\} \mid   S \right\}\right\} \\
& \geq \frac{7}{8} \mathbb{P}\left\{\mathbb{E}\left\{g_{\hat{f}_N}(Z) \mid   S \right\}-\frac{1}{N} \sum_{i=1}^{N} g_{\hat{f}_N}\left(Z_{i}\right) \geq \epsilon(\alpha+\beta)+\epsilon \mathbb{E}\left\{g_{\hat{f}_N}(Z) \mid   S \right\}\right\} \\
& =\frac{7}{8} \mathbb{P}\left\{\exists f \in \mathcal{F}_N: \mathbb{E} g_{f}(Z)-\frac{1}{N} \sum_{i=1}^{N} g_{f}\left(Z_{i}\right) \geq \epsilon(\alpha+\beta)+\epsilon \mathbb{E} g_{f}(Z)\right\}
\end{aligned}
$$

where the last inequality follows from (\ref{bound1}). Thus we have shown that, for $N>\frac{32\mathcal{B}}{\epsilon^{2}(\alpha+\beta)}$,

\begin{align}\notag
&\mathbb{P}\left\{\exists f \in \mathcal{F}_N: \mathbb{E} g_{f}(Z)-\frac{1}{N} \sum_{i=1}^{N} g_{f}\left(Z_{i}\right) \geq \epsilon(\alpha+\beta)+\epsilon \mathbb{E} g_{f}(Z)\right\} \\ \label{bound2}
&\qquad \leq \frac{8}{7} \mathbb{P}\left\{\exists f \in \mathcal{F}_N: \frac{1}{N} \sum_{i=1}^{N} g_{f}\left(Z_{i}^{\prime}\right)-\frac{1}{N} \sum_{i=1}^{N} g_{f}\left(Z_{i}\right)\right. \left.\geq \frac{\epsilon}{2}(\alpha+\beta)+\frac{\epsilon}{2} \mathbb{E} g_{f}(Z)\right\} .
\end{align}

{\noindent STEP 2. Replacement of the expectation in (\ref{bound2}) by an empirical mean of the ghost sample.}

First, we introduce additional conditions in the probability (\ref{bound2}),

$$
\begin{aligned}
& \mathbb{P}\left\{\exists f \in \mathcal{F}_N: \frac{1}{N} \sum_{i=1}^{N} g_{f}\left(Z_{i}^{\prime}\right)-\frac{1}{N} \sum_{i=1}^{N} g_{f}\left(Z_{i}\right) \geq \frac{\epsilon}{2}(\alpha+\beta)+\frac{\epsilon}{2} \mathbb{E} g_{f}(Z)\right\} \\
& \leq \mathbb{P}\left\{\exists f \in \mathcal{F}_N: \frac{1}{N} \sum_{i=1}^{N} g_{f}\left(Z_{i}^{\prime}\right)-\frac{1}{N} \sum_{i=1}^{N} g_{f}\left(Z_{i}\right) \geq \frac{\epsilon}{2}(\alpha+\beta)+\frac{\epsilon}{2} \mathbb{E} g_{f}(Z),\right. \\
&\qquad\quad \frac{1}{N} \sum_{i=1}^{N} g_{f}^{2}\left(Z_{i}\right)-\mathbb{E} g_{f}^{2}(Z) \leq \epsilon\left(\alpha+\beta+\frac{1}{N} \sum_{i=1}^{N} g_{f}^{2}\left(Z_{i}\right)+\mathbb{E} g_{f}^{2}(Z)\right)\left.\right.,
\end{aligned}
$$

\begin{align}\notag
&\left.\qquad\quad \frac{1}{N} \sum_{i=1}^{N} g_{f}^{2}\left(Z_{i}^{\prime}\right)-\mathbb{E} g_{f}^{2}(Z) \leq \epsilon\left(\alpha+\beta+\frac{1}{N} \sum_{i=1}^{N} g_{f}^{2}\left(Z_{i}^{\prime}\right)+\mathbb{E} g_{f}^{2}(Z)\right)\right\} \\\label{bound3}
&\qquad+2 \mathbb{P}\left\{\exists f \in \mathcal{F}_N: \frac{\frac{1}{N} \sum_{i=1}^{N} g_{f}^{2}\left(Z_{i}\right)-\mathbb{E} g_{f}^{2}(Z)}{\left(\alpha+\beta+\frac{1}{N} \sum_{i=1}^{N} g_{f}^{2}\left(Z_{i}\right)+\mathbb{E} g_{f}^{2}(Z)\right)}>\epsilon\right\} 
\end{align}

Application of Theorem 11.6 in \cite{gyorfi2002distribution} to the second probability on the right-hand side of (\ref{bound3}) yields

$$
\begin{aligned}
& \mathbb{P}\left\{\exists f \in \mathcal{F}_N: \frac{\frac{1}{N} \sum_{i=1}^{N} g_{f}^{2}\left(Z_{i}\right)-\mathbb{E} g_{f}^{2}(Z)}{\left(\alpha+\beta+\frac{1}{N} \sum_{i=1}^{N} g_{f}^{2}\left(Z_{i}\right)+\mathbb{E} g_{f}^{2}(Z)\right)}>\epsilon\right\} \\
& \leq 4  \mathcal{N}_{N}\left(\frac{(\alpha+\beta) \epsilon}{5},\left\{g_{f}: f \in \mathcal{F}_N\right\},  \Vert\cdot\Vert_\infty \right) \exp \left(-\frac{3 \epsilon^{2}(\alpha+\beta) N}{40\mathcal{B}^2}\right) .
\end{aligned}
$$

Now we consider the first probability on the right-hand side of (\ref{bound3}). The second inequality inside the probability implies

$$
(1+\epsilon) \mathbb{E} g_{f}^{2}(Z) \geq(1-\epsilon) \frac{1}{N} \sum_{i=1}^{N} g_{f}^{2}\left(Z_{i}\right)-\epsilon(\alpha+\beta)
$$

which is equivalent to

$$
\frac{1}{2\mathcal{B}} \mathbb{E} g_{f}^{2}(Z) \geq \frac{1-\epsilon}{2 \mathcal{B}(1+\epsilon)} \frac{1}{N} \sum_{i=1}^{N} g_{f}^{2}\left(Z_{i}\right)-\epsilon \frac{(\alpha+\beta)}{2\mathcal{B}(1+\epsilon)}
$$

We can deal similarly with the third inequality. Using this and the inequality $\mathbb{E} g_{f}(Z) \geq \frac{1}{\mathcal{B}} \mathbb{E} g_{f}^{2}(Z)=2 \frac{1}{2\mathcal{B}} \mathbb{E} g_{f}^{2}(Z)$ we can bound the first probability on the right-hand side of  (\ref{bound3}) by

$$
\begin{aligned}
\mathbb{P}\Bigg\{& \exists f \in \mathcal{F}_N: \frac{1}{N} \sum_{i=1}^{N} g_{f}\left(Z_{i}^{\prime}\right)-\frac{1}{N} \sum_{i=1}^{N} g_{f}\left(Z_{i}\right)\geq \epsilon(\alpha+\beta) /2\\
& \qquad+ \frac{\epsilon}{2}\left(\frac{1-\epsilon}{2\mathcal{B}(1+\epsilon)} \frac{1}{N} \sum_{i=1}^{N} g_{f}^{2}\left(Z_{i}\right)-\frac{\epsilon(\alpha+\beta)}{2\mathcal{B}(1+\epsilon)}+\frac{1-\epsilon}{2\mathcal{B}(1+\epsilon)} \frac{1}{N} \sum_{i=1}^{N} g_{f}^{2}\left(Z_{i}^{\prime}\right)-\frac{\epsilon(\alpha+\beta)}{2\mathcal{B}(1+\epsilon)}\right)\Bigg\}.
\end{aligned}
$$

This shows
\begin{align}\notag
&\mathbb{P}\left\{\exists f \in \mathcal{F}_N: \frac{1}{N} \sum_{i=1}^{N} g_{f}\left(Z_{i}^{\prime}\right)-\frac{1}{N} \sum_{i=1}^{N} g_{f}\left(Z_{i}\right) \geq \epsilon(\alpha+\beta) / 2+\epsilon \mathbb{E} g_{f}(Z) / 2\right\}\\\notag
& \leq \mathbb{P}\left\{\exists f \in \mathcal{F}_N: \frac{1}{N} \sum_{i=1}^{N}\left(g_{f}\left(Z_{i}^{\prime}\right)-g_{f}\left(Z_{i}\right)\right)\right. \\\notag
& \left.\quad \geq \epsilon(\alpha+\beta) / 2-\frac{\epsilon^{2}(\alpha+\beta)}{2 \mathcal{B}(1+\epsilon)}+\frac{\epsilon(1-\epsilon)}{4 \mathcal{B}(1+\epsilon)} \frac{1}{N} \sum_{i=1}^{N}\left(g_{f}^{2}\left(Z_{i}\right)+g_{f}^{2}\left(Z_{i}^{\prime}\right)\right)\right\} \\\label{bound4}
& \quad+  8 \mathcal{N}_{N}\left(\frac{(\alpha+\beta) \epsilon}{5},\left\{g_{f}: f \in \mathcal{F}_N\right\},  \Vert\cdot\Vert_\infty \right) \exp \left(-\frac{3 \epsilon^{2}(\alpha+\beta) N}{40\mathcal{B}^2}\right) 
\end{align}

{\noindent STEP 3. Additional randomization by random signs.}

Let $U_{1}, \ldots, U_{N}$ be independent and uniformly distributed over the set $\{-1,1\}$ and independent of $Z_{1}, \ldots, Z_{N}, Z_{1}^{\prime}, \ldots, Z_{N}^{\prime}$. Because of the independence and identical distribution of $Z_{1}, \ldots, Z_{N}^{\prime}$ the joint distribution of $S,  S^\prime $ is not affected by the random interchange of the corresponding components in $  S $ and $  S^\prime $. Therefore the first probability on the right-hand side of inequality (\ref{bound4}) is equal to
$$
\begin{aligned}
& \mathbb{P}\left\{\exists f \in \mathcal{F}_N: \frac{1}{N} \sum_{i=1}^{N} U_{i}\left(g_{f}\left(Z_{i}^{\prime}\right)-g_{f}\left(Z_{i}\right)\right)\right. \\
& \left.\geq \frac{\epsilon}{2}(\alpha+\beta)-\frac{\epsilon^{2}(\alpha+\beta)}{2\mathcal{B}(1+\epsilon)}+\frac{\epsilon(1-\epsilon)}{4\mathcal{B}(1+\epsilon)}\left(\frac{1}{N} \sum_{i=1}^{N}\left(g_{f}^{2}\left(Z_{i}\right)+g_{f}^{2}\left(Z_{i}^{\prime}\right)\right)\right)\right\}
\end{aligned}
$$

and this, in turn, by the union bound, is bounded by

\begin{align}\notag
& \mathbb{P}\left\{\exists f \in \mathcal{F}_N:\left|\frac{1}{N} \sum_{i=1}^{N} U_{i} g_{f}\left(Z_{i}^{\prime}\right)\right|\right. \\\notag
& \left.\quad \geq \frac{1}{2}\left(\epsilon(\alpha+\beta) / 2-\frac{\epsilon^{2}(\alpha+\beta)}{2\mathcal{B}(1+\epsilon)}\right)+\frac{\epsilon(1-\epsilon)}{4\mathcal{B}(1+\epsilon)} \frac{1}{N} \sum_{i=1}^{N} g_{f}^{2}\left(Z_{i}^{\prime}\right)\right\} \\\notag
& +\mathbb{P}\left\{\exists f \in \mathcal{F}_N:\left|\frac{1}{N} \sum_{i=1}^{N} U_{i} g_{f}\left(Z_{i}\right)\right|\right. \\\notag
& \left.\quad \geq \frac{1}{2}\left(\epsilon(\alpha+\beta) / 2-\frac{\epsilon^{2}(\alpha+\beta)}{2\mathcal{B}(1+\epsilon)}\right)+\frac{\epsilon(1-\epsilon)}{4\mathcal{B}(1+\epsilon)} \frac{1}{N} \sum_{i=1}^{N} g_{f}^{2}\left(Z_{i}\right)\right\} \\\notag
& =2 \mathbb{P}\left\{\exists f \in \mathcal{F}_N:\left|\frac{1}{N} \sum_{i=1}^{N} U_{i} g_{f}\left(Z_{i}\right)\right|\right. \\ \label{bound5}
& \left.\quad \geq \epsilon(\alpha+\beta) / 4-\frac{\epsilon^{2}(\alpha+\beta)}{4\mathcal{B}(1+\epsilon)}+\frac{\epsilon(1-\epsilon)}{4\mathcal{B}(1+\epsilon)} \frac{1}{N} \sum_{i=1}^{N} g_{f}^{2}\left(Z_{i}\right)\right\}.
\end{align}

{\noindent STEP 4. Conditioning and using covering.}

Next, we condition the probability on the right-hand side of (\ref{bound5}) on $ S $,
which is equivalent to fixing $z_{1}, \ldots, z_{N}$ and considering

$$
\begin{aligned}
\mathbb{P}\{\exists f & \in \mathcal{F}_N:\left|\frac{1}{N} \sum_{i=1}^{N} U_{i} g_{f}\left(z_{i}\right)\right| \\
& \left.\geq \epsilon(\alpha+\beta) / 4-\frac{\epsilon^{2}(\alpha+\beta)}{64 B^{2}(1+\epsilon)}+\frac{\epsilon(1-\epsilon)}{64 B^{2}(1+\epsilon)} \frac{1}{N} \sum_{i=1}^{N} g_{f}^{2}\left(z_{i}\right)\right\}
\end{aligned}
$$

Let $\delta>0$ and let $\mathcal{G}_{\delta}$ be an  $\delta$-cover of $\mathcal{G}_{\mathcal{F}_N}=\left\{g_{f}: f \in \mathcal{F}_N\right\}$ on $z_{1}, \ldots, z_{N}$. Fix $f \in \mathcal{F}_N$. Then there exists $g \in \mathcal{G}_{\delta}$ such that

$$
\frac{1}{N} \sum_{i=1}^{N}\left|g\left(z_{i}\right)-g_{f}\left(z_{i}\right)\right|<\delta
$$

Without losing generality we can assume $-\mathcal{B} \leq g(z) \leq \mathcal{B}$. This implies

$$
\begin{aligned}
\left|\frac{1}{N} \sum_{i=1}^{N} U_{i} g_{f}\left(z_{i}\right)\right| & =\left|\frac{1}{N} \sum_{i=1}^{N} U_{i} g\left(z_{i}\right)+\frac{1}{N} \sum_{i=1}^{N} U_{i}\left(g_{f}\left(z_{i}\right)-g\left(z_{i}\right)\right)\right| \\
& \leq\left|\frac{1}{N} \sum_{i=1}^{N} U_{i} g\left(z_{i}\right)\right|+\frac{1}{N} \sum_{i=1}^{N}\left|g_{f}\left(z_{i}\right)-g\left(z_{i}\right)\right| \\
& <\left|\frac{1}{N} \sum_{i=1}^{N} U_{i} g\left(z_{i}\right)\right|+\delta
\end{aligned}
$$

and

$$
\begin{aligned}
\frac{1}{N} \sum_{i=1}^{N} g_{f}^{2}\left(z_{i}\right) & =\frac{1}{N} \sum_{i=1}^{N} g^{2}\left(z_{i}\right)+\frac{1}{N} \sum_{i=1}^{N}\left(g_{f}^{2}\left(z_{i}\right)-g^{2}\left(z_{i}\right)\right) \\
& =\frac{1}{N} \sum_{i=1}^{N} g^{2}\left(z_{i}\right)+\frac{1}{N} \sum_{i=1}^{N}\left(g_{f}\left(z_{i}\right)+g\left(z_{i}\right)\right)\left(g_{f}\left(z_{i}\right)-g\left(z_{i}\right)\right) \\
& \geq \frac{1}{N} \sum_{i=1}^{N} g^{2}\left(z_{i}\right)-\mathcal{B} \frac{1}{N} \sum_{i=1}^{N}\left|g_{f}\left(z_{i}\right)-g\left(z_{i}\right)\right| \\
& \geq \frac{1}{N} \sum_{i=1}^{N} g^{2}\left(z_{i}\right)-\mathcal{B} \delta
\end{aligned}
$$

It follows

\begin{align*}
&\mathbb{P}\left\{\exists f \in \mathcal{F}_N:\left|\frac{1}{N} \sum_{i=1}^{N} U_{i} g_{f}\left(z_{i}\right)\right| \geq\epsilon(\alpha+\beta)/4-\frac{\epsilon^{2}(\alpha+\beta)}{4\mathcal{B}(1+\epsilon)}+\frac{\epsilon(1-\epsilon)}{4\mathcal{B}(1+\epsilon)}\frac{1}{n}\sum_{i=1}^{n} g_{f}^{2}\left(z_{i}\right) \right\}\\
&\leq \mathbb{P}\left\{\exists g \in \mathcal{G}_{\delta}:\left|\frac{1}{N} \sum_{i=1}^{N} U_{i} g\left(z_{i}\right)\right|+\delta\geq \epsilon(\alpha+\beta)/4-\frac{\epsilon^{2}(\alpha+\beta)}{4\mathcal{B}(1+\epsilon)}+\frac{\epsilon(1-\epsilon)}{4\mathcal{B}(1+\epsilon)}\left(\frac{1}{N} \sum_{i=1}^{N} g^{2}\left(z_{i}\right)-\mathcal{B}\delta\right)\right\} \\
&\leq\left|\mathcal{G}_{\delta}\right| \max _{g \in \mathcal{G}_{\delta}} \mathbb{P}\left\{\left|\frac{1}{N} \sum_{i=1}^{N} U_{i} g\left(z_{i}\right)\right|\geq \epsilon(\alpha+\beta)/4-\frac{\epsilon^{2}(\alpha+\beta)}{4\mathcal{B}(1+\epsilon)}-\delta-\delta\frac{\epsilon(1-\epsilon)}{4(1+\epsilon)}+\frac{\epsilon(1-\epsilon)}{4\mathcal{B}(1+\epsilon)} \frac{1}{N} \sum_{i=1}^{N} g^{2}\left(z_{i}\right)\right\}.
\end{align*}

Next we set $\delta=\epsilon \beta / 7$. This, together with $\mathcal{B} \geq 1$ and $0<\epsilon \leq \frac{1}{2}$, implies
\begin{align*}
& \frac{\epsilon \beta}{4}-\frac{\epsilon^{2} \beta}{4\mathcal{B}(1+\epsilon)}-\delta-\delta \frac{\epsilon(1-\epsilon)}{4(1+\epsilon)} \\
& =\epsilon\beta(\frac{1}{4}-\frac{1}{12}-\frac{1}{7}-\frac{1}{7}\frac{1}{20}) \\
& \ge 0.
\end{align*}
Thus
\begin{align*}
&\mathbb{P}\left\{\exists f \in \mathcal{F}_N:\left|\frac{1}{N} \sum_{i=1}^{N} U_{i} g_{f}\left(z_{i}\right)\right|\geq \epsilon(\alpha+\beta) / 4-\frac{\epsilon^{2}(\alpha+\beta)}{4\mathcal{B}(1+\epsilon)}+\frac{\epsilon(1-\epsilon)}{4\mathcal{B}(1+\epsilon)} \frac{1}{N} \sum_{i=1}^{N} g_{f}^{2}\left(z_{i}\right)\right\} \\
&\leq\left|\mathcal{G}_{\frac{\epsilon \beta}{7}}\right| \max _{g \in\mathcal{G}_{\frac{\epsilon \beta}{7}}} \mathbb{P}\left\{\left|\frac{1}{N} \sum_{i=1}^{N} U_{i} g\left(z_{i}\right)\right|\geq \frac{\epsilon \alpha}{4}-\frac{\epsilon^{2}\alpha}{4\mathcal{B}(1+\epsilon)}+\frac{\epsilon(1-\epsilon)}{4\mathcal{B}(1+\epsilon)} \frac{1}{N} \sum_{i=1}^{N} g^{2}\left(z_{i}\right)\right\}.
\end{align*}

{\noindent STEP 5. Application of Bernstein's inequality.}

In this step, we use Bernstein's inequality to bound

$$
\mathbb{P}\left\{\left|\frac{1}{N} \sum_{i=1}^{N} U_{i} g\left(z_{i}\right)\right| \geq \frac{\epsilon \alpha}{4}-\frac{\epsilon^{2} \alpha}{4\mathcal{B}(1+\epsilon)}+\frac{\epsilon(1-\epsilon)}{4\mathcal{B}(1+\epsilon)} \frac{1}{N} \sum_{i=1}^{N} g^{2}\left(z_{i}\right)\right\},
$$
where $z_{1}, \ldots, z_{N} $ are fixed and $g$ satisfies $-\mathcal{B} \leq$ $g(z) \leq \mathcal{B}$. First we relate $\frac{1}{N} \sum_{i=1}^{N} g^{2}\left(z_{i}\right)$ to the variance of $U_{i} g\left(z_{i}\right)$,

$$
\frac{1}{N} \sum_{i=1}^{N} \operatorname{Var}\left(U_{i} g\left(z_{i}\right)\right)=\frac{1}{N} \sum_{i=1}^{N} g^{2}\left(z_{i}\right) \operatorname{Var}\left(U_{i}\right)=\frac{1}{N} \sum_{i=1}^{N} g^{2}\left(z_{i}\right)
$$

Thus the probability above is equal to
$$
\mathbb{P}\left\{\left|\frac{1}{N} \sum_{i=1}^{N} V_{i}\right| \geq A_{1}+A_{2} \sigma^{2}\right\}
$$
where
$$
\begin{gathered}
V_{i}=U_{i} g\left(z_{i}\right), \quad \sigma^{2}=\frac{1}{N} \sum_{i=1}^{N} \operatorname{Var}\left(U_{i} g\left(z_{i}\right)\right) \\
A_{1}=\frac{\epsilon \alpha}{4}-\frac{\epsilon^{2} \alpha}{4\mathcal{B}(1+\epsilon)}, \quad A_{2}=\frac{\epsilon(1-\epsilon)}{4\mathcal{B}(1+\epsilon)} .
\end{gathered}
$$

Observe that $V_{1}, \ldots, V_{N}$ are independent random variables satisfying $\left|V_{i}\right| \leq$ $\left|g\left(z_{i}\right)\right| \leq \mathcal{B} (i=1, \ldots, N)$, and that $A_{1}, A_{2} \geq 0$. We have, by Bernstein's inequality,

\begin{align}\notag
\mathbb{P}\left\{\left|\frac{1}{N} \sum_{i=1}^{N} V_{i}\right| \geq A_{1}+A_{2} \sigma^{2}\right\} & \leq 2 \exp \left(-\frac{N\left(A_{1}+A_{2} \sigma^{2}\right)^{2}}{2 \sigma^{2}+2\left(A_{1}+A_{2} \sigma^{2}\right) \frac{2\mathcal{B}}{3}}\right) \\\notag
& =2 \exp \left(-\frac{N A_{2}^{2}}{\frac{4}{3} \mathcal{B} A_{2}} \cdot \frac{\left(\frac{A_{1}}{A_{2}}+\sigma^{2}\right)^{2}}{\frac{A_{1}}{A_{2}}+\left(\frac{1}{2}+\frac{3}{2\mathcal{B} A_{2}}\right) \sigma^{2}}\right) \\ \label{bound6}
& =2 \exp \left(-\frac{3N \cdot A_{2}}{4 \mathcal{B}} \cdot \frac{\left(\frac{A_{1}}{A_{2}}+\sigma^{2}\right)^{2}}{\frac{A_{1}}{A_{2}}+\left(\frac{1}{2}+\frac{3}{2\mathcal{B} A_{2}}\right) \sigma^{2}}\right).
\end{align}

An easy calculation (cf. Problem 11.1 in \cite{gyorfi2002distribution}) shows that, for arbitrary $a, b, u>0$, one has
$$
\frac{(a+u)^{2}}{a+b \cdot u} \geq \frac{\left(a+\frac{b-2}{b} a\right)^{2}}{a+b \frac{b-2}{b} a}=4 a \frac{b-1}{b^{2}}
$$
Thus setting $a=A_{1} / A_{2}, b=\left(\frac{1}{2}+\frac{3}{2\mathcal{B} A_{2}}\right), u=\sigma^{2}$, and using the bound above we get, for the exponent in (\ref{bound6}),

$$
\begin{aligned}
\frac{3 N \cdot A_{2}}{4\mathcal{B} \cdot \frac{\left(\frac{A_{1}}{A_{2}}+\sigma^{2}\right)^{2}}{\frac{A_{1}}{A_{2}}+\left(\frac{1}{2}+\frac{3}{2\mathcal{B} A_{2}}\right) \sigma^{2}}} & \geq \frac{3 N\cdot A_{2}}{2\mathcal{B}} \cdot 4 \cdot \frac{A_{1}}{A_{2}} \frac{\frac{3}{2\mathcal{B} A_{2}}-\frac{1}{2}}{\left(\frac{1}{2}+\frac{3}{2\mathcal{B} A_{2}}\right)^{2}} \\
& =12 N \frac{A_{1} A_{2}(3-\mathcal{B}A_2)}{\left(\mathcal{B} A_{2}+3\right)^{2}}\\
&\ge \frac{177}{5} N \frac{A_{1} A_{2}}{\left(\mathcal{B} A_{2}+3\right)^{2}}.
\end{aligned}
$$

Substituting the formulas for $A_{1}$ and $A_{2}$ and noticing

$$
A_{1}=\frac{\epsilon \alpha}{4}-\frac{\epsilon^{2} \alpha}{4 \mathcal{B}(1+\epsilon)} \geq \frac{\epsilon \alpha}{4}-\frac{\epsilon \alpha}{12}=\frac{\epsilon \alpha}{6}
$$

we obtain

$$
\begin{aligned}
\frac{177}{5} N \frac{A_{1} A_{2}}{\left(\mathcal{B} A_{2}+3\right)^{2}} & \geq \frac{177}{5}  N \frac{\epsilon \alpha}{6} \cdot \frac{\epsilon(1-\epsilon)}{4\mathcal{B}(1+\epsilon)} \cdot \frac{1}{\left(\frac{\epsilon(1-\epsilon)}{4(1+\epsilon)}+3\right)^{2}} \\
& \geq \frac{177}{5} N \frac{\epsilon^{2}(1-\epsilon) \cdot \alpha}{24 \mathcal{B}(1+\epsilon)} \cdot \frac{1}{\left(\frac{1}{20}+3\right)^{2}} \\
& =\frac{590}{3721} \cdot \frac{\epsilon^{2}(1-\epsilon)}{1+\epsilon} \cdot \frac{\alpha \cdot N}{\mathcal{B}} \\
& \geq \frac{\epsilon^{2}(1-\epsilon) \cdot \alpha \cdot N}{7 \mathcal{B}(1+\epsilon)} .
\end{aligned}
$$

Plugging the lower bound above into (\ref{bound6}) we finally obtain

$$
\begin{aligned}
\mathbb{P}\left\{\mid \frac{1}{N}\sum_{i=1}^{N} U_{i} g\left(z_{i}\right) \mid \geq \frac{\epsilon \alpha}{4}-\frac{\epsilon^{2} \alpha}{4 \mathcal{B}(1+\epsilon)}+\frac{\epsilon(1-\epsilon)}{4 \mathcal{B}(1+\epsilon)} \frac{1}{N} \sum_{i=1}^{N} g^{2}\left(z_{i}\right)\right\}\leq 2 \exp \left(-\frac{\epsilon^{2}(1-\epsilon) \alpha N}{7 \mathcal{B}(1+\epsilon)}\right) .
\end{aligned}
$$

{\noindent STEP 6. Bounding the covering number.}

In this step we construct an $\frac{\epsilon \beta}{7}$-cover of $\left\{g_{f}: f \in \mathcal{F}_N\right\}$ on $z_{1}, \ldots, z_{N}$. Let $f_{1}, \ldots, f_{l}, l=\mathcal{N}_{N}\left(\frac{\epsilon \beta}{7}, \mathcal{F}_N, \Vert\cdot\Vert_\infty\right)$ be an $\frac{\epsilon \beta}{7}$-cover of $\mathcal{F}_N$ on $z_{1}^{N}$. By assumption, $\Vert f_{j}\Vert_\infty \leq \mathcal{B}$ for all $j$. Let $f \in \mathcal{F}_N$ be arbitrary. Then there exists an $f_{j}$ such that $\frac{1}{N} \sum_{i=1}^{N}\left|f\left(x_{i}\right)-f_{j}\left(x_{i}\right)\right|<\frac{\epsilon \beta}{7}$. We have
$$
\begin{aligned}
\frac{1}{N} \sum_{i=1}^{N}\left|g_{f}\left(z_{i}\right)-g_{f_{j}}\left(z_{i}\right)\right| 
& \leq \frac{1}{N} \sum_{i=1}^{N}\left|f\left(x_{i}\right)-f_{j}\left(x_{i}\right)\right|<\frac{\epsilon \beta}{7},
\end{aligned}
$$
by the Lipschitz property of $g_f$ with respect to $f$. Thus $g_{f_{1}}, \ldots, g_{f_{l}}$ is an $\frac{\epsilon \beta}{7}$-cover of $\left\{g_{f}: f \in \mathcal{F}_N\right\}$ on $ S$ of size $\mathcal{N}_{N}\left(\frac{\epsilon \beta}{7}, \mathcal{F}_N, \Vert\cdot\Vert_\infty\right)$. Steps 3 through 6 imply
\begin{align*}
&{\small \mathbb{P}\left\{\exists f \in \mathcal{F}_N: \frac{1}{N} \sum_{i=1}^{N}\left(g_{f}\left(Z_{i}^{\prime}\right)-g_{f}\left(Z_{i}\right)\right)\geq \frac{\epsilon}{2}(\alpha+\beta)-\frac{\epsilon^{2}(\alpha+\beta)}{4\mathcal{B}(1+\epsilon)}+\frac{\epsilon(1-\epsilon)}{4\mathcal{B}(1+\epsilon)} \frac{1}{N} \sum_{i=1}^{N}\left(g_{f}^{2}\left(Z_{i}\right)+g_{f}^{2}\left(Z_{i}^{\prime}\right)\right)\right\} }\\
&\qquad\qquad\qquad\leq 4 \mathcal{N}_{N}\left(\frac{\epsilon \beta}{7}, \mathcal{F}_N, \Vert\cdot\Vert_\infty\right) \exp \left(-\frac{\epsilon^{2}(1-\epsilon) \alpha N}{7\mathcal{B}(1+\epsilon)}\right) .
\end{align*}

{\noindent STEP 7. Conclusion.}

Steps 1, 2, and 6 imply, for $N>\frac{64\mathcal{B}}{\epsilon^{2}(\alpha+\beta)}$,

$$
\begin{aligned}
& \mathbb{P}\left\{\exists f \in \mathcal{F}_N: \mathbb{E} g_{f}(Z)-\frac{1}{N} \sum_{i=1}^{N} g_{f}\left(Z_{i}\right)>\epsilon\left(\alpha+\beta+\mathbb{E} g_{f}(Z)\right)\right\} \\
& \leq \frac{32}{7} \mathcal{N}_{N}\left(\frac{\epsilon \beta}{7}, \mathcal{F}_N, \Vert\cdot\Vert_\infty\right) \exp \left(-\frac{\epsilon^{2}(1-\epsilon) \alpha N}{7\mathcal{B}(1+\epsilon)}\right) \\
& \quad+\frac{64}{7} \mathcal{N}_{N}\left(\frac{(\alpha+\beta) \epsilon}{5},\left\{g_{f}: f \in \mathcal{F}_N\right\},  \Vert\cdot\Vert_\infty \right) \exp \left(-\frac{3 \epsilon^{2}(\alpha+\beta) N}{40\mathcal{B}^2}\right) \\
& \leq 14 \mathcal{N}_{N}\left(\frac{\epsilon \beta}{7}, \mathcal{F}_N, \Vert\cdot\Vert_\infty \right) \exp \left(-\frac{3\epsilon^{2}(1-\epsilon) \alpha N}{40(1+\epsilon) \mathcal{B}^2}\right).
\end{aligned}
$$
For $N\le\frac{64\mathcal{B}}{\epsilon^{2}(\alpha+\beta)}$,  we have 
$$
\exp \left(-\frac{3\epsilon^{2}(1-\epsilon) \alpha N}{40(1+\epsilon) \mathcal{B}^2}\right) \geq \exp \left(-\frac{12}{5}\right) \ge \frac{1}{14}
$$
and hence the assertion follows trivially.

\subsection{Proof of Lemma \ref{app_error}.}
Recall that the target quantile curves $Q_Y^{\tau_1},\ldots,Q_Y^{\tau_K}$ are H\"older functions in $\mathcal{H}^\beta([0,1]^{d_0},B)$. On one hand, it is easy to show that the average of target quantile curves $\bar{Q}_Y= \sum_{k=1}^KQ_Y^{\tau_k}$ also belongs to $\mathcal{H}^\beta([0,1]^{d_0}, B)$. On the other hand, it is noted that the target function of value layer $v(\cdot;\theta_v)$ is just the averaged quantile curve $\bar{Q}_Y$. Then based on Theorem 3.3 in \cite{jiao2023deep}, for any positive integer $M,U$, there exists an ReLU network $v$ with width $\mathcal{W}=38(\lfloor\beta\rfloor+1)^2{d_0}^{\lfloor\beta\rfloor+1}U\lceil\log_2(8U)\rceil$ and depth $\mathcal{D}=21(\lfloor\beta\rfloor+1)^2M\lceil\log_2(8M)\rceil$ so that
\begin{equation*}
|v(x)- \bar{Q}_Y(x)|\leq 18B (\lfloor\beta\rfloor+1)^2d_0^{\lfloor\beta\rfloor+(\beta\vee 1)/2}(UM)^{-2\beta/d_0},
\end{equation*}
for $x\in[0,1]^{d_0}$ excluding a set $\Omega$ with Lebesgue measure $\delta K{d_0}$, where $\delta$ can be arbitrarily small. Therefore, we have
\begin{align}\label{expectation_approximation_bound}
\mathbb{E}|v(X)- \bar{Q}_Y(X)|&\leq 18B_0 (\lfloor\beta\rfloor+1)^2{d_0}^{\lfloor\beta\rfloor+(\beta\vee 1)/2}(UM)^{-2\beta/{d_0}}+\mathbb{P}(\Omega)\cdot \underset{x\in\Omega}{\sup}\{|v(x)- \bar{Q}_Y(x)|\}.
\end{align}
By Assumption (C1), the marginal distribution of $X$ is continuous with respect to the Lebesgue measure, which means that $\underset{\delta\rightarrow0}{\lim\sup} \mathbb{P}(\Omega)=0$. Meanwhile, we know from the 
assumption and the definition that both $\|v \|_{\infty}$ and $\|\bar{Q}_Y\|_{\infty}$ are bounded. Therefore, by taking limit infimum with respect to $\delta$ on both sides of \eqref{expectation_approximation_bound}, we have 
\begin{equation*}
\mathbb{E}|v(X)- \bar{Q}_Y(X)|\leq 18B (\lfloor\beta\rfloor+1)^2{d_0}^{\lfloor\beta\rfloor+(\beta\vee 1)/2}(UM)^{-2\beta/{d_0}}.
\end{equation*}

Next, we consider the approximation of the delta layer on the gaps of target quantile curves. Since the target quantile curves $Q_Y^{\tau_1},\ldots, Q_Y^{\tau_K}$ are H\"older functions in $\mathcal{H}^\beta([0,1]^{d_0}, B)$,  it is obvious that the difference of each pair of consecutive target quantile curves $\Delta^{k}_Y= Q_Y^{\tau_k}-Q_Y^{\tau_{k-1}}\in(0, 
B]$ and is $\beta$-H\"older smooth for $k=1,\ldots,K$. Then $g_Y^{(k)}:=ELU^{-1}(\Delta^{k}_Y-1)$ is the target function of the $k$th component of delta layer ouput $g_k(\cdot;\theta_\delta)$. Note that $g_Y^{(k)}=ELU^{-1}(\Delta^{k}_Y-1)$ may not be lower bounded, we thus focus on the approximation of its truncated version $\bar{g}_Y^{(k)}=\max\{g_Y^{(k)},-\mathcal{B}\}$ for $\mathcal{B}\ge B$. Then again by Theorem 3.3 in \cite{jiao2023deep}, any positive integer $M,N$, there exists an ReLU network $g_k$ with width $\mathcal{W}=38(\lfloor\beta\rfloor+1)^2{d_0}^{\lfloor\beta\rfloor+1}U\lceil\log_2(8U)\rceil$ and depth $\mathcal{D}=21(\lfloor\beta\rfloor+1)^2M\lceil\log_2(8M)\rceil$ such that
\begin{equation*}
\mathbb{E}\vert g_k(X)- \bar{g}_Y^{(k)}(X)\vert\leq 18\mathcal{B}(\lfloor\beta\rfloor+1)^2d_0^{\lfloor\beta\rfloor+(\beta\vee 1)/2}(UM)^{-2\beta/d_0},
\end{equation*}
for $k=1,\ldots,K$. Let $\sigma(x)=ELU(x)+1$, we can bound the approximation error of the $\sigma$-activated output of delta layer on the quantile curve gaps:
\begin{align*}
\mathbb{E}\vert \sigma(g_k(X))- \Delta^{k}_Y(X)\vert&\leq\mathbb{E}\vert \sigma(g_k(X))- \sigma(\bar{g}_Y^{(k)}(X))\vert +\mathbb{E}\vert\sigma(\bar{g}_Y^{(k)}(X))-\Delta^{k}_Y(X)\vert\\
&\leq 18\mathcal{B} (\lfloor\beta\rfloor+1)^2d_0^{\lfloor\beta\rfloor+(\beta\vee 1)/2}(UM)^{-2\beta/d_0} + \exp(-\mathcal{B}),
\end{align*}
where the second inequality follows since $\sigma(\cdot)$ is 1-Lipschitz and the fact that the difference between $\sigma(\bar{g}_Y^{(k)}(X))$ and $\Delta^{k}_Y(X)$ is bounded by $\exp(-\mathcal{B})-\exp(-\infty)=\exp(-\mathcal{B})$ almost surely.

Last, we can parallelly stack the neural networks $v,g_1,\ldots,g_K$ to calculate the neural network $f=(f_1,\ldots,f_K)$ according to (\ref{nq}), and we have
\begin{align*}
\mathbb{E}\vert f_k(X)- Q^{\tau_k}_Y(X)\vert&\leq
18(K+2)\mathcal{B} (\lfloor\beta\rfloor+1)^2d_0^{\lfloor\beta\rfloor+(\beta\vee 1)/2}(UM)^{-2\beta/d_0} + (K+2)\exp(-\mathcal{B}),
\end{align*}
for $k=1,\ldots,K$,  where the neural networks in $\mathcal{F}_N$ has width $\mathcal{W}=38(K+1)(\lfloor\beta\rfloor+1)^2{d_0}^{\lfloor\beta\rfloor+1}U\lceil\log_2(8U)\rceil$ and depth $\mathcal{D}=21(\lfloor\beta\rfloor+1)^2M\lceil\log_2(8M)\rceil$ for any positive integer $M,U$. Lastly, by the Lipschitz property of the quantile check loss $\rho_{\tau_k}$, and the definition of the risk $\mathcal{R}$, it is easy to show
\begin{align*}
\underset{f\in\mathcal{F}_{N}}{\inf}\mathcal{R}(f)\leq 18(K+2)\mathcal{B} (\lfloor\beta\rfloor+1)^2d_0^{\lfloor\beta\rfloor+(\beta\vee 1)/2}(UM)^{-2\beta/d_0} + (K+2)\exp(-\mathcal{B}),
\end{align*}
which completes the proof.

\subsection{Proof of Lemma \ref{app_error1}.}
The Lemma \ref{app_error1} can be proved by following the proof of Lemma \ref{app_error}. With the same construction of neural network $f=(f_1,\ldots,f_K)$ in the proof of Lemma \ref{app_error}, and we have
\begin{align*}
\mathbb{E}\vert f_k(X)- Q^{\tau_k}_Y(X)\vert^2&\leq
648\mathcal{B}^2 (K+2)^2(\lfloor\beta\rfloor+1)^4d_0^{2\lfloor\beta\rfloor+(\beta\vee 1)}(UM)^{-4\beta/d_0} + 2(K+2)^2\exp(-2\mathcal{B}),
\end{align*}
for $k=1,\ldots,K$,  where the neural networks in $\mathcal{F}_N$ has width $\mathcal{W}=38(K+1)(\lfloor\beta\rfloor+1)^2{d_0}^{\lfloor\beta\rfloor+1}U\lceil\log_2(8U)\rceil$ and depth $\mathcal{D}=21(\lfloor\beta\rfloor+1)^2M\lceil\log_2(8M)\rceil$ for any positive integer $M,U$. Lastly, by the local quadratic property in Assumption \ref{assump2} of the excess risk $\mathcal{R}$, it is easy to show
\begin{align*}
\underset{f\in\mathcal{F}_{N}}{\inf}\mathcal{R}(f)\leq 648(K+2)^2\mathcal{B}^2(\lfloor\beta\rfloor+1)^4d_0^{2\lfloor\beta\rfloor+(\beta\vee 1)}(UM)^{-4\beta/d_0} + 2(K+2)^2\exp(-2\mathcal{B}),
\end{align*}
which completes the proof.

\subsection{Proof of Theorem \ref{nonasymp_error}.}
Combining Lemma \ref{decomposition}, Lemma \ref{app_error1} and Theorem \ref{sto_error}, the bounds directly follows:
\begin{align}\label{first_error_term}
\mathbb{E}[\mathcal{R}(\hat{f}_N)]&\le C_2\cdot K\mathcal{B}^3\frac{\mathcal{S}\mathcal{D}\log(\mathcal{S})\log(N)}{N}\\\label{second_error_term}
&+C_1(K+2)^2\left[\mathcal{B}^2(\lfloor\beta\rfloor+1)^4d_0^{2\lfloor\beta\rfloor+(\beta\vee 1)}(UM)^{-4\beta/d_0} + \exp(-2\mathcal{B})\right]
\end{align}
where $N\geq C\cdot \mathcal{S}\mathcal{D}
\log(\mathcal{S})$ for a large enough $C>0$ and $M,U\in\mathbb{N}_+$. 

\subsection{Proof of Corollary \ref{cor1}}
To balance these two error terms, we choose proper integers $M$ and $U$ with respect to sample size $N$ to optimize the convergence rate. We notice that the network size $\mathcal{S}$ grows linear in its depth $\mathcal{D}$ but quadratically in width $\mathcal{W}$. To reach the optimal rate, and at the same time, to use a smaller network size, we fix $U=1$ so that the width $\mathcal{W}=114(K+1)(\lfloor\beta\rfloor+1)^2{d_0}^{\lfloor\beta\rfloor+1}$. We set $M=\lfloor N^t\rfloor$ with $t>0$ to be determined later such that  $\mathcal{D}=21(\lfloor\beta\rfloor+1)^2M\lceil\log_2(8M)\rceil=O(N^t\log(N))$
and
\begin{align*}
\mathcal{S}&\leq \mathcal{W}(d_0+1)+(\mathcal{W}^2+\mathcal{W})(\mathcal{D}-1)+\mathcal{W}+1=O(\mathcal{W}^2\mathcal{D})=O(N^t\log(N)).
\end{align*}
Neglecting the $\log(N)$ factors, and choosing the truncation bound $\mathcal{B}=\log(N)$ in \eqref{second_error_term}, we then check the orders of the stochastic error and approximation error in the excess risk bound:
\begin{equation*}
\eqref{first_error_term}=O(\frac{\mathcal{S}\mathcal{D}
}{N})=O(N^{2t-1}) \quad\text{ and }\quad \eqref{second_error_term}=O((MU)^{-4\beta/d_0})=O(N^{-4t\beta/d_0}).
\end{equation*}
Let $2t-1=-4t\beta/d_0$, we obtain $t=\frac{d_0}{2d_0+4\beta}$. Plugging it back into the relevant expressions above and letting $C$ be a generic constant whose value may change in different scenarios, we get
\begin{equation*}
\mathcal{D}\leq C\frac{d_0}{2d_0+4\beta}(\lfloor\beta\rfloor+1)^2N^{\frac{d_0}{2d_0+4\beta}}\log(N)\leq C(\lfloor\beta\rfloor+1)^2N^{\frac{d_0}{2d_0+4\beta}}\log(N)
\end{equation*}
and 
\begin{equation*}
\mathcal{S}\leq C\mathcal{W}^2\mathcal{D}\leq C(\lfloor\beta\rfloor+1)^6 {d_0}^{2(\lfloor\beta\rfloor+1)}N^{\frac{d_0}{2d_0+4\beta}}\log(N).
\end{equation*}
Therefore,
\begin{align*}
\mathbb{E}[\mathcal{R}(\hat{f}_N)]\le C\cdot K^2(\log N)^7 N^{-\frac{2\beta}{d_0+2\beta}}
\end{align*}
for $N$ large enough where $C>0$ is a constant depending only on $\beta$ and polynomially in $d_0$. Finally, we notice that the stochastic error bound is valid only when $N\geq C\cdot \mathcal{S}\mathcal{D}\log(\mathcal{S})$. Under the current setting of $\mathcal{S}$ and $\mathcal{D}$, we know that $\mathcal{S}\mathcal{D}\log(\mathcal{S})$ grows slower than $N$. This condition then holds when $N$ is large enough.

\subsection{Proof of Lemma \ref{app_error_lowdim}}
We prove the lemma following the ideas in \cite{shen2019deep,jiao2023deep}. We first project the data to a low-dimensional space and then construct an NQ network to approximate the low-dimensional function. 

Based on Theorem 3.1 in \cite{baraniuk2009random}, there exists a linear projector $A\in\mathbb{R}^{d_0^*\times d_0}$ with $d_0^*\le d_0$ that maps a low-dimensional manifold in a $d_0$-dimensional space to a $d_0^*$-dimensional space nearly preserving the distance. To be specific, there exists a matrix $A\in\mathbb{R}^{d_0^*\times d_0}$ such that
$AA^T=(d_0/d_0^*)I_{d_0^*}$ where $I_{d_0^*}$ is an identity matrix of size $d_0^*\times d_0^*$, and
$$(1-\delta)\Vert x_1-x_2\Vert_2\leq\Vert Ax_1-Ax_2\Vert_2\leq(1+\delta)\Vert x_1-x_2\Vert_2,$$
for any $x_1,x_2\in\mathcal{M}.$ And it is easy to check
$$A(\mathcal{M}_\rho)\subseteq A([0,1]^{d_0})\subseteq E:=[-\sqrt{{d_0}/{d_0^*}},\sqrt{{d_0}/{d_0^*}}]^{d_0^*}.$$

Since $A$ nearly preserves the distance on $\mathcal{M}$, it is easy to see that for any $z\in A(\mathcal{M})$, there exists a unique $x\in\mathcal{M}$ such that $Ax=z$. For any $z\in A(\mathcal{M})$, we define $x_z=\mathcal{SL}(\{x\in\mathcal{M}: Ax=z\})$ where $\mathcal{SL}(\cdot)$ returns a unique element of a set. And we can see that $\mathcal{SL}: A(\mathcal{M})\to\mathcal{M}$ is a differentiable function with the norm of its derivative bounded in $[1/(1+\delta),1/(1-\delta)]$, because
$$\frac{1}{1+\delta}\Vert z_1-z_2\Vert_2\leq\Vert x_{z_1}-x_{z_2}\Vert_2\leq\frac{1}{1-\delta}\Vert z_1-z_2\Vert_2,$$
for any $z_1,z_2\in A(\mathcal{M})\subseteq E$.
For the high-dimensional target function $Q_Y: [0,1]^{d_0}\to\mathbb{R}^K$, we define its low-dimensional representation $\tilde{Q}_Y:\mathbb{R}^{d_0^*}\to\mathbb{R}^K$ by
$$\tilde{Q}_Y(z)=Q_Y(x_z), \quad {\rm for\ any} \ z\in A(\mathcal{M})\subseteq\mathbb{R}^{d_0^*}.$$
By assumption,  $Q_Y^{\tau_k}\in\mathcal{H}^\beta([0,1]^{d_0},B)$ for $k=1,\ldots,K$, then $\tilde{Q}_Y^{\tau_k}\in\mathcal{H}^\beta(A(\mathcal{M}),B/(1-\delta)^\beta)$. By the extended version of Whitney' extension theorem in \cite{fefferman2006whitney}, functions $\tilde{Q}_Y^{\tau_k}$ for $k=1,\ldots,K$ can be extended such that $\tilde{Q}_Y^{\tau_k}\in\mathcal{H}^\beta(E,B/(1-\delta)^\beta)$ and $\tilde{Q}_Y(z)=Q_Y(x_z)$ for any $z\in A(\mathcal{M})$. 

For H\"older smooth function $\tilde{Q}_Y$ defined on low-dimentional set $E=[-\sqrt{{d_0}/{d_0^*}},\sqrt{{d_0}/{d_0^*}}]^{d_0^*}$, based on Lemma \ref{app_error}, for any positive integer $M,U$, there exists an NQ network $\tilde{f}=(\tilde{f}^{\tau_1},\ldots,\tilde{f}^{\tau_K})$ with width $\mathcal{W}=38(K+1)(\lfloor\beta\rfloor+1)^2{d_0^*}^{\lfloor\beta\rfloor+1}U\lceil\log_2(8U)\rceil$ and depth $\mathcal{D}=21(\lfloor\beta\rfloor+1)^2M\lceil\log_2(8M)\rceil$ so that
\begin{equation*}
|\tilde{f}^{\tau_k}(z)- \tilde{Q}^{\tau_k}_Y(z)|\leq 18(K+2)\mathcal{B} (\lfloor\beta\rfloor+1)^2(d_0^*)^{\lfloor\beta\rfloor+(\beta\vee 1)/2}(UM)^{-2\beta/(d_0^*)} + (K+2)\exp(-\mathcal{B}),
\end{equation*}
for all $z\in E\backslash\Omega(E)$ where $\Omega(E)$ is a subset of $E$ with an arbitrarily small Lebesgue measure as well as $\Omega:=\{x\in\mathcal{M}_\rho: Ax\in\Omega(E)\}$ does.

Then we construct a network $f=\tilde{f}\circ A$ such that $f(x)=\tilde{f}(Ax)$ for any $x\in[0,1]^{d_0}$, then $f\in\mathcal{F}_N$ is also NQ network with one more linear layer  than $\tilde{f}$. Then for $k=1,\ldots,K$, and any $x\in\mathcal{M}_\rho\backslash\Omega$ and $z=Ax$, , there exists a $\tilde{x}\in\mathcal{M}$ such that $\Vert x-\tilde{x}\Vert_2\leq \rho$ by definition of neighborhood, and we have
\begin{align*}
&\vert f^{\tau_k}(x)-Q^{\tau_k}_Y(x)\vert=\vert \tilde{f}^{\tau_k}(Ax)-\tilde{Q}^{\tau_k}_0(Ax)+\tilde{Q}^{\tau_k}_0(Ax)-\tilde{Q}^{\tau_k}_0(A\tilde{x})+\tilde{Q}^{\tau_k}_0(A\tilde{x})-{Q}^{\tau_k}_0(x)\vert\\
&\leq\vert \tilde{f}^{\tau_k}(Ax)-\tilde{Q}^{\tau_k}_0(Ax)\vert+\vert\tilde{Q}^{\tau_k}_0(Ax)-\tilde{Q}^{\tau_k}_0(A\tilde{x})\vert+\vert\tilde{Q}^{\tau_k}_0(A\tilde{x})-{Q}^{\tau_k}_0(x)\vert\\
&\leq \vert \tilde{f}^{\tau_k}(Ax)-\tilde{Q}^{\tau_k}_0(Ax)\vert+\frac{B}{1-\delta}\Vert Ax-A\tilde{x}\Vert_2+\vert+\vert{Q}^{\tau_k}_0(\tilde{x})-{Q}^{\tau_k}_0(x)\vert\\
&= \vert \tilde{f}^{\tau_k}(Ax)-\tilde{Q}^{\tau_k}_0(Ax)\vert+\frac{\rho B}{1-\delta}\sqrt{\frac{d_0}{d_0^*}}+\rho B\\
&\le(K+2)[18\mathcal{B} (\lfloor\beta\rfloor+1)^2(d_0^*)^{\lfloor\beta\rfloor+(\beta\vee 1)/2}(UM)^{-2\beta/(d_0^*)} +\exp(-\mathcal{B})]+\rho B\{(1-\delta)^{-1}\sqrt{{d_0}/{d_0^*}}+1\}\\
&\leq (18+C_2)(K+2)\mathcal{B} (\lfloor\beta\rfloor+1)^2(d_0^*)^{\lfloor\beta\rfloor+(\beta\vee 1)/2}(UM)^{-2\beta/(d_0^*)} +(K+2)\exp(-\mathcal{B})
\end{align*}
where $C_2>0$ is a constant not depending on any parameter. The last inequality follows from $\rho\leq C_2(UM)^{-2\beta/d_0^*}(\beta+1)^2(d_0^*)^{3\beta/2}\{(1-\delta)^{-1}\sqrt{{d_0}/{d_0^*}}+1\}^{-1}$. Since  the probability measure of $X$ is absolutely continuous with respect to the Lebesgue measure, we have
\begin{align*}
\mathbb{E}\Vert f^{\tau_k}(X) -Q_Y^{\tau_k}(X)\Vert^2 &\leq (18+C_2)^2(K+2)^2\mathcal{B}^2(\lfloor\beta\rfloor+1)^4(d_0^*)^{2\lfloor\beta\rfloor+(\beta\vee 1)}(UM)^{-4\beta/(d_0^*)}\\
&\qquad\qquad\qquad\qquad+(K+2)^2\exp(-2\mathcal{B}),
\end{align*}
where $d_0^*=O(d_\mathcal{M}{\log(d/\delta)}/{\delta^2})$ is assumed
to satisfy $d_0^* \ll d$. Then by Assumption \ref{assump2}, the local quadratic property of excess risk completes the proof.

\subsection{Proof of Theorem \ref{theorem_rl}}

One may expect to obtain a convergence result for the iteration of the distributional RL algorithm as in Algorithm \ref{alg-1}. However, it is shown that the Bellman optimality operator $\mathcal{T}$ is in general not a contraction, and not all optimality operators (with respect to optimal policies) have a fixed point. Even worse, the optimality operator $\mathcal{T}$ has a fixed point that is insufficient to guarantee the convergence of $Z^{(M)}$ towards optimal value distributions $\mathcal{Z}^*$.

In fact, we can establish the error propagation results for the distributional RL iterations as \citep{farahmand2010error}, which relates our target to the maximum estimation errors for each iteration.
\begin{customthm}{S2}[Error Propagation]\label{propagation} Let $\{Z^{(m)}\}_{m=0}^M$ be the iterates in Algorithm \ref{alg-1}. Let $\pi_M$ be the greedy policy with respect to $Z^{(M)}$, and let $Z^{\pi_M}$ be the action-value distribution corresponding to $\pi_M$. Then for $\mu\in\mathcal{P}(\mathcal{S}\times\mathcal{A})$ being a distribution on $\mathcal{S}\times\mathcal{A}$, we have
\begin{align}
	\Vert \mathbb{E} Z^{\pi_M}-\mathbb{E}Z^*\Vert_{1,\mu}\le c_{M,\mu}\frac{2\gamma}{(1-\gamma)^2}\cdot \max_{m=1,\ldots,M}\Vert \mathbb{E}\mathcal{T}Z^{(m-1)}-\mathbb{E} Z^{(m)}\Vert_{\sigma_m} +\frac{4\gamma^{M+1}}{(1-\gamma)^2} C_{p,R}
\end{align}
where $c_{M,\mu}$ is a constant that only depends on the sampling distribution $\mu$ and $\sigma_1,\ldots,\sigma_M$ (the distributions of state-action pairs $(S^{(m)},A^{(m)}),m=1,\ldots,M$). The definition of $c_{M,\mu}$ can be found in Definition \ref{coeff}.
\end{customthm}
We defer the proof of Lemma \ref{propagation} to the next subsection. Lemma \ref{propagation} indicates that the prediction error for the expected action-value distribution of Algorithm \ref{alg-1} can be bounded by the estimation errors $\Vert \mathcal{T}Z^{(m-1)}-Z^{(m)}\Vert_{\sigma_m}$ and the algorithmic error $4\gamma^{M+1}\cdot C_{R}/(1-\gamma)^2$, which is exponentially decreasing in the number of iterations. Notably, the algorithmic error in Lemma \ref{propagation} depends on the first moment of the reward and does not require the reward to be bounded, which improves the existing error propagation results for fitted value iteration algorithm \cite{munos2008finite,farahmand2010error,scherrer2015approximate,farahm2016regularized,fan2020theoretical,li2022testing}.

With Lemma \ref{propagation}, we start to prove the theorem. Given policy  $\pi^*$, we let $\mu_{\pi^*}$ denote the distribution of $(S_0,A_0)$ such that $(A_0|S_0)$ follows distribution $\pi^*(\cdot|S_0)$. We let $\mu_{\pi_M}$ denote the the distribution of $(S_0,A_0)$ with respect to policy $\pi_M$.
Notice that
\begin{align*}
J(\pi^*)-J(\pi_M)&=\mathbb{E}[\pi^*(a,S_0)Z^*(S_0,a)]-\mathbb{E}[\pi^M(a,S_0)Z^{\pi_M}(S_0,a)]\\
&\le \left\vert\mathbb{E}[\pi^*(a,S_0)Z^*(S_0,a)]-\mathbb{E}[\pi^*(a,S_0)Z^{\pi_M}(S_0,a)]\right\vert \\
&+\left\vert\mathbb{E}[\pi^{M}(a,S_0)Z^*(S_0,a)]-\mathbb{E}[\pi^M(a,S_0)Z^{\pi_M}(S_0,a)]\right\vert\\
&\le\left\|\mathbb{E}Z^*-\mathbb{E}Z^{\pi_M}\right\|_{1, \mu_{\pi^*}}+\left\|\mathbb{E}Z^*-\mathbb{E}Z^{\pi_M}\right\|_{1, \mu_{\pi_M}}.
\end{align*}
Then by Lemma \ref{propagation}, the main part of the proof of the theorem is to establish the upper bounds of the estimation error $\Vert \mathbb{E}\mathcal{T}Z^{(m-1)}-\mathbb{E} Z^{(m)}\Vert_{1,\sigma_m}$ for $m=1,\ldots, M$. It is worth noting that $\mathbb{E}\mathcal{T}Z^{(m-1)}$ and $\mathbb{E} Z^{(m)}$ are the expectations of the value distribution, while our estimates $Z^{(m)}=(Z^{(m)}_1,\ldots,Z^{(m)}_K)$ are the conditional quantiles at level $1/(K+1),\ldots, K/(K+1)$. In light of this, we first build up the connection between the distribution mean and the average of $K$ equal-spaced quantiles. Let $\mathbb{E}_K Z^{(m)}:=\sum_{k=1}^K Z^{(m)}_k$ denote the empirical mean and let $\mathbb{E}_K \mathcal{T}Z^{(m-1)}:=\sum_{k=1}^K (\mathcal{T}Z^{(m-1)})_k$. To this end, we prove the moment condition of $Z^\pi$ for any $\pi$ to apply Lemma \ref{lemma_K}. Recall that for a fixed policy $\pi$ and a given initial state-action pair $(s,a)$, its return 
$$Z^\pi(s,a)=\sum_{t=0}^\infty\gamma^t R(S_t,A_t),$$
is a random variable representing the sum of discounted rewards observed along a trajectory $\{(S_t, A_t)\}_{t\ge 0}$ following $\pi$, conditional on that $S_0=s$ and $A_0=a$. By Fatou's Lemma and Minkowski's inequality, we have
\begin{align*}
&[\mathbb{E}\vert Z^\pi(s,a)\vert^p]^{1/p}\le\sum_{t=0}^\infty (\mathbb{E}\vert\gamma^tR(S_t,A_t)\vert^p)^{1/p}\le \sum_{t=0}^\infty \gamma^t C_{p,R}\le \frac{C_{p,R}}{1-\gamma}.
\end{align*}
Then by Lemma \ref{lemma_K}, we have
$$\Vert\mathbb{E}_K Z^{(m)}-\mathbb{E} Z^{(m)}\Vert_{\sigma_m}\le \frac{C_1\times C_{p,K}}{(1-\gamma)K^{(p-1)/p}},\quad{\rm and}\quad \Vert \mathbb{E}_K \mathcal{T}Z^{(m-1)}-\mathbb{E}\mathcal{T}Z^{(m-1)}\Vert_{\sigma_m}\le \frac{C_2\times C_{p,K}}{(1-\gamma)K^{(p-1)/p}}$$
for some universal constants $C_1,C_2>0$.
Then we focus on deriving the bounds for
$$\Vert \mathbb{E}_K\mathcal{T}Z^{(m-1)}-\mathbb{E}_K Z^{(m)}\Vert_{\sigma_m}\le \frac{1}{K}\sum_{k=1}^K\Vert \mathcal{T}Z^{(m-1)}_k-Z^{(m)}_k\Vert_{\sigma_m},$$
for each iteration. In the $m$th iteration, $\mathcal{T}Z^{(m-1)}_k$ is the target function of the $\tau_k$ conditional quantile of the propagated value distribution $\mathcal{T}Z^{(m-1)}$, the function $Z^{(m)}_k$ is an estimator of $\mathcal{T}Z^{(m-1)}$ by minimizing the empirical risk for non-crossing quantile networks. Then by the calibration condition in Assumption \ref{assump_RL2}, in the rest of the proof it is sufficient to give upper bounds for the excess risk $\mathcal{R}^{(m)}$ where 
$$\mathcal{R}^{(m)}(f):=\mathbb{E}_{(S,A)\sim\sigma_m}\frac{1}{K}\sum_{k=1}^K[\rho_{\tau_k}(\mathcal{T}Z^{(m-1)}(S,A)-f_k(S,A))-\rho_{\tau_k}(\mathcal{T}Z^{(m-1)}(S,A)-\mathcal{T}Z^{(m-1)}_k(S,A))],$$
for $f\in\mathcal{F}^{(RL)}_N$ and $Z^{(m)}=(Z^{(m)}_1,\ldots,Z^{(m)}_K)$ is the estimation in the $m$th iteration of the fitted NQ algorithm, and $\mathcal{T}Z^{(m-1)}(S,A)$ is a random variable following the value distribution $\mathcal{T}Z^{(m-1)}$ given state $S$ and action $A$.

Note that the estimation $Z^{(m)}$ is actually the minimizer of the empirical risk $\mathcal{R}^{(m)}_N$ in the $m$th iteration with respect to the sample $\{(S^{(m)}_i,A^{(m)}_i,R^{(m)}_i)\}_{i=1}^N$. Then by the error decomposition as in Lemma \ref{decomposition},  for each $m=1,\ldots,M$ we have
\begin{align*}
\mathcal{R}^{(m)}(Z^{(m)})\le 2\sup_{f\in\mathcal{F}_N^{(RL)}}\vert \mathcal{R}^{(m)}(f)-\mathcal{R}^{(m)}_N(f)\vert +\inf_{f\in\mathcal{F}_N^{(RL)}}\mathcal{R}^{(m)}(f),
\end{align*}
where $\sup_{f\in\mathcal{F}_N^{(RL)}}\vert \mathcal{R}^{(m)}(f)-\mathcal{R}^{(m
)}_N(f)\vert$ is the {\it stochastic error} and $\inf_{f\in\mathcal{F}_N^{(RL)}}[\mathcal{R}^{(m)}(f)]$ is the {\it approximation error}. Next, we bound the {\it stochastic error} and the {\it approximation error} in each iteration $m=1,\ldots, M$.

To analyze the stochastic error, we apply McDiarmid’s inequality for dependent data in our theory. We denote $D^{(m)}:=\{(S^{(m)}_i,A^{(m)}_i,R^{(m)}_i)\}_{i=1}^N$ by the sequence of training data which is possibly dependent and let $\tilde{D}^{(m)}$ be another data that is different from $D^{(m)}$ by only one sample. For instance, $\tilde{D}^{(m)}$ can be the same as $D^{(m)}$ but only for the $j$th sample $(S^{(m)}_j,A^{(m)}_j,R^{(m)}_j)$ for some $j=1,\ldots,N$. Recall that the empirical risk $\mathcal{R}_N^{(m)}$ is defined with respect to $D^{(m)}$ and we let $\mathcal{R}_N^{(m)\prime}$ be the empirical risk defined with respect to $\tilde{D}^{(m)}$. Then by the Lipschitz property of the quantile check loss function and the bound assumption in Assumption \ref{assump_RL1}, it is easy to check that
$$\vert\mathcal{R}_N^{(m)}(f)-\mathcal{R}_N^{(m)\prime}(f)\vert\le 2\mathcal{B},$$
for any $f\in\mathcal{F}^{(RL)}_N.$ In addition, it is also easy to check that $\mathbb{E}R^{(m)}_N(f)= R^{(m)}(f)$ since the data $D^{(m)}$ is stationary. Then we apply McDiarmid’s inequality (Theorem 1 in \cite{lauer2023uniform}) for the dependent data $D^{(m)}$ to get
$$\mathbb{P}(\vert\mathcal{R}^{(m)}(f)-\mathcal{R}_N^{(m)}(f)\vert\ge t)\le 2\exp\left( -\frac{Nt^2}{2\mathcal{B}^2}\right),$$
for any given $f\in\mathcal{F}^{(RL)}_N$ and any $t>0$.

For any given $\epsilon>0$, let $f_1,f_2,\ldots,f_{\mathcal{N}}$ be the anchor points of an $\epsilon$-covering for the function class $\mathcal{F}_{N}^{(RL)}$ where $\mathcal{N}:=\mathcal{N}_N(\epsilon,\mathcal{F}_{N}^{(RL)},\Vert\cdot\Vert_\infty)$ denotes the covering number of $\mathcal{F}_{N}^{(RL)}$ with radius $\epsilon$ under the norm $\Vert\cdot\Vert_\infty$. It is worth noting that the definition of the covering number is uniform and irrelevant to the dependence of the data. Then for any $f\in\mathcal{F}_{N}^{(RL)}$, there exists an anchor $f_j$ for $j\in\{1,\ldots,\mathcal{N}\}$ such that 
$\vert f_j(s,a)-f(s,a)\vert\le \epsilon$ for any $(s,a)\in\{(s^{(m)}_i,a^{(m)}_i)\}_{i=1}^N$ regardless the dependence of $\{(s^{(m)}_i,a^{(m)}_i)\}_{i=1}^N$. Lipschitz property of quantile check loss further implies 
\begin{align}
\vert\mathcal{R}^{(m)}(f)-\mathcal{R}^{(m)}_N(f)\vert
&\le \vert\mathcal{R}^{(m)}(f_j)-\mathcal{R}^{(m)}_N(f_j)\vert+2\epsilon
\end{align}
Therefore, with a fixed $t>0$,
\begin{align}
\mathbb{P}\Big(\sup_{f\in\mathcal{F}_{N}^{(RL)}}\vert\mathcal{R}^{(m)}(f)-\mathcal{R}^{(m)}_N(f) \vert&\ge t+2\epsilon \Big){\leq} \mathbb{P}\Big(\exists j\in\{1,\ldots,\mathcal{N}\} : \vert\mathcal{R}^{(m)}(f_j)-\mathcal{R}^{(m)}_N(f_j)\vert\ge t\Big)\notag\\
&{\leq}\mathcal{N}_N(\epsilon,\mathcal{F}_{N}^{(RL)},\Vert\cdot\Vert_\infty)\max_{j\in\{1,\ldots,\mathcal{N}\}}\mathbb{P}\Big( \vert\mathcal{R}^{(m)}(f_j)-\mathcal{R}^{(m)}_N(f_j)\vert\ge t\Big)\notag\\
&{\leq}2\mathcal{N}_N(\epsilon,\mathcal{F}_{N}^{(RL)},\Vert\cdot\Vert_\infty)\exp\left( -\frac{Nt^2}{2\mathcal{B}^2}\right).
\end{align}

Note that for any random variable $W$, we have $$\mathbb{E}[W]\le \mathbb{E}[W\times I(W>0)]\le \int_{t=0}^\infty \mathbb{P}(W>t).$$
Then taking $\epsilon=1/N$ and following the proof of Lemma \ref{sto_error}, we can show that 
\begin{align*}
\mathbb{E}\sup_{f\in\mathcal{F}_N^{(RL)}}\vert \mathcal{R}^{(m)}(f)-\mathcal{R}^{(m
	)}_N(f)\vert&\leq \frac{8\mathcal{B}^2\sqrt{\log 2\mathcal{N}_N(1/N,\mathcal{F}_{N}^{(RL)},\Vert\cdot\Vert_\infty)}}{\sqrt{N}}
	\end{align*}
	for $m=1,\ldots,M$. Note that the covering number in above bound $\mathcal{N}_N(1/N,\mathcal{F}_{N}^{(RL)},\Vert\cdot\Vert_\infty)$ is for the neural networks class $\mathcal{F}_{N}^{(RL)}$ where for each function $f\in\mathcal{F}_{N}^{(RL)}$ and action $a\in\mathcal{A}$, $f(\cdot,a)\in\mathcal{F}_N$ is a NQ network with depth $\mathcal{D}$, width $\mathcal{W}$ and size $\mathcal{S}$ and bound $\mathcal{B}$. By our assumption the action space $\mathcal{A}$ is finite, it is easy to see that $\mathcal{N}_N(1/N,\mathcal{F}_{N}^{(RL)},\Vert\cdot\Vert_\infty)\le \mathcal{N}_N(1/N,\mathcal{F}_{N},\Vert\cdot\Vert_\infty)^{\vert\mathcal{A}\vert}$.  Then by Lemma \ref{thm:covering_number} and Theorems 3 and 6 of \cite{bartlett2019nearly}, we have 
	\begin{align*}
\mathbb{E}\sup_{f\in\mathcal{F}_N^{(RL)}}\vert \mathcal{R}^{(m)}(f)-\mathcal{R}^{(m
	)}_N(f)\vert&\leq \frac{8\mathcal{B}^2\sqrt{\vert\mathcal{A}\vert K\log(\mathcal{D})\mathcal{S}\log(\mathcal{S})}}{\sqrt{N}}
	\end{align*}
	for $N\geq C_1\cdot \mathcal{S}\mathcal{D}\log(\mathcal{S})$ and $m=1,\ldots,M$ where $C_1>0$ is some universal constant.
	
	To obtain an approximation error bound, we follow the proof of Lemma \ref{app_error}. We see that 
	\begin{align*}
\inf_{f\in\mathcal{F}^{(RL)}_N}\mathcal{R}^{(m)}(f)&\le \inf_{f\in\mathcal{F}^{(RL)}_N}\frac{1}{K}\sum_{k=1}^K \Vert \mathcal{T}Z^{(m-1)}_k-f_k\Vert_{1,\sigma_m}\\
&\le \inf_{f\in\mathcal{F}^{(RL)}_N}\frac{1}{K}\sum_{k=1}^K \sum_{a\in\mathcal{A}} \mathbb{E}_{S\sim\sigma_{m,S}} \Vert \mathcal{T}Z^{(m-1)}_k(S,a)-f_k(S,a)\Vert,\\
&\le \inf_{f\in\mathcal{F}^{(RL)}_N}\frac{1}{K}\sum_{k=1}^K \vert\mathcal{A}\vert \max_{a\in\mathcal{A}}\mathbb{E}_{S\sim\sigma_{m,S}} \Vert \mathcal{T}Z^{(m-1)}_k(S,a)-f_k(S,a)\Vert,
\end{align*}
where $\sigma_{m,S}$ denotes the marginal distribution of $S$ and $(f_1(\cdot,a),\ldots, f_K(\cdot,a))$ is a NQ neural network defined in \ref{NQ-networks} with width $\mathcal{W}$ and depth $\mathcal{D}$. Then by Theorem 3.3 in \cite{jiao2023deep} and following the proof of Lemma \ref{app_error}, there exists a network $f=(f_1,\ldots,f_K)\in\mathcal{F}_N^{(RL)}$ such that for $k=1,\ldots,K$
\begin{align*}
&\max_{a\in\mathcal{A}} \mathbb{E}_{S\sim\sigma_{m,S}} \Vert \mathcal{T}Z^{(m-1)}_k(S,a)-f_k(S,a)\Vert\\
&\qquad\leq
18(K+2)\mathcal{B} (\lfloor\beta\rfloor+1)^2d_0^{\lfloor\beta\rfloor+(\beta\vee 1)/2}(VM)^{-2\beta/d_0} + (K+2)\exp(-\mathcal{B}),
\end{align*}
where the neural networks $f(\cdot,a)=(f_1(\cdot,a),\ldots,f_K(\cdot,a))\in\mathcal{F}_N$ has width $\mathcal{W}=38(K+1)(\lfloor\beta\rfloor+1)^2{d_0}^{\lfloor\beta\rfloor+1}U\lceil\log_2(8U)\rceil$ and depth $\mathcal{D}=21(\lfloor\beta\rfloor+1)^2V\lceil\log_2(8V)\rceil$ for any positive integer $V,U$. 
Setting $U=1$, $V=\lfloor N^{d_0/[d_0+4\beta]}\rfloor$, $\mathcal{B}=\log(N)$, $\mathcal{S}=C(K+1)^{-2}(\lfloor\beta\rfloor+1)^6{d_0}^{2\lfloor\beta\rfloor+2}\lfloor N^{d_0/[d_0+4\beta]}\rfloor$ and combining above inequalities completes the proof.

\subsection{Proof of Lemma \ref{propagation}}

The proof can be proceeded in three steps following the idea of the proof of Theorem 6.1 in \cite{fan2020theoretical}. Before we present the proof, we introduce some notation. Recall that, for any $m \in\{0, \ldots, M\}$, the $Z^{(m)}$ denotes the quantile estimates of state-action value distribution. With a bit of abuse of notation, we let $Z^{m+1}:=\mathcal{T} Z^{(m)}$ denote the push-forward distribution of $\mathcal{T}$ operating on the distribution of $Z^{(m)}$ and let $\varrho_{m}=Z^{m}-Z^{(m)}$. 
We denote by $\pi_{m}$ the greedy policy with respect to $Z^{(k)}$. In addition, throughout the proof, for two functions $\mathbb{E}Z_{1}, \mathbb{E}Z_{2}: \mathcal{S} \times \mathcal{A} \rightarrow \mathbb{R}$, we use the notation $\mathbb{E}Z_{1} \geq \mathbb{E}Z_{2}$ if $\mathbb{E}Z_{1}(s, a) \geq \mathbb{E}Z_{2}(s, a)$ for any $s \in \mathcal{S}$ and any $a \in \mathcal{A}$, and define $\mathbb{E}Z_{1} \leq \mathbb{E}Z_{2}$ similarly. For any policy $\pi$, we define the operator $P^{\pi}$ by
$$
\left(P^{\pi} Z\right)(s, a)=\mathbb{E}\left[Z\left(S^{\prime}, A^{\prime}\right) \mid S^{\prime} \sim P(\cdot \mid s, a), A^{\prime} \sim \pi\left(\cdot \mid S^{\prime}\right)\right].
$$


{\noindent Step (1):} In the first step, we establish a recursion that relates $\mathbb{E}Z^{*}-\mathbb{E} Z^{(m)}$ with $\mathbb{E}Z^{*}-\mathbb{E} Z^{(m-1)}$ to measure the sub-optimality of the value function $\mathbb{E} Z^{(m-1)}$.

We first establish an upper bound for $\mathbb{E}Z^*-\mathbb{E}Z^{(m+1)}$ as follows. For each $m \in\{0, \ldots, M-1\}$, we have
\begin{align*}
\mathbb{E}Z^*-\mathbb{E}Z^{(m+1)} & =\mathbb{E}Z^*-\left(\mathbb{E}Z^{m+1}-\mathbb{E}\varrho_{m+1}\right)=\mathbb{E}Z^*-\mathbb{E}Z^{m+1}+\mathbb{E}\varrho_{m+1}=\mathbb{E}Z^*-\mathcal{T} \mathbb{E}Z^{(m)}+\mathbb{E}\varrho_{m+1} \\
& =\mathbb{E}Z^*-\mathcal{T}^{\pi^{*}} \mathbb{E}Z^{(m)}+\left(\mathbb{E}\mathcal{T}^{\pi^{*}} Z^{(m)}- \mathbb{E}\mathcal{T}Z^{(m)}\right)+\mathbb{E}\varrho_{m+1}\\
&\le \mathbb{E}Z^*- \mathbb{E}\mathcal{T}^{\pi^{*}}Z^{(m)}+\mathbb{E}\varrho_{m+1}
\end{align*}
where $\pi^{*}$ is the greedy policy with respect to $\mathbb{E}Z^*$, and the inequality follows from the fact that $ \mathbb{E} \mathcal{T}^{\pi^{*}}Z^{(m)} \leq  \mathbb{E} \mathcal{T}Z^{(m)}$.

Next, we establish a lower bound for $\mathbb{E}Z^*-\mathbb{E}Z^{(m+1)}$ based on $\mathbb{E}Z^*-\mathbb{E}Z^{(m)}$. By the definition of Bellman optimality operator, we have $ \mathbb{E}\mathcal{T}^{\pi_{m}}Z^{(m)}=\mathbb{E}\mathcal{T} Z^{(m)}$ and $ \mathbb{E}\mathcal{T}Z^* \geq \mathbb{E}\mathcal{T}^{\pi_{m}} Z^*$. Since $\mathbb{E}Z^*$ is the unique fixed point of $\mathcal{T}$, it holds that
\begin{align*}
\mathbb{E}Z^*-\mathbb{E}Z^{(m+1)} & =\mathbb{E}Z^*- \mathbb{E}\mathcal{T}Z^{(m)}+\mathbb{E}\varrho_{m+1}=\mathbb{E}Z^*- \mathbb{E}\mathcal{T}^{\pi_{m}}Z^{(m)}+\mathbb{E}\varrho_{m+1}\\
&=\mathbb{E}Z^*-\mathbb{E}\mathcal{T}^{\pi_{m}} Z^*+\left(\mathbb{E}\mathcal{T}^{\pi_{m}} Z^*-\mathbb{E}\mathcal{T}^{\pi_{m}} Z^{(m)}\right)+\mathbb{E}\varrho_{m+1} \\
& =\left(\mathbb{E}\mathcal{T} Z^*-\mathbb{E}\mathcal{T}^{\pi_{m}} Z^*\right)+\left( \mathbb{E}\mathcal{T}^{\pi_{m}}Z^*-\mathbb{E}\mathcal{T}^{\pi_{m}} Z^{(m)}\right)+\mathbb{E}\varrho_{m+1} \\
&\geq\left(\mathbb{E}\mathcal{T}^{\pi_{m}} Z^*-\mathbb{E}\mathcal{T}^{\pi_{m}} Z^{(m)}\right)+\mathbb{E}\varrho_{m+1}. 
\end{align*}

Thus, combining the lower and upper bound, we have for any $m \in\{0, \ldots, M-1\}$,
$$\mathbb{E}\mathcal{T}^{\pi_{m}}Z^*-\mathbb{E}\mathcal{T}^{\pi_{m}} Z^{(m)}+\mathbb{E}\varrho_{m+1} \leq \mathbb{E}Z^*-\mathbb{E}Z^{(m+1)} \leq \mathbb{E}\mathcal{T}^{\pi^{*}} Z^*-\mathbb{E}\mathcal{T}^{\pi^{*}} Z^{(m)}+\mathbb{E}\varrho_{m+1}.$$

The inequalities above show that the error $\mathbb{E}Z^*-\mathbb{E}Z^{(m+1)}$ can be controlled by the summation of a term involving $\mathbb{E}Z^*-\mathbb{E}Z^{(m)}$ and the error $\mathbb{E}\varrho_{m+1}$, which is induced by approximating the action-value function. Using the operator $P^{\pi}$, we can express inequality in a more neat form,
$$\gamma \cdot P^{\pi^{*}}\left(Z^*-Z^{(m)}\right)+\mathbb{E}\varrho_{m+1} \geq \mathbb{E}Z^*-\mathbb{E}Z^{(m+1)} \geq \gamma \cdot P^{\pi_{m}}\left(Z^*-Z^{(m)}\right)+\mathbb{E}\varrho_{m+1}.$$
It is worth noting that that $P^{\pi}$ is a linear operator, this leads to the following characterization of the multi-step error. For any $m, \ell \in\{0,1, \ldots, M-1\}$ with $m<\ell$, we have
\begin{align*}
& \mathbb{E}Z^*-\mathbb{E}Z^{(\ell)} \leq \sum_{i=k}^{\ell-1} \gamma^{\ell-1-i} \cdot\left(P^{\pi^{*}}\right)^{\ell-1-i} \varrho_{i+1}+\gamma^{\ell-k} \cdot\left(P^{\pi^{*}}\right)^{\ell-k}\left(Z^*-Z^{(m)}\right), \\
& \mathbb{E}Z^*-\mathbb{E}Z^{(\ell)} \geq \sum_{i=k}^{\ell-1} \gamma^{\ell-1-i} \cdot\left(P^{\pi_{\ell-1}} P^{\pi_{\ell-2}} \cdots P^{\pi_{i+1}}\right) \varrho_{i+1}+\gamma^{\ell-k} \cdot\left(P^{\pi_{\ell-1}} P^{\pi_{\ell-2}} \cdots P^{\pi_{m}}\right)\left(Z^*-Z^{(m)}\right).
\end{align*}
Here  $P^{\pi} P^{\pi^{\prime}}$ and $\left(P^{\pi}\right)^{k}$ to denote the composition of operators.

{\noindent Step (2):} In the second step, our goal is to quantify the suboptimality of $\mathbb{E} Z^{\pi_{m}}$, which is the action-value function corresponding to $\pi_{m}$ where $\pi_{m}$ is the greedy policy with respect to $Z^{(m)}$. We establish an upper bound for $\mathbb{E}Z^*-\mathbb{E}Z^{(\pi_{m})}$. Note that we have $\mathbb{E}Z^*=\mathcal{T}^{\pi^{*}} \mathbb{E}Z^*$ and $\mathbb{E}Z^{\pi_{m}}=\mathcal{T}^{\pi_{k}} \mathbb{E}Z^{\pi_{m}}$. Then we have
\begin{align*}
\mathbb{E}Z^*-\mathbb{E}Z^{\pi_{m}} & = \mathbb{E}\mathcal{T}^{\pi^{*}}Z^*- \mathbb{E}\mathcal{T}^{\pi_{m}}Z^{\pi_{m}}\\
&=\mathbb{E}\mathcal{T}^{\pi^{*}} Z^*+\left(- \mathbb{E}\mathcal{T}^{\pi^{*}}Z^{(m)}+\mathbb{E}\mathcal{T}^{\pi^{*}} Z^{(m)}\right)+\left(- \mathbb{E}\mathcal{T}^{\pi_{m}}Z^{(m)}+\mathbb{E}\mathcal{T}^{\pi_{m}} Z^{(m)}\right)- \mathbb{E}\mathcal{T}^{\pi_{m}}Z^{\pi_{m}} \\
& =\left(\mathbb{E}\mathcal{T}^{\pi^{*}} Z^{(m)}- \mathbb{E}\mathcal{T}^{\pi_{m}}Z^{(m)}\right)+\left(\mathbb{E}\mathcal{T}^{\pi^{*}} Z^*- \mathbb{E}\mathcal{T}^{\pi^{*}}Z^{(m)}\right)+\left(\mathbb{E}\mathcal{T}^{\pi_{m}} Z^{(m)}-\mathbb{E}\mathcal{T}^{\pi_{m}} Z^{\pi_{m}}\right).
\end{align*}
Next, we quantify the three terms on the right-hand side respectively. First, it is easy to see that $
\mathbb{E}\mathcal{T}^{\pi^{*}} Z^{(m)}-\mathbb{E}\mathcal{T}^{\pi_{m}} Z^{(m)}=\mathbb{E}\mathcal{T}^{\pi^{*}} Z^{(m)}-\mathbb{E}\mathcal{T} Z^{(m)} \leq 0$. Second, by the definition of the operator $P^{\pi}$, we have $\mathbb{E}\mathcal{T}^{\pi^{*}} Z^*- \mathbb{E}\mathcal{T}^{\pi^{*}}Z^{(m)}=\gamma \cdot P^{\pi^{*}}\left(Z^*-Z^{(m)}\right)$ and $\mathbb{E}\mathcal{T}^{\pi_{m}}Z^{(m)}- \mathbb{E}\mathcal{T}^{\pi_{m}}Z^{\pi_{m}}=\gamma \cdot P^{\pi_{m}}\left(Z^{(m)}-Z^{\pi_{m}}\right)
$. Combining the above results, we have
\begin{align*}
\mathbb{E}Z^*-\mathbb{E}Z^{\pi_{m}} & \leq \gamma \cdot P^{\pi^{*}}\left(Z^*-Z^{(m)}\right)+\gamma \cdot P^{\pi_{m}}\left(Z^{(m)}-Z^{(\pi_{m})}\right) \\
& =\gamma \cdot\left(P^{\pi^{*}}-P^{\pi_{m}}\right)\left(Z^*-Z^{(m)}\right)+\gamma \cdot P^{\pi_{m}}\left(Z^*-Z^{\pi_{m}}\right),
\end{align*}
which further implies that
$$\left(I-\gamma \cdot P^{\pi_{m}}\right)\left(Z^*-Z^{\pi_{m}}\right) \leq \gamma \cdot\left(P^{\pi^{*}}-P^{\pi_{m}}\right)\left(Z^*-Z^{(m)}\right).$$
Here $I$ is the identity operator. Since $\mathcal{T}^{\pi}$ is a $\gamma$-contractive operator for any policy $\pi, I-\gamma \cdot P^{\pi}$ is invertible. Then, we have
$$0 \leq \mathbb{E}Z^*-\mathbb{E}Z^{\pi_{m}} \leq \gamma \cdot\left(I-\gamma \cdot P^{\pi_{m}}\right)^{-1}\left[P^{\pi^{*}}\left(Z^*-Z^{(m)}\right)-P^{\pi_{m}}\left(Z^*-Z^{(m)}\right)\right],$$
which relates $\mathbb{E}Z^*-\mathbb{E}Z^{\pi_{m}}$ with $\mathbb{E}Z^*-\mathbb{E}Z^{(m)}$. In the following, we obtain the multiple-step error bounds for $\mathbb{E}Z^{\pi_{m}}$. First note that, by the definition of $P^{\pi}$, for any functions $f_{1}, f_{2}: \mathcal{S} \times \mathcal{A} \rightarrow \mathbb{R}$ satisfying $f_{1} \geq f_{2}$, we have $P^{\pi} f_{1} \geq P^{\pi} f_{2}$. Combining this inequality with the upper and lower bound in step (i), we have that, for any $m<\ell$,
\begin{gather*}
P^{\pi^{*}}\left(Z^*-Z^{(\ell)}\right) \leq \sum_{i=m}^{\ell-1} \gamma^{\ell-1-i} \cdot\left(P^{\pi^{*}}\right)^{\ell-i} \varrho_{i+1}+\gamma^{\ell-m} \cdot\left(P^{\pi^{*}}\right)^{\ell-m+1}\left(Z^*-Z^{(m)}\right), \\
P^{\pi_{\ell}}\left(Z^*-Z^{(\ell)}\right) \geq \sum_{i=m}^{\ell-1} \gamma^{\ell-1-i} \cdot\left(P^{\pi_{\ell}} P^{\pi_{\ell-1}} \cdots P^{\pi_{i+1}}\right) \varrho_{i+1}+\gamma^{\ell-m} \cdot\left(P^{\pi_{\ell}} P^{\pi_{\ell-1}} \cdots P^{\pi_{m}}\right)\left(Z^*-Z^{(m)}\right) .
\end{gather*}
Combining the above results, we further have
\begin{align*}
0 \leq \mathbb{E}Z^*-\mathbb{E}Z^{\pi_{\ell}} \leq\left(I-\gamma \cdot P^{\pi_{\ell}}\right)^{-1}\{ & \sum_{i=k}^{\ell-1} \gamma^{\ell-i} \cdot\left[\left(P^{\pi^{*}}\right)^{\ell-i}-\left(P^{\pi_{\ell}} P^{\pi_{\ell-1}} \cdots P^{\pi_{i+1}}\right)\right] \varrho_{i+1} \\
& \left.+\gamma^{\ell+1-k} \cdot\left[\left(P^{\pi^{*}}\right)^{\ell-k+1}-\left(P^{\pi_{\ell}} P^{\pi_{\ell-1}} \cdots P^{\pi_{k}}\right)\right]\left(Z^*-Z^{(m)}\right)\right\}
\end{align*}
for any $m<\ell$. To quantify the error of $\mathbb{E}Z^{\pi_K}$, we set $\ell=M$ and $m=0$ in above inequality to obtain
\begin{align*}
0 \leq \mathbb{E}Z^*-\mathbb{E}Z^{\pi_M} \leq\left(I-\gamma P^{\pi_{M}}\right)^{-1}\{ & \sum_{i=0}^{M-1} \gamma^{M-i} \cdot\left[\left(P^{\pi^{*}}\right)^{M-i}-\left(P^{\pi_{M}} P^{\pi_{M-1}} \cdots P^{\pi_{i+1}}\right)\right] \varrho_{i+1} \\
& \left.+\gamma^{M+1} \cdot\left[\left(P^{\pi^{*}}\right)^{M+1}-\left(P^{\pi_{M}} P^{\pi_{M-1}} \cdots P^{\pi_{0}}\right)\right]\left(Z^*-Z^{(0)}\right)\right\}.
\end{align*}

To simplify the presentation, we denote
$$\alpha_{i}=\frac{(1-\gamma) \gamma^{M-i-1}}{1-\gamma^{M+1}}, \text { for } 0 \leq i \leq M-1, \quad \text { and } \quad \alpha_{M}=\frac{(1-\gamma) \gamma^{M}}{1-\gamma^{M+1}},$$
where It is easy to check$\sum_{i=0}^{M} \alpha_{i}=1$. Also we define linear operators $\left\{O_{m}\right\}_{m=0}^{M}$ by
\begin{align*}
O_{i} & =(1-\gamma) / 2 \cdot\left(I-\gamma P^{\pi_{M}}\right)^{-1}\left[\left(P^{\pi^{*}}\right)^{M-i}+\left(P^{\pi_{M}} P^{\pi_{M-1}} \cdots P^{\pi_{i+1}}\right)\right], \quad \text { for } 0 \leq i \leq M-1 \\
O_{M} & =(1-\gamma) / 2 \cdot\left(I-\gamma P^{\pi_{M}}\right)^{-1}\left[\left(P^{\pi^{*}}\right)^{M+1}+\left(P^{\pi_{M}} P^{\pi_{M-1}} \cdots P^{\pi_{0}}\right)\right].
\end{align*}
Then for any $(s, a) \in \mathcal{S} \times \mathcal{A}$, we have
\begin{align*}
& \mathbb{E}Z^*(s, a)-\mathbb{E}Z^{\pi_M}(s, a) \leq \frac{2 \gamma\left(1-\gamma^{M+1}\right)}{(1-\gamma)^{2}} \cdot\left[\sum_{i=0}^{M-1} \alpha_{i} \cdot\left(O_{i}\varrho_{i+1}\right)(s, a)+\alpha_{M} \cdot\left(O_{M}(Z^*-Z^{(0)})\right)(s, a)\right],
\end{align*}
where both $O_{i}(\varrho_{i+1})$ and $O_{M}(Z^*-Z^{(0)})$ are functions defined on $\mathcal{S} \times \mathcal{A}$.

{\noindent Step (3):} In the third step, we can conclude the proof by establishing an upper bound for $\left\|\mathbb{E}Z^{*}-\mathbb{E}Z^{\pi_{M}}\right\|_{1,\mu}$. Here $\mu \in \mathcal{P}(\mathcal{S} \times \mathcal{A})$ is a fixed probability distribution. To simplify the notation, for any measurable function $f: \mathcal{S} \times \mathcal{A} \rightarrow \mathbb{R}$, we denote $\mu(f)$ to be the expectation of $f$ under $\mu$, i.e., $\mu(f)=\int_{\mathcal{S} \times \mathcal{A}} f(s, a) \mathrm{d} \mu(s, a)$. Then we have
\begin{align*}
\left\|\mathbb{E}Z^*-\mathbb{E}Z^{\pi_M}\right\|_{1, \mu} & =\mu\left(\left|\mathbb{E}Z^*-\mathbb{E}Z^{\pi_M}\right|\right) \\
& \leq \frac{2 \gamma\left(1-\gamma^{M+1}\right)}{(1-\gamma)^{2}} \cdot \mu\left[\sum_{i=0}^{M-1} \alpha_{i} \cdot\left(O_{i}\left|\varrho_{i+1}\right|\right)+\alpha_{M} \cdot\left|O_{M}(Z^*-Z^{(0)})\right|\right]\\
&\leq \frac{2 \gamma\left(1-\gamma^{M+1}\right)}{(1-\gamma)^{2}} \cdot\left[\sum_{i=0}^{M-1} \alpha_{i} \cdot \mu\left(O_{i}\left|\varrho_{i+1}\right|\right)+\alpha_{M} \cdot \mu\left(\left|O_{M}(Z^*-Z^{(0)})\right|\right)\right].
\end{align*}
Note that the rewards $\mathbb{E}\vert R(s, a)\vert< C_{p,R}$ for any $s\in\mathcal{S}$ and $a\in\mathcal{A}$. Then $\mu\left(\left|(O_{M}(Z^*-Z^{(0)})\right|\right) \leq 2 \cdot C_{p,R} /(1-\gamma)$. For any $i \in\{0, \ldots, M-1\}$, by expanding $\left(1-\gamma P^{\pi_{M}}\right)^{-1}$ we have
\begin{align*}
\mu\left(O_{i}\left|\varrho_{i+1}\right|\right) & =\mu\left\{\frac{1-\gamma}{2} \cdot\left(1-\gamma P^{\pi_{M}}\right)^{-1}\left[\left(P^{\pi^{*}}\right)^{M-i}+\left(P^{\pi_{M}} P^{\pi_{M-1}} \cdots P^{\pi_{i+1}}\right)\right]\left|\varrho_{i+1}\right|\right\} \\
& =\frac{1-\gamma}{2} \cdot \mu\left\{\sum_{j=0}^{\infty} \gamma^{j} \cdot\left[\left(P^{\pi_{M}}\right)^{j}\left(P^{\pi^{*}}\right)^{M-i}+\left(P^{\pi_{M}}\right)^{j+1}\left(P^{\pi_{M-1}} \cdots P^{\pi_{i+1}}\right)\right]\left|\varrho_{i+1}\right|\right\}.
\end{align*}
To give an upper bound of the right-hand side, we consider the following quantity
$$\mu\left[\left(P^{\pi_{M}}\right)^{j}\left(P^{\tau_{m}} P^{\tau_{m-1}} \cdots P^{\tau_{1}}\right) f\right]=\int_{\mathcal{S} \times \mathcal{A}}\left[\left(P^{\pi_{M}}\right)^{j}\left(P^{\tau_{m}} P^{\tau_{m-1}} \cdots P^{\tau_{1}}\right) f\right](s, a) \mathrm{d} \mu(s, a).$$
Here $\tau_{1}, \ldots, \tau_{m}$ are $m$ policies. Recall that $P^{\pi}$ is the transition operator of a Markov process defined on $\mathcal{S} \times \mathcal{A}$ for any policy $\pi$. Then the integral on the right-hand side in the above equation is the expectation of the function $f\left(X_{t}\right)$, where $\left\{X_{t}\right\}_{t \geq 0}$ is a Markov process defined on $\mathcal{S} \times \mathcal{A}$. Such a Markov process has initial distribution $X_{0} \sim \mu$. The first $m$ transition operators are $\left\{P^{\tau_{j}}\right\}_{j \in[m]}$, followed by $j$ identical transition operators $P^{\pi_{K}}$. Hence, $\left(P^{\pi_{K}}\right)^{j}\left(P^{\tau_{m}} P^{\tau_{m-1}} \cdots P^{\tau_{1}}\right) \mu$ is the marginal distribution of $X_{j+m}$, which we denote by $\widetilde{\mu}_{j}$ for notational simplicity. Then we have
$$\mu\left[\left(P^{\pi_{K}}\right)^{j}\left(P^{\tau_{m}} P^{\tau_{m-1}} \cdots P^{\tau_{1}}\right) f\right]=\mathbb{E}\left[f\left(X_{j+m}\right)\right]=\widetilde{\mu}_{j}(f)=\int_{\mathcal{S} \times \mathcal{A}} f(s, a) \mathrm{d} \widetilde{\mu}_{j}(s, a)$$
for any measurable function $f$ on $\mathcal{S} \times \mathcal{A}$. By Cauchy-Schwarz inequality, for any measure $\sigma$ we have
$$\widetilde{\mu}_{j}(f) \leq\left[\int_{\mathcal{S} \times \mathcal{A}}\left|\frac{\mathrm{d} \widetilde{\mu}_{j}}{\mathrm{~d} \sigma}(s, a)\right|^{2} \mathrm{~d} \sigma(s, a)\right]^{1 / 2}\left[\int_{\mathcal{S} \times \mathcal{A}}|f(s, a)|^{2} \mathrm{~d} \sigma(s, a)\right]^{1 / 2}\le C(m+j ; \mu, \sigma) \cdot\|f\|_{\sigma},$$
where $\mathrm{d} \widetilde{\mu}_{j} / \mathrm{d} \sigma: \mathcal{S} \times \mathcal{A} \rightarrow \mathbb{R}$ is the Radon-Nikodym derivative. Then we have
\begin{align*}
\mu\left(O_{i}\left|\varrho_{i+1}\right|\right) & =\frac{1-\gamma}{2} \cdot \sum_{j=0}^{\infty} \gamma^{j} \cdot\left\{\mu\left[\left(P^{\pi_{M}}\right)^{j}\left(P^{\pi^{*}}\right)^{M-i}\left|\varrho_{i+1}\right|\right]+\mu\left[\left(P^{\pi_{M}}\right)^{j+1}\left(P^{\pi_{M-1}} \cdots P^{\pi_{i+1}}\right)\left|\varrho_{i+1}\right|\right]\right\} \\
& \leq(1-\gamma) \cdot \sum_{j=0}^{\infty} \gamma^{j} \cdot C(M-i+j ; \mu, \sigma_{i+1}) \cdot\left\|\varrho_{i+1}\right\|_{\sigma_{i+1}}.
\end{align*}
Combining the above inequalities, we have
\begin{align*}
\| \mathbb{E}Z^* & -\mathbb{E}Z^{\pi_K} \|_{1, \mu} \leq \frac{2 \gamma\left(1-\gamma^{M+1}\right)}{(1-\gamma)^{2}} \cdot\left[\sum_{i=0}^{M-1} \alpha_{i} \cdot \mu\left(O_{i}\left|\varrho_{i+1}\right|\right)+\alpha_{M} \cdot \mu\left(\left|O_{M}(Z^*-Z^{(0)})\right|\right)\right] \\
& \leq \frac{2 \gamma\left(1-\gamma^{M+1}\right)}{(1-\gamma)} \cdot\left[\sum_{i=0}^{M-1} \sum_{j=0}^{\infty} \alpha_{i} \cdot \gamma^{j} \cdot C(M-i+j ; \mu, \sigma_{i+1}) \cdot\left\|\varrho_{i+1}\right\|_{\sigma_{i+1}}\right]+\frac{4  \alpha_{M} C_{p,R}\gamma\left(1-\gamma^{M+1}\right)}{(1-\gamma)^{3}}.
\end{align*}
Recall that we define $\varepsilon_{\max}=\max _{i \in[M]}\left\|\varrho_{i}\right\|_{\sigma_i}$. Then $\left\|\mathbb{E}Z^*-\mathbb{E}Z^{\pi_M}\right\|_{1, \mu}$ is further upper bounded by
\begin{align*}
& \frac{2 \gamma\left(1-\gamma^{K+1}\right)}{(1-\gamma)} \cdot\left[\sum_{i=0}^{M-1} \sum_{j=0}^{\infty} \alpha_{i} \cdot \gamma^{j} \cdot C(M-i+j ; \mu, \sigma_{i+1})\right] \cdot \varepsilon_{\max }+\frac{4 C_{p,R}\gamma\left(1-\gamma^{M+1}\right)}{(1-\gamma)^{3}} \cdot \alpha_{M}  \\
=& \frac{2 \gamma\left(1-\gamma^{M+1}\right)}{(1-\gamma)} \cdot\left[\sum_{i=0}^{M-1} \sum_{j=0}^{\infty} \frac{(1-\gamma) \gamma^{M-i-1}}{1-\gamma^{M+1}} \cdot \gamma^{j} \cdot C(M-i+j ; \mu, \sigma_{i+1})\right] \cdot \varepsilon_{\max }+\frac{4 C_{p,R}\gamma^{M+1}}{(1-\gamma)^{2}}
\end{align*}
where the last equality follows from the definition of $\left\{\alpha_{i}\right\}_{0 \leq i \leq M}$. We can further simplify the summation to obtain
\begin{align*}
& \sum_{i=0}^{M-1} \sum_{j=0}^{\infty} \frac{(1-\gamma) \gamma^{M-i-1}}{1-\gamma^{M+1}} \cdot \gamma^{j} \cdot C(M-i+j ; \mu, \sigma_{i+1}) \\
& \quad=\frac{1-\gamma}{1-\gamma^{M+1}} \sum_{j=0}^{\infty} \sum_{i=0}^{M-1} \gamma^{M-i+j-1} \cdot C(M-i+j ; \mu, \sigma_{i+1}) \\
& \quad\leq\frac{1-\gamma}{1-\gamma^{M+1}} \sum_{j=1}^{\infty} \sum_{m=0}^{M} \gamma^{j} \cdot C(M-m+j ; \mu, \sigma_{m}) 
\leq \frac{c_{M,\mu}}{\left(1-\gamma^{M+1}\right)(1-\gamma)},
\end{align*}
which concludes the proof.

\section{Supporting Definitions and Lemmas}\label{appendix_supp}
In this subsection, we present some preliminary concepts and theorems that will be used later regarding distributional reinforcement learning, complexity theory (Chapter 3 of \citealt{mohri2018foundations}) and network learning (Chapters 11 and 12 of \citealt{anthony1999neural}) to prove the non-asymptotic error bounds in Section \ref{sec_theory}.

\begin{customtdef}{S1}[Covering Number]\label{cove_number}
Let $\mathcal{F}$ be a class of functions from $\mathcal{X}^p$ to $\mathbb{R}$.
For any given sequence $x=(x_1,\ldots,x_p)\in\mathcal{X}^p$, let
\begin{equation*}
	\mathcal{F}|_{x}:= \{(f(x_1),\ldots,f(x_p)): f\in\mathcal{F}\}\subset \mathbb{R}^p.
\end{equation*}
For a positive constant $\delta$, the covering number $\mathcal{N}(\delta,\mathcal{F}|_{x},\|\cdot\|)$ is the smallest positive integer $N$ such that there exist  $f_1,\ldots,f_N\in\mathcal{F}|_{x}$ and  $\mathcal{F}|_{x}\subset \bigcup_{i=1}^N\{f: \|f-f_i\|\leq \delta\}$, i.e., $\mathcal{F}|_{x}$ is fully covered by $N$ spherical balls centered at $f_i$ with radius $\delta$ under the norm $\|\cdot\|$. Moreover, the uniform covering number $\mathcal{N}_p(\delta,\mathcal{F},\|\cdot\|)$ is defined to be the maximum covering number over all  $x$ in $\mathcal{X}^p$, i.e.,
\begin{equation*}
	\mathcal{N}_p(\delta,\mathcal{F},\|\cdot\|):=\max\{\mathcal{N}(\delta,\mathcal{F}|_{x},\|\cdot\|): x\in\mathcal{X}^p\}.
\end{equation*}
\end{customtdef}
\begin{customtdef}{S2}[Pseudo-Dimension]\label{def:pseudo_dimension}
The pseudo-dimension of $\mathcal{F}$, denoted as Pdim$(\mathcal{F})$, is defined as the largest integer $m$ for which there exist $(x_1,y_1),\ldots,(x_m,y_m)\in\mathcal{X}^p\times\mathbb{R}$ such that for all $\boldsymbol{\eta}\in\{0,1\}^m$, there exists $f\in\mathcal{F}$ so that the following two arguments are equivalent:
\begin{equation*}
	f(x_i)>y_i  \iff \eta_i, \quad \text{ for } i=1,\ldots,m.
\end{equation*}
\end{customtdef}
Both the covering number and pseudo-dimension quantify the complexity of a given function class from different aspects. Indeed, the following results are available to establish the connections between them.

\begin{customthm}{S3}[{Theorem 12.2 in \citealt{anthony1999neural}}]\label{thm:covering_number}
If $\mathcal{F}$ is a set of real functions from a domain to a bounded interval $[0, B]$ and the pseudo-dimension of $\mathcal{F}$ is $\text{Pdim}(\mathcal{F})$, then
\begin{equation*}
	\mathcal{N}_N(\delta,\mathcal{F},\|\cdot\|_{\infty})\leq\sum_{i=1}^{\text{Pdim}(\mathcal{F})}\binom{N}{i}\bigg(\frac{B}{\delta}\bigg)^i<\bigg(\frac{eNB}{\delta \cdot \text{Pdim}(\mathcal{F})}\bigg)^{\text{Pdim}(\mathcal{F})},
\end{equation*}
for $N\geq \text{Pdim}(\mathcal{F})$.
\end{customthm}

\begin{customthm}{S4}\label{lemma_K}
Suppose $X$ is a continuous random variable with finite $p$th moment $\mathbb{E}\vert X\vert^p<\infty$. Let $\mu = E[X]$ denote the mean, $F$ denote the cumulative distributional function and let \( q_k = F^{-1}\left(\frac{k}{K+1}\right) \) for \( k = 1, 2, \ldots, K \) denote the quantiles of $X$. Then we have
$$\left\vert\frac{1}{K} \sum_{k=1}^{K} q_k -\mathbb{E}[X]\right\vert\le \frac{C}{K},$$
where $C>0$ is a constant not depending on $K$.
\end{customthm}

{\noindent \bf Proof}:
Given $F^{-1}$ is continuous, by Whitney approximation theorem \citep[Chapter 2.4,][]{hirsch2012differential}, there exists a sufficient smooth function ${F}_*^{-1}$ that approximates $F^{-1}$ arbitrarily well such that $\int_{0}^1 F^{-1}(u)\, du=\int_{0}^1{F}_*^{-1}(u)\, du$ and ${F}_*^{-1}(u)={F}^{-1}(u)$ for $u=\frac{1}{K+1},\ldots,\frac{K}{K+1}$. In light of this, we can work with ${F}_*^{-1}$ or assume $F^{-1}$ is smooth in our analysis.

First, we prove the lemma when $X$ is bounded. When $X$ is bounded, the lower and upper limits $F^{-1}(0)$ and $F^{-1}(1)$ are finite.  Note that the true mean of the distribution is given by $\mathbb{E}(X) = \int_{0}^{1} F^{-1}(u)\, du.$ 
The integral can be approximated by the Riemann sum $\int_{0}^{1} F^{-1}(u) \, du \approx \frac{1}{K} \sum_{k=1}^{K} F^{-1}\left(\frac{k}{K}\right)$, and the approximation error can be bounded using the Euler-Maclaurin formula, which states that for a sufficiently smooth function \( g \):
\[
\int_{a}^{b} g(x) \, dx - \frac{b-a}{n} \sum_{k=1}^{n} g\left(a + k \frac{b-a}{n}\right) = -\frac{b-a}{2n} \left( g(a) + g(b) \right) + O\left(\frac{1}{n^2}\right).
\]
Applying this to our case with \( g(u) = F^{-1}(u) \), \( a = 0 \), \( b = 1 \), and \( n = K \), we have
\[
\int_{0}^{1} F^{-1}(u) \, du - \frac{1}{K} \sum_{k=1}^{K} F^{-1}\left(\frac{k}{K}\right) = -\frac{1}{2K} \left( F^{-1}(0) + F^{-1}(1) \right) + O\left(\frac{1}{K^2}\right).
\]
Since \( F^{-1}(0) \) and \( F^{-1}(1) \) are finite, then the leading error term is \( O\left(\frac{1}{K}\right) \). Recall we are using \( \frac{k}{K+1} \) instead of \( \frac{k}{K} \) for the approximation, we then need to control the difference using the Lipschitz continuity of \( F^{-1} \). Let \( L \) denote the Lipschitz constant of \( F^{-1} \), then:
$|F^{-1}\left({k}/{K}\right) - F^{-1}\left({k}/(K+1)\right)| \leq L \left|{k}/{K} - {k}/(K+1)\right|\le {L}/{K}.$
Summing over all \( k \), we have
$$\left\vert\frac{1}{K} \sum_{k=1}^{K} q_k -\mathbb{E}[X]\right\vert \leq \frac{C}{K},$$
where $C>0$ is a constant not depending on $K$.

Now we prove the lemma when $X$ is unbounded. 
Given random variable $X$ has finite $p$th moment, we have $\mathbb{P}(\vert X\vert>t)\le \mathbb{E}\vert X\vert^p/t^p$ for $t>0$ by the Markov inequality. Let $T=(\mathbb{E}\vert X\vert^p)^{1/p} \cdot (K+1)^{1/p}$, we obtain $\mathbb{P}(\vert X\vert>T)\le 1/(K+1)$, which implies $-T<q_1<q_K<T.$ Let $\tilde{X}=X\cdot I(\vert X\vert\le T)$, we have 
\begin{align*}
\left| \mathbb{E}[X]-\mathbb{E}[\tilde{X}]\right|\le  \mathbb{E}\left|X-\tilde{X}\right|&=\mathbb{E}\left|X\cdot I(|X|>T)\right|\\
&=\int_0^\infty \mathbb{P}(\left|X\right|\cdot I(|X|>T) >u)\, du\\
&=\int_0^T \mathbb{P}(|X|>T))\, du +\int_T^\infty \mathbb{P}(|X|>u)\, du\\
&\le \frac{2(\mathbb{E}\vert X\vert^p)^{1/p}}{(K+1)^{(p-1)/p}}.
\end{align*}
Meanwhile, $\tilde{X}$ is random variable bounded by $T=(\mathbb{E}\vert X\vert^p)^{1/p} \cdot (K+1)^{1/p}$. By the previous argument, we also have
$$\left\vert\frac{1}{K} \sum_{k=1}^{K} q_k -\mathbb{E}[\tilde{X}]\right\vert \leq \frac{C}{(K+1)^{(p-1)/p}},$$
where $C>0$ is a constant not depending on $K$. Combining the above results, we obtain 
$$\left\vert\frac{1}{K} \sum_{k=1}^{K} q_k -\mathbb{E}[X]\right\vert \leq \frac{C}{K^{(p-1)/p}},$$
for some constant $C>0$ not depending on $K$.

\begin{customtdef}{S3}[Expected Concentration Coefficients]\label{coeff}

Given probability measures $\nu_1,\nu_2$ on $\mathcal{S}\times\mathcal{A}$ that are absolutely continuous with respect to Lebesgue measure. Let $\{\pi_m\}_{m=1}^M$ be a sequence of policies. Suppose the initial state-action pair $(S_0,A_0)$ of the MDP has distribution $\nu_1$ and it takes actions following the sequence of policies $\{\pi_m\}_{m=1}^M$. We denote the distribution of $\{(S_m,A_m)\}_{m=0}^{M}$ by $P^{\pi_M}P^{\pi_{M-1}}\cdots P^{\pi_1}\nu_1$ and define the following concentrability coefficients
$$C(M,\nu_1,\nu_2):=\sup_{\pi_1,\ldots,\pi_M}\left(\mathbb{E}_{\nu_1}\left[ \left\vert \frac{dP^{\pi_M}P^{\pi_{M-1}}\cdots P^{\pi_1}\nu_1}{d\nu_2}\right\vert^2\right]\right)^{1/2},$$
where the supremum is taken over all possible policies.  Let $\sigma_m$ be the marginal distribution of the sampled station-action pair $(S^{(m)}_i,A^{(m)}_i)$ in the $m$th iteration in Algorithm \ref{alg-1}, let $\mu$ denote the distribution of $(S_0,A_0)$ and let 
$c_{M,\mu}:=(1-\gamma)^2\sum_{j=1}^{\infty} \sum_{m=0}^{M-1} \gamma^{j} \cdot C(M-m+j ; \mu, \sigma_{m}).$
We assume the constant $c_{M,\mu}<\infty$. In addition, given the distribution of $S_0$ and any policy $\pi$, we let $\mu_\pi$ denote the joint distribution of $(S_0,A_0)$ satisfying $(A_0|S_0)\sim \pi(\cdot|S_0)$. We denote $C_M:=\sup_{\pi} C_{M,\mu_\pi}$ by the supremum of $C_{M,\mu_\pi}$ over all possible policies $\pi$ and assume $C_M<\infty$. \end{customtdef}

Concentration coefficients $C_{M,\mu}$ in Definition \ref{coeff} quantify the similarity between $\mu$ and the distribution of the future states of the MDP when starting from $\sigma_m$. The finite concentration coefficients hold for a large class of systems MDPs, which essentially assumes that the sampling distribution $\mu$ has sufficient coverage over $\mathcal{S}\times\mathcal{A}$. It is also necessary for the theoretical development of batch reinforcement learning \citep{antos2007value,chen2019information,fan2020theoretical}.

\end{document}